\documentclass[pdflatex,sn-nature]{sn-jnl}

\usepackage{graphicx}%
\usepackage{multirow}%
\usepackage{amsmath,amssymb,amsfonts}%
\usepackage{amsthm}%
\usepackage{mathrsfs}%
\usepackage[title]{appendix}%
\usepackage{xcolor}%
\usepackage{textcomp}%
\usepackage{manyfoot}%
\usepackage{booktabs}%
\usepackage{algorithm}%
\usepackage{algorithmicx}%
\usepackage{algpseudocode}%
\usepackage{listings}%
\usepackage{enumitem}%
\usepackage{tabularx} 
\usepackage{booktabs} 
\usepackage{tcolorbox}
\usepackage{amssymb}
\usepackage{pifont}
\usepackage{graphicx}
\definecolor{myblue}{HTML}{4A90E2}
\definecolor{mygreen}{HTML}{50C878}
\definecolor{myred}{HTML}{D9534F}
\definecolor{mygray}{rgb}{0.5,0.5,0.5}
\definecolor{myorange}{HTML}{FF7A33}
\newcommand{\cmark}{\ding{51}}
\newcommand{\xmark}{\ding{55}}

\theoremstyle{thmstyleone}%
\theoremstyle{thmstyletwo}%

\theoremstyle{thmstylethree}%

\usepackage{multibib}
\newcites{app}{References}

\begin{document}

\title[Article Title]{Scaling Participation in Modular AI Systems}



\author*[1]{\fnm{Shangbin} \sur{Feng}}\email{shangbin@cs.washington.edu}

\author[1]{\fnm{Yike} \sur{Wang}}

\author[1]{\fnm{Weijia} \sur{Shi}}

\author[1]{\fnm{Luke} \sur{Zettlemoyer}}

\author[2]{\fnm{Yejin} \sur{Choi}}

\author[1]{\fnm{Yulia} \sur{Tsvetkov}}

\affil[1]{\orgname{University of Washington}, \orgaddress{\city{Seattle}, \state{WA}, \country{USA}}}

\affil[2]{\orgname{Stanford University}, \orgaddress{\city{Stanford}, \state{CA}, \country{USA}}}


\abstract{
Humanity is a mosaic of multifaceted talents and needs, and any truly intelligent AI must reflect that richness. Yet the LLMs used by all are built by the few -- a centralized market of monolithic AI models structurally ill-suited to capture the diversity of human knowledge, reasoning, and values. 
Here we introduce \emph{scaling participation}, a new paradigm in which \emph{modular}, \emph{community-sourced} AI systems are built from the bottom up through the contributions of diverse stakeholders. Participants contribute small models trained on their own interests and priorities; these models then collaborate in modular frameworks as compositional AI systems, repurposing existing collaboration algorithms for this bottom-up paradigm. 
Participatory AI systems outperform monolithic LLMs by up to 15.42\% (95\% CI: [10.09\%, 21.13\%]) across 15 tasks, such as reasoning and factuality, surpassing models with more parameters than all contributed components combined. Further experiments show that these systems are especially strong at representing diverse cultures, values, and communities, benefit from contributor diversity, substantially improve on each contributor's original priorities, and exhibit emergent capabilities that allow them to solve over 15\% of problems where all individual models fail. Scaling participation provides a technical foundation, demonstrated here with academic contributors and benchmark evaluations, for transitioning from the monolithic status quo toward an open, bottom-up, and collaborative AI future.}




\maketitle

\section{Introduction}\label{sec1}

As AI becomes more deeply intertwined with society's institutions -- guiding decisions in policy \cite{ingrams2022ai}, education \cite{aieducation2023}, healthcare \cite{chinta2025ai}, economy \cite{bughin2018notes}, and science \cite{asai2026synthesizing, lu2026towards} -- it becomes crucial that AI better reflect society's pluralistic values and priorities. But today, AI's promise and power are concentrated in the hands of a few private firms, embodied in black-box, monolithic models developed in secrecy. This makes accountability difficult to sustain, because opaque models erode public trust, and monolithic systems developed on skewed data and benchmarks cannot credibly claim to represent and serve all populations fairly \cite{sorensenposition, feng2025one}. If these AI systems are to mediate decisions that affect everyone, they must be open to scrutiny and shaped through broader participation. This paper advances a technical foundation toward this goal: a proof-of-concept demonstrating that modular AI systems built from decentralized contributions can rival centralized, monolithic development.

\begin{figure}[!t]
\centering
\includegraphics[width=0.9\textwidth]{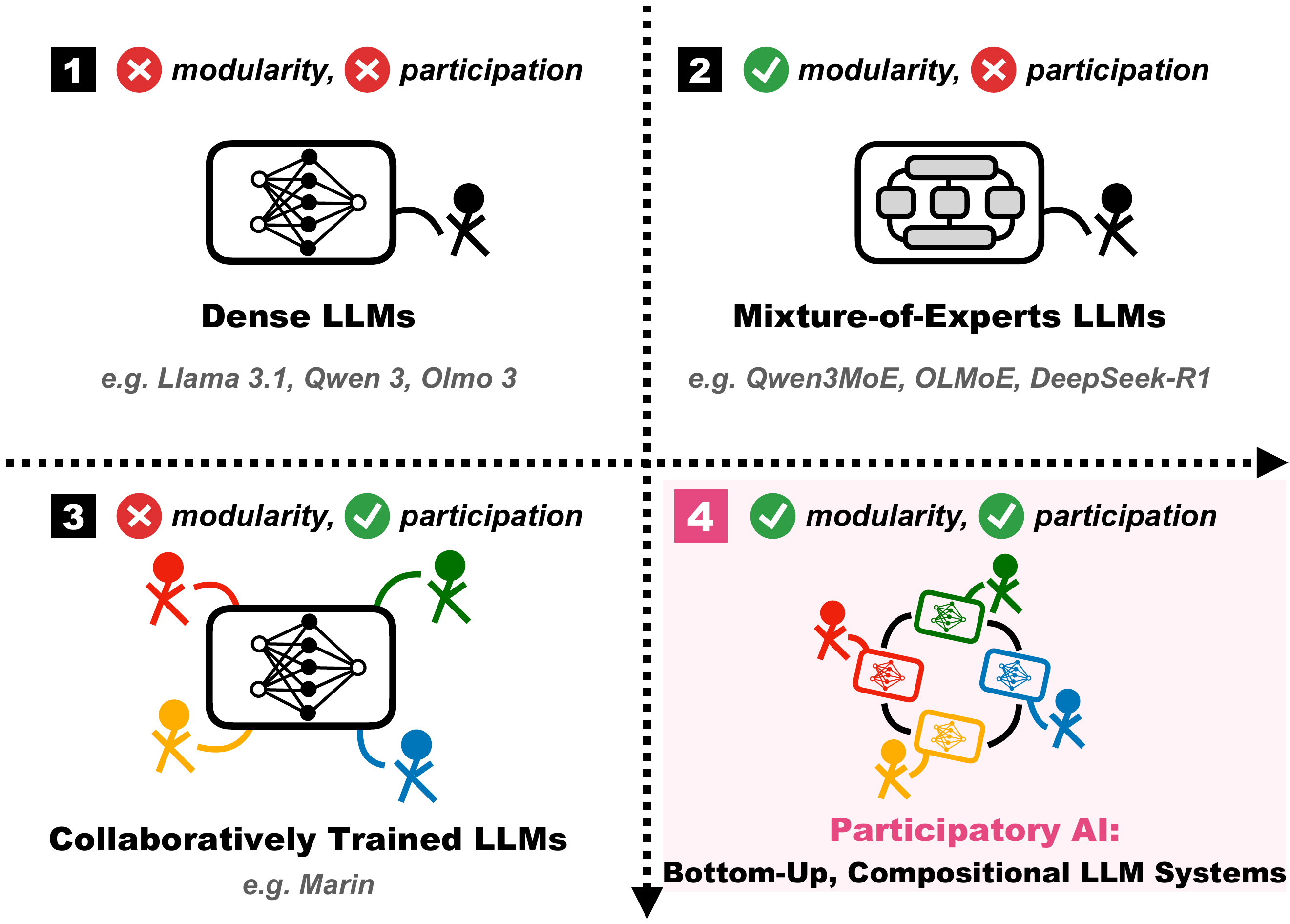}
\caption{\textbf{Overview of four paradigms of AI development.} 1) Single dense LLMs are developed without modularity or participation: they employ monolithic neural architectures and are solely developed by a single organization. 2) Mixture-of-Experts LLMs employ modular model architectures, but all the modular components are still solely developed by a single entity. 3) Initiatives such as Marin \cite{marin2025} take a participatory approach in the model training pipeline, while all efforts are eventually merged into a single monolithic model. 4) Our work uniquely champions participatory AI, specifically scaling participation in modular AI systems: participants contribute specialized language models to build modular systems from the bottom up, where diverse stakeholders will have a say in the futuristic paradigm of bottom-up AI development.}
\label{fig:overview}
\end{figure}

We propose a new paradigm: modular, participatory AI systems built to reflect the variety of the stakeholders they serve. We develop a framework in which many small and large models collaborate through explicit coordination algorithms. (Figure \ref{fig:overview}) It is modular because intelligence is composed from interacting components, rather than compressed into one model. It is participatory in two related senses: \emph{sociologically}, diverse stakeholders actively contribute to AI development by sharing models trained on their own data and priorities; \emph{technically}, the resulting system aggregates these decentralized contributions into a compositional AI system. The two senses are unified by design: sociological participation is operationalized through technical model composition algorithms. These participatory AI systems are, in this sense, \emph{community-sourced}: assembled from decentralized contributions, each retaining an identifiable, modular component. This framework sets two practical goals. The first is to scale up collaboration methods that combine the strengths of diverse models and produce outcomes stronger than any individual contributor or larger monolithic models alone. The second is to scale participation itself, so that many stakeholders can contribute models grounded in their own knowledge and priorities, allowing the system to adapt to particular populations, domains, and objectives.


We operationalize modular participatory AI by soliciting language models (LMs) released as academic research artifacts, as those are models designed with different architectures, trained on different domains, and represent diverse research agendas and priorities beyond commercial interests. With the 61 LMs obtained from decentralized contributors, we scale up diverse model collaboration systems where up to 32 LMs collaborate, compose, and complement each other as compositional AI systems. Specifically, these models act as modular components in 14 collaborative systems such as routing \cite{ongroutellm, fenggraphrouter}, multi-agent debate \cite{du2023improving, jiangsparta}, and model fusion \cite{yadav2023ties, yu2024language}. These collaboration algorithms are drawn from prior work; our contribution is harnessing them to instantiate a new paradigm: scaling participation by building modular AI from the bottom up through the real-world contributions of diverse stakeholders. We contrast this paradigm with prior approaches in Appendix \ref{sec:comparison_table}.

Extensive experiments demonstrate that successful participatory AI systems outperform non-modular and/or non-participatory LLMs -- some with more parameters than the 32 contributed LMs combined -- by up to 15.42\% (95\% CI: [10.09\%, 21.13\%]) across QA, reasoning, knowledge, safety, and instruction following tasks. Further analysis reveals that participatory AI is particularly strong at representing diverse cultures, values, and communities, benefits from the diversity of contributed model pools, solves problems where the participating LMs struggle individually, and consistently improves on the original priorities and use cases of model contributors. Overall, scaling participation improves AI capability and creates systems that are more transparent, more pluralistic, easier to update, more effective at reusing existing progress in AI, and better empowering contributors' agency to join or withdraw their modular components. Transitioning from centralized silos to participatory mosaics, our prototype of academic contributors and benchmark evaluations demonstrates the potential for future work in scaling participation in modular AI beyond academic scales, offering a new paradigm for building state-of-the-art AI systems and a technical foundation for a more open, democratized, bottom-up, and collaborative AI future.

\section{Methods}
\label{sec:method}

We propose to scale participation in modular AI systems with two steps: 1) soliciting participation, recruiting specialized language models trained in the wild as modular building blocks; 2) orchestrating participation, building bottom-up and compositional AI systems with these models and diverse model collaboration algorithms.

\vspace*{10pt}
\noindent \textbf{Soliciting Participation} We prototype participatory AI with language models trained as academic research artifacts: these models are generally specialized and represent the diverse priorities of the AI research agenda around the world. Specifically, to curate a pool of language models, we reach out to academic labs working on language models and natural language processing and ask for their participation by submitting the language models they trained in their research. We specifically employ the form shown in Figure \ref{fig:solicit_parti} (Appendix) to inform contributors of our objective and collect information about their model and research. By reaching out to 236 researchers around the world, we source a pool of 61 models from the research community spanning diverse institutions, countries, and model families; a full diversity breakdown across five dimensions is provided in Appendix B (Figure \ref{fig:pool_diversity}).

\vspace*{10pt}
\noindent \textbf{Orchestrating Participation} We employ a wide spectrum of model collaboration algorithms to prototype participatory AI, where models trained by diverse stakeholders collaborate, compose, and complement each other. Specifically, we employ 14 collaborative systems across three levels of information exchange across language models. \cite{feng2025one} We provide details for each collaborative algorithm in Appendix \ref{sec:details}.

For \emph{API-level collaboration}, featuring the routing and selection among candidate models:

\begin{itemize}[leftmargin=*]
    \item \textbf{Prompt Routing} \ Given a user instruction, we prompt an LLM to select the most fitting candidate model based on model descriptions provided by participants.
    \item \textbf{Trained Router} \ We evaluate all candidate models on a development set and determine the best model for each instruction. We then train a causal language model to learn to predict the best model for each instruction. At inference time, the trained router model dynamically assigns instructions to candidate models based on its predictions. \cite{ongroutellm}
    \item \textbf{Graph Routing} \ Similar to Trained Router, but a graph neural network is employed as the neural architecture for routing prediction, specifically by representation learning on a task-instruction-model network. \cite{fenggraphrouter}
    \item \textbf{Switch Generation} \ Instead of selecting one model to generate the full response, we train a switcher model to govern how multiple candidate models take turns to generate text patches towards a full response, whenever the skills and strengths of a particular model are most needed. \cite{feng2025don}
\end{itemize}

For \emph{text-level collaboration}, featuring the exchange of generated texts across models:

\begin{itemize}[leftmargin=*]
    \item \textbf{Multi-Agent Refine} \ Each candidate model first independently generates a response, then refines its response based on the responses from other models. After a few iterations, an LLM is instructed to summarize all models' responses in the last iteration. \cite{du2023improving}
    \item \textbf{Multi-Agent Finetuning} \ A training-based enhancement of multi-agent refine, where both generation and critique datasets are generated in the iterative process to fine-tune candidate models. At inference-time, the finetuned and adapted models generate responses and aggregate through majority vote. \cite{subramaniammultiagent}
    \item \textbf{LLM Blender} \ All candidate models first independently generate responses, we then train a ranker LLM to rank the responses and train a fuser LLM to aggregate the top-$k$ responses into a final response. \cite{jiang2023llm}
    \item \textbf{AggLM} \ An aggregator LLM is trained with reinforcement learning with verifiable rewards to balance ``easy'' cases where majority vote succeeds and ``hard'' cases, where minority answer recovery is required. At inference time, all candidate models independently generate responses and are processed by the aggregator LLM for a final response. \cite{zhao2025majority}
    \item \textbf{Sparta Alignment} \ Two candidate models enter a contest of instruction following and other candidate models evaluate their responses. The winning model would gain in reputation and vice versa, where the reputation of models indicates the weight of their judgments when evaluating other models' responses. All models are trained on the (winning response, losing response) pairs obtained in collective judgement through preference optimization. \cite{jiangsparta}
    \item \textbf{Heterogeneous Swarms} \ We orchestrate multiple candidate models in a directed acyclic graph (DAG), where one model's output becomes part of another model's input based on directed edges. We maintain a population of $k$ DAGs and employ particle optimization on continuous adjacency matrices to optimize the graph-based collaboration structures. \cite{fengheterogeneous}
\end{itemize}

For \emph{Weight-level collaboration}, featuring the arithmetic and search with model parameters:

\begin{itemize}[leftmargin=*]
    \item \textbf{Dare-Ties} \ Dare-Ties is a combination of two popular model merging techniques, focusing on pruning and the sign consensus of parameter values across multiple models. \cite{yadav2023ties, yu2024language}
    \item \textbf{Greedy Soup} \ We first evaluate all candidate models on the development set and sort them in descending order of performance. Starting from the best model, we iteratively add the next model to the ``soup'' (the current pool of models being considered), average the weights of models in the soup, and evaluate the merged model on the development set. We retain the model if performance improves over the previous best, and move on to the next model. \cite{wortsman2022model}
    \item \textbf{Weight Extrapolation (ExPO)} \ We evaluate models on the development set, select the top-$k$ and bottom-$k$ performing models, and extrapolate their parameters into a new model with extrapolate with $\mathbf{x}_{\textit{expo}} = \mathbf{x}_{\textit{top-k}} + \alpha (\mathbf{x}_{\textit{top-k}} - \mathbf{x}_{\textit{bottom-k}})$. \cite{zheng2024weak}
    \item \textbf{Model Swarms} \ Multiple candidate models collaboratively search in the model parameter space to improve performance on the development set, where the optimization is instantiated with particle swarm optimization, treating the pool of models as the population. \cite{feng2025model}
\end{itemize}
Please note that weight-level collaboration often requires participating models to share the same model architecture: we take one extra step to perform knowledge distillation \cite{kim2016sequence} from the contributed models to the same model architecture, so that we unify the model architecture for weight-level methods while preserving the specialization of participating models. Using the original training data instead is not available by design, since participants share trained models rather than data. We give the distillation setup in Section \ref{sec:experiment_settings} and Appendix \ref{sec:details}, and compare it against jointly distilling from all 32 models at once in Appendix \ref{sec:joint_distill}.

To sum up, we experiment with 61 models sourced from diverse academic research around the globe and 14 collaboration algorithms for bottom-up, compositional AI systems. To the best of our knowledge, this work is the first attempt at a modular and participatory approach towards AI development. The 14 collaboration algorithms are drawn from prior published work; our unique contribution is the paradigm of \emph{scaling participation} itself, where real-world contributions from decentralized stakeholders are orchestrated into compositional AI systems at scale.

\section{Experiment Settings}
\label{sec:experiment_settings}

\noindent \textbf{Data and Evaluation} We evaluate models and systems across 5 domains and 15 tasks in total: \textbf{general QA} (\emph{AGIEval}, \emph{ARC-challenge}, \emph{MMLU-redux}), \textbf{reasoning} (\emph{BigBench-Hard}, \emph{GSM8k}, \emph{MATH}, \emph{Sciriff}), \textbf{knowledge} (\emph{WikiDYK}, \emph{PopQA}, \emph{BLEND}), \textbf{safety} (\emph{TruthfulQA}, \emph{CocoNot}), and \textbf{instruction following} (\emph{AlpacaEval}, \emph{Wildchat}, \emph{Human Interest}); full dataset details in Appendix \ref{sec:details}. For each task, we create a held-out split for model selection as well as a development and a test split for evaluation. We report the macro-average performance of tasks in each domain.

\vspace*{10pt}
\noindent \textbf{Models and Implementation} We first evaluate the 61 community-sourced models contributed by diverse academics (Section \ref{sec:method}) on the mixture of held-out splits across tasks. Since we aim to scale collaborative systems up to 32 models, we retain the top 32 models as the default pool while employing other models for further analysis. This selection is not merely the source of the headline gains: collaborative systems built from random draws of the same pool size still show the same scaling trend, though below top-selected pools (Appendix \ref{sec:random_draws}). To run weight-level collaboration where models usually need to share the same architecture, we employ knowledge distillation with each of the 32 models as teacher and \textsc{Qwen-2.5-7B} as student, to obtain distilled models representing the contributed models. For these weight-level methods the modular components are therefore not the original contributed checkpoints: they are \textsc{Qwen-2.5-7B} student checkpoints separately sequence-distilled from each contributor, so the weight-level experiments measure collaboration among compressed, contributor-derived behavioral signals under a shared architecture. API-level and text-level methods use the original contributed models directly. We then build model collaboration systems with the first 2, 4, 8, 16, and 32 models in the pool: we specifically employ MoCo \cite{feng2026moco}, a one-stop library for executing diverse model collaboration protocols, including the 14 frameworks in Section \ref{sec:method}.

\vspace*{10pt}
\noindent \textbf{Baselines} We compare the bottom-up and collaborative AI systems with non-modular and non-participatory baselines. \textbf{Since the median year where the contributed models are trained is 2024}, we mainly compare with state-of-the-art open models and systems in that time range for fair comparison. 1) \emph{No modularity, no participation}: single dense LLMs including \textsc{Llama 3.1} (8B, 70B, 405B) and \textsc{Qwen 2.5} (7B, 14B, 32B, 72B); 2) \emph{Yes modularity, no participation}: mixture-of-experts LLMs including \textsc{Mixtral} (8*7B, 8*22B) and \textsc{Deepseek-MoE} (16B, 236B-v2); 3) \emph{No modularity, yes participation}: \textsc{Marin} (8B); we exclude the larger Marin-32B from the main scaling comparison since only its base, not post-trained, checkpoint is released. We separately conduct a dedicated experiment with only base model checkpoints in Appendix \ref{sec:base_to_base}.

As the average size of our model pool is 5.32B, two of these baselines (\textsc{Llama 3.1} 405B and \textsc{Deepseek-MoE} 236B) exceed the combined parameter count of all 32 participating models. We also compare with the best single model, the model in our pool that performs best on each specific task. If scaling up participation yields strong bottom-up, compositional systems that outperform these baselines, our work spearheads a new paradigm of modular, bottom-up, community-sourced, and participatory AI systems.

\section{Results}
\label{sec:results}
\begin{figure}[!t]
\centering
\includegraphics[width=1.0\textwidth]{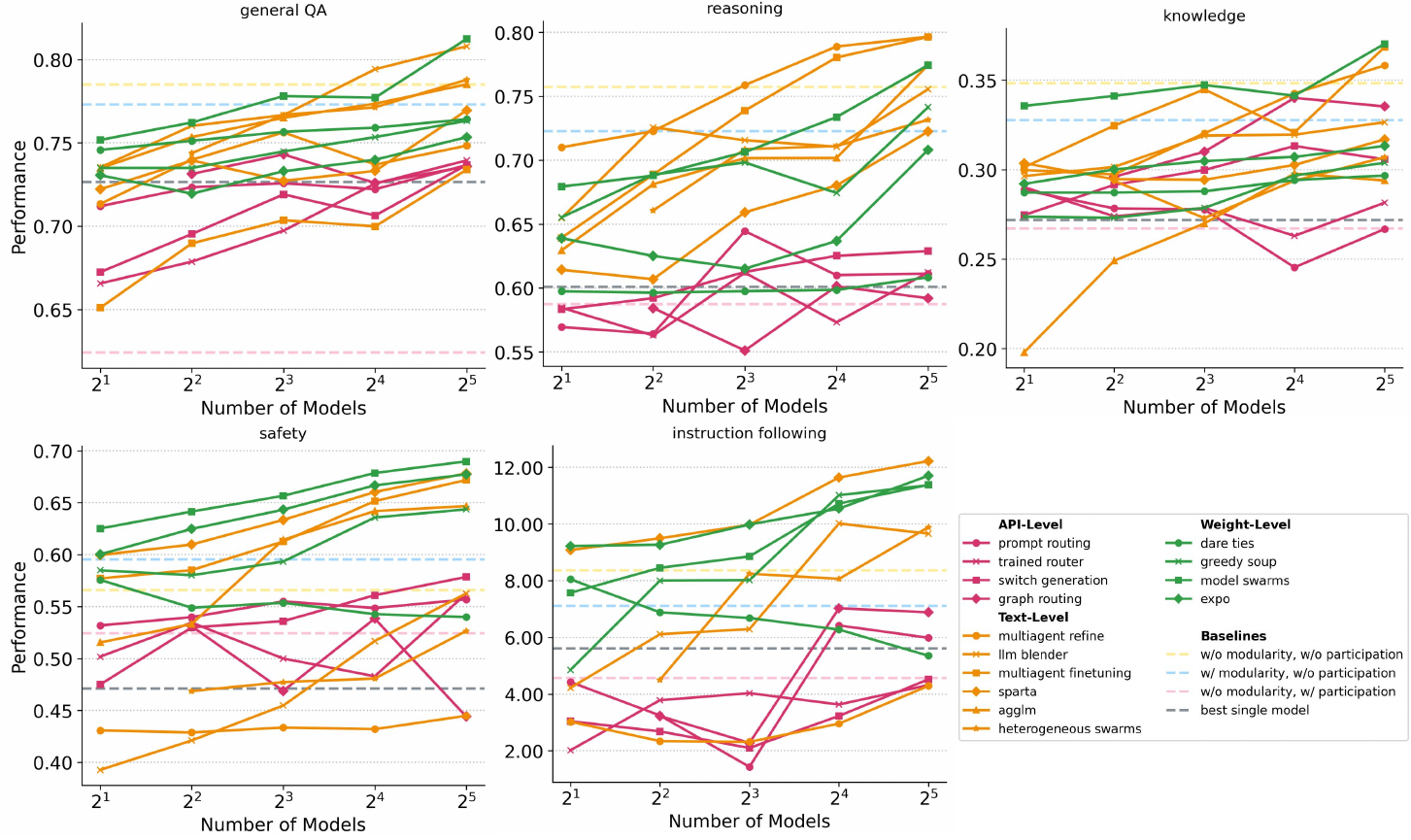}
\caption{\textbf{Results of Scaling Participation.} Performance (macro-averaged per-domain; accuracy for most domains, reward-model score for instruction following, see Appendix \ref{sec:metric_scales} for details) of diverse model collaboration systems and non-modular and/or non-participatory baselines. We report the best performance within each category of baselines as a horizontal dashed line (often the largest model in the category) and the scaling trend of model collaboration systems, from 2 models to 32 models contributed by diverse stakeholders. Graph routing and heterogeneous swarms do not have results for 2 models as 2 models do not constitute a graph network. Many methods generally improve with the scaling of participation in building collaborative AI systems, with the most successful systems outperforming non-modular and/or non-participatory baselines by up to 15.42\% (95\% CI: [10.09\%, 21.13\%]).}
\label{fig:big}
\end{figure}

We report the performance of diverse model collaboration systems and baselines in Figure \ref{fig:big}.

\vspace*{10pt}
\noindent \textbf{Overall Results} By scaling participation through scaling the number of contributed models, diverse model collaboration systems generally trend up by 28.78\% (95\% CI: [22.31\%, 37.86\%]) on average from 2 to 32 models. The most successful collaborative systems outperform all baselines: even though \textsc{Llama 3.1} 405B (in the category of w/o modularity, w/o participation) has 2.38 times the combined parameter count of all 32 models, the best collaborative systems still outperform it by 17.15\% (95\% CI: [11.07\%, 23.67\%]) on average across evaluation domains, and this improvement mostly remains significant after statistical correction (Appendix \ref{sec:multiple_comparisons}). Parameter count is not a proxy for total compute: some collaboration methods exceed the cost of a single \textsc{Llama 3.1} 405B call, while the most cost-efficient ones operate at the inference cost of a single small model (Appendix E). This advantage also holds under compute-matched comparisons against test-time scaling baselines on the dense model (Appendix \ref{sec:compute_match}). Per-domain confidence intervals for the top methods are reported in Appendix E (Table \ref{tab:ci}). This suggests that \emph{scaling participation via scaling model collaboration} is a promising paradigm: bottom-up, community-sourced, compositional AI systems built with the contributions of diverse stakeholders could rival the centralized and monolithic status quo of AI development.

\vspace*{10pt}
\noindent \textbf{Domain Insights} Successful collaborative systems outperform the strongest baseline in each evaluation domain, with gains observed across general QA, reasoning, knowledge, safety, and instruction following. Aside from aggregate-level results, we also observe that there is no one-size-fits-all collaboration system: different types of collaborative systems suit different domains and applications, with concrete takeaways analyzed further below (full per-domain method rankings in Appendix \ref{sec:method_ranking}).

\vspace*{10pt}
\noindent \textbf{Collaborative Systems} API-level methods, featuring the routing and selection among candidate models, are not very scalable, and performance may even degrade when introducing more models: as the model pool grows with participation, the strategy of shallow selection may be susceptible to noise and struggles to find the best individual model. Specifically, prompt routing and trained router both show non-monotonic or degrading trends at large pool sizes (Figure \ref{fig:big}), as a larger candidate set makes accurate routing harder. A controlled leave-one-out study (Appendix \ref{sec:specialist}) shows routing benefits most when a relevant specialist is in the pool; two additional coordination mechanisms, model selection (SLM-Mux) and cost-aware deferral (Cascade), do not fully resolve the degradation either (Appendix \ref{sec:additional_coordination}). In contrast, diverse models are more deeply integrated in text-level and weight-level collaboration, benefiting more from scaling participation, where all models have a chance to contribute their specialization and augment the compositional AI system. Scaling each collaborative system to 32 models, the top-3 ranked methods across evaluation domains (full per-domain rankings in Appendix \ref{sec:method_ranking}) all involve \emph{training} the parameters of participating models: \emph{model swarms} \cite{feng2025model}, \emph{sparta alignment} \cite{jiangsparta}, and \emph{multiagent finetuning} \cite{subramaniammultiagent}. Deeper integration of decentralized contributions, where models collectively evolve and adapt, outperforms methods that merely route or prompt models without updating their parameters. This suggests that the current emphasis on prompting and agentic orchestration of frozen models may not be sufficient for maximum collaborative utility, and that training-based participation is a promising direction for realizing more of what diverse contributions have to offer. It boils down to a single best practice for anyone building bottom-up and modular AI systems: \textbf{curate a diverse and specialized model pool, and collaborate through methods with deeper levels of integration and adaptation.}
\section{Analysis}

\vspace*{10pt}
\noindent \textbf{Evaluations on Reflecting Human Diversity} In addition to excelling on general-purpose tasks (Section \ref{sec:results}), we posit that community-sourced, participatory AI systems should be particularly strong at \emph{representing human diversity}: as more models contributed by diverse stakeholders in AI join the system, the values, preferences, and priorities of them will be better reflected than monolithic models. We investigate this by evaluating participatory AI systems and the two largest baseline models on tasks about human diversity: CulturalBench \cite{chiu2025culturalbench}, testing cultural knowledge with questions constructed via human-AI red-teaming: with more than one pluralistically correct answers, success cannot come from mode maxing alone; Value Kaleidoscope \cite{sorensen2024value}, testing whether a model identifies the values relevant to a situation and correctly judges whether they are supported or opposed, treating value pluralism itself as the target rather than one universally correct value; and InfinityChat \cite{jiangartificial}, open-ended queries with no single ground truth, scored via reward-model quality and inter-model response diversity (full descriptions, success criteria, and benchmark-specific limitations in Appendix \ref{sec:details}). Together, these evaluations quantify how well models and systems represent diverse cultures, values, and employ those knowledge in real-world user queries.

\begin{figure}[!t]
\centering
\includegraphics[width=1.0\textwidth]{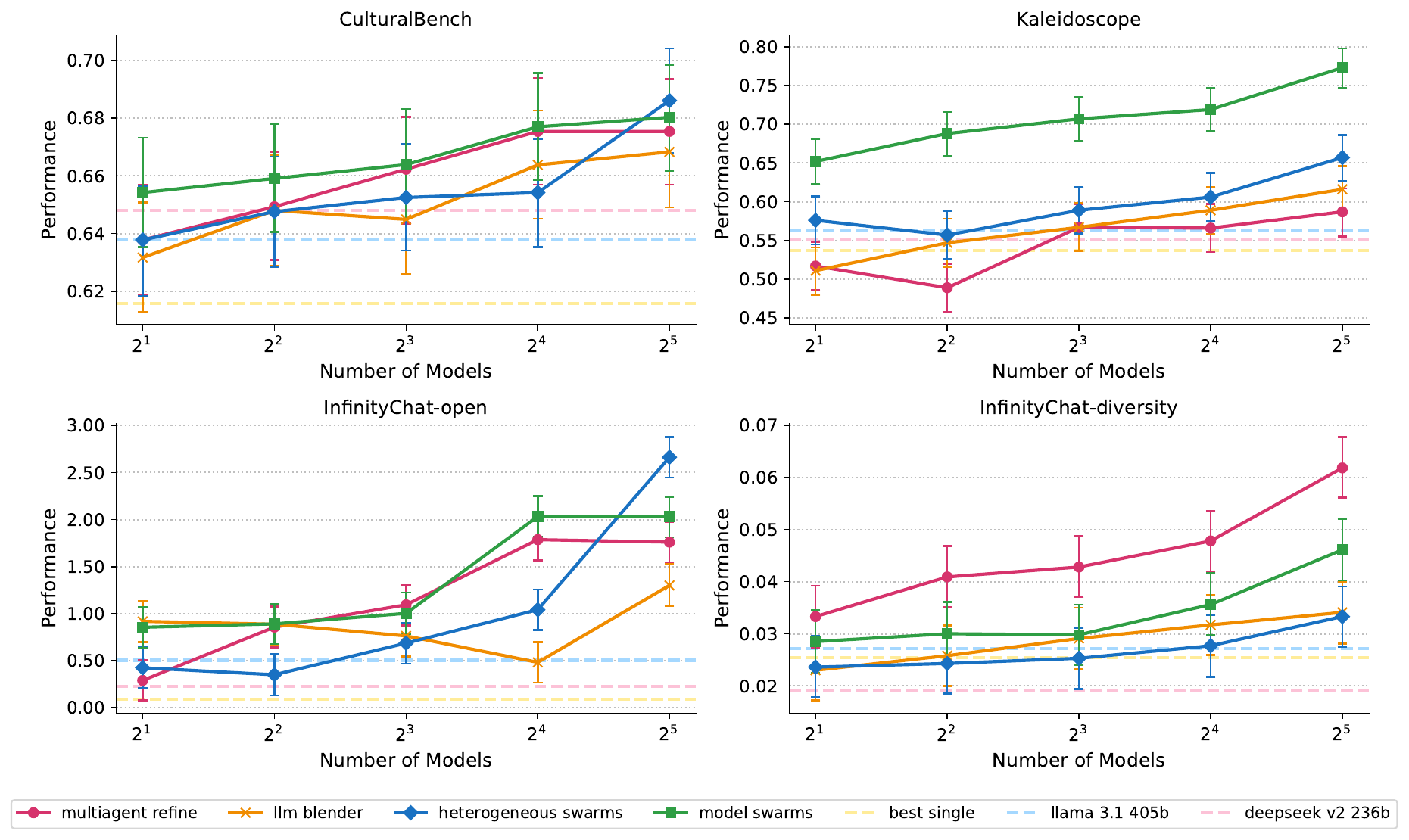}
\caption{\textbf{Evaluations on representing human diversity}: representing diverse culture, values, and employing them for real-world human queries. Metrics differ by dataset (accuracy for CulturalBench and Kaleidoscope, reward-model score for the open setting of InfinityChat, and inter-model diversity metric for the diversity setting; see Appendix \ref{sec:metric_scales}); the improvement over monolithic models reaches 21.58\% here against 17.15\% on general-utility tasks in Figure \ref{fig:big}. Error bars represent 95\% confidence intervals for the collaboration methods; baseline CIs are reported in Table \ref{tab:diversity_ci} in the Appendix.}
\label{fig:diversity_evals}
\end{figure}

Results in Figure \ref{fig:diversity_evals} demonstrate that: 1) scaling participation continues to yield promising trends on these tasks, improving from 2 to 32 models; 2) the improvements over large monolithic models reach 21.58\%, compared with 17.15\% on general-purpose tasks (Section \ref{sec:results}). This shows that incorporating diverse contributions from decentralized stakeholders yields systems that perform better on tasks probing diverse cultures, values, and preferences. To confirm that these gains do not reflect majority-class prediction, we compute majority baselines for the two objectively scored datasets: 50.24\% on CulturalBench-Hard and 50.50\% on Value Kaleidoscope. All collaborative systems exceed both, and the margin is largest for the strongest systems: on Value Kaleidoscope model swarms reaches 77.30\% against a 50.50\% majority baseline, while the monolithic baselines reach only 53.70\% to 56.30\% and therefore sit substantially closer to majority prediction than the participatory systems do. A majority baseline is not applicable to InfinityChat, which is open-ended generation; its diversity setting is instead scored by inter-model homogeneity, a measure of generation diversity rather than of correctness, so improvement there cannot be obtained by mode-seeking behavior.

\vspace*{10pt}
\noindent \textbf{Scaling Model Diversity} Scaling participation via scaling model collaboration led to consistent improvements with larger and more diverse AI systems (Figure \ref{fig:big}). Are the improvements merely due to spending more compute on the collaboration of more models, or are they also due to the increasing diversity and complementary strengths of contributed models? To disentangle the effect of compute and contribution diversity, we conduct a controlled experiment where collaborative systems are always built with 8 models, but models varying in different prongs of diversity. We follow the model ranking order in Section \ref{sec:experiment_settings} to select eligible models.

\begin{itemize}[leftmargin=*]
    \item \textbf{Setting 1: Distinct Models} We employ settings of $a \times b$ where $a$ distinct models are each repeated for $b$ times in the pool. ($a \times b = 8$) Specifically, four model pools of $1 \times 8$, $2 \times 4$, $4 \times 2$, and $8 \times 1$ offer an increasing scale of diversity. \vspace{5pt}
    \item \textbf{Setting 2: Diverse Institutions} Among the pool of 61 models, researchers from the University of Washington (UW) contributed the most. We investigate whether a compositional AI system built with cross-institution collaboration would be helpful. Specifically, we employ a pool of 8 UW models, 6 UW and 2 from other institutions, 4 and 4, as well as 1 from each institution: these pools offer a progressive scaling of multi-institution collaboration.
    \item \textbf{Setting 3: International Collaboration} While most of the models were contributed by researchers in the United States, there are also models from international scholars. We investigate the case of international collaboration, with 8 US models, 6 US and 2 else, 4 and 4, as well as 2 and 6. We hope to understand whether the diverse research agenda and priorities around the globe would strengthen collaborative AI systems.
    \item \textbf{Setting 4: Generation Diversity} An important measurement of model diversity is the diversity of their generated texts: we employ the inter-model homogeneity with the InfinityChat dataset \cite{jiangartificial}, a repository of open-ended user questions, to evaluate the diversity of model pools. We draw random samples of 8 models from the whole pool 4 times, rank their diversity level based on inter-model homogeneity (quantitative measurement and raw statistics in Appendix \ref{sec:metric_scales}), and build collaborative systems with those models.
\end{itemize}

\begin{figure}[!t]
\centering
\includegraphics[width=1.0\textwidth]{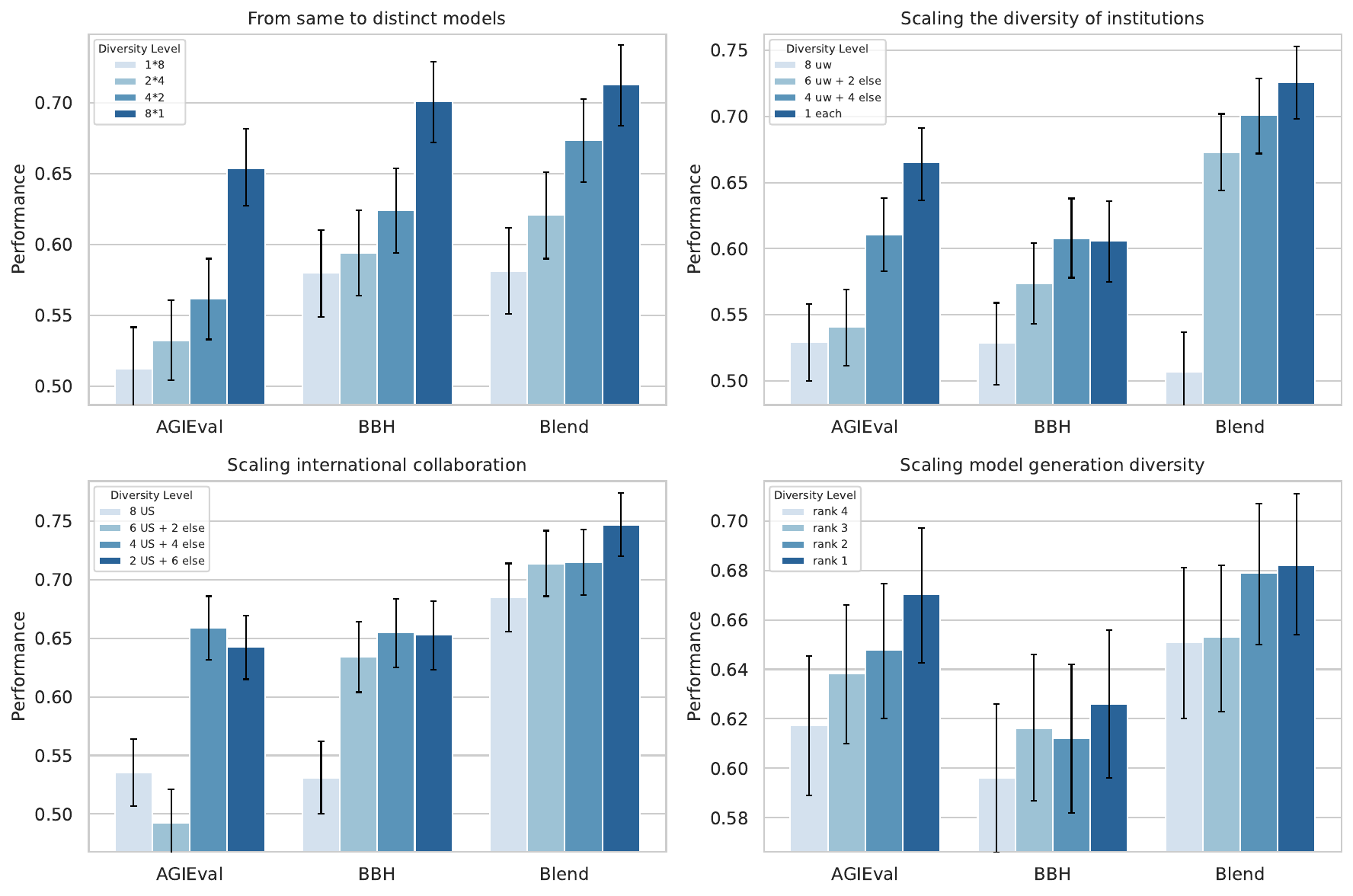}
\caption{\textbf{Scaling Model Diversity.} We investigate the impact of scaling model diversity by keeping the model pool size as eight, but employing increasingly diverse model compositions across four aspects. Performance is the same macro-averaged per-domain metric as Figure \ref{fig:big} (details in Appendix \ref{sec:metric_scales}). An average improvement of 13.8\% from the least diverse to most diverse settings indicates that the successes of scaling participation are not only due to compute, but also crucially due to the diversity and complementary strengths of models contributed by diverse stakeholders. Error bars represent 95\% confidence intervals.}
\label{fig:model_diversity}
\end{figure}

We employ multi-agent refine, multi-agent finetuning, heterogeneous swarms, and model swarms as collaborative protocols to build AI systems with each model pool. We report performance for each model pool averaged across the four collaboration strategies in Figure \ref{fig:model_diversity}. From the least to most diverse settings, we observe an average improvement of 13.8\%, indicating that model diversity is crucial in the success of scaling participation. Across the four settings, the average improvement is 18.8\% (setting 3), 18.0\% (setting 2), 14.5\% (setting 1), and 3.8\% (setting 4), in descending order, indicating that diverse sources of contributed models are critical in our success. Together, these results demonstrate that through scaling participation, diverse preferences and priorities about AI models are holistically reflected to build stronger AI systems. This pattern echoes the classical wisdom-of-crowds effect \cite{galton1907vox} and the Condorcet Jury Theorem's result that aggregating independent, better-than-chance judgments improves collective accuracy \cite{austen1996information}, and is consistent with the ensemble learning literature's diversity-accuracy relationship \cite{kuncheva2003measures}.

\begin{figure}[!t]
\centering
\includegraphics[width=1.0\textwidth]{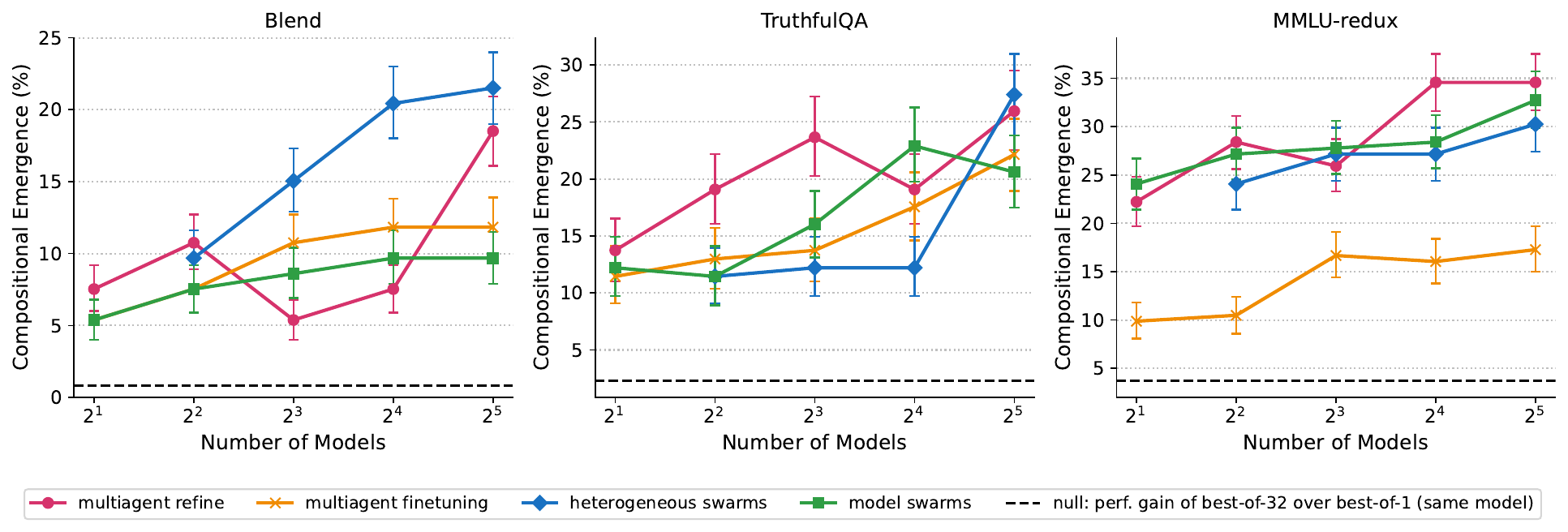}
\caption{\textbf{Scaling collaborative emergence}: for problems where all participating models could not individually solve, what percentage of them are now solvable with collaborative systems of varying scales? The black dashed line is the null baseline: the performance improvement from best-of-32 over best-of-1 with the best single model (0.84\%, 2.27\%, 3.69\% across datasets). All collaborative methods substantially exceed it, so the gains cannot be explained by repeated sampling alone. Error bars represent 95\% confidence intervals.}
\label{fig:collaborative_emergence}
\end{figure}

\vspace*{10pt}
\noindent \textbf{Collaborative Emergence} In addition to what models are individually capable of, collaborative AI systems of multiple models could solve problems where individual models struggle, \emph{i.e.,} collaborative emergence. \cite{feng2026moco} This is due to collaborative systems tapping into the complementary strengths of diverse models and generalizing beyond their existing capabilities. To investigate whether collaborative emergence scales with participation, we calculate the percentage of problems that 1) all 32 models in the pool could not individually solve, and 2) collaborative systems of 2 to 32 models could now solve. Figure \ref{fig:collaborative_emergence} demonstrates that collaborative emergence is happening with participatory AI systems, and such emergence and generalization further scale with participation. To establish a null baseline, we compute the percentage of these problems solvable by best-of-32 sampling from the strongest single model (0.84\%, 2.27\%, 3.69\% across datasets) compared to best-of-1, shown as the black dashed line. All collaborative methods substantially exceed it, so the gains cannot be explained by repeated sampling alone. This highlights a new possibility: that generalization does not need to come from excessive scaling of a single model, but from scaling the collaboration and representation of diverse contributed models.

\vspace*{10pt}
\begin{figure}[!t]
\centering
\includegraphics[width=1.0\textwidth]{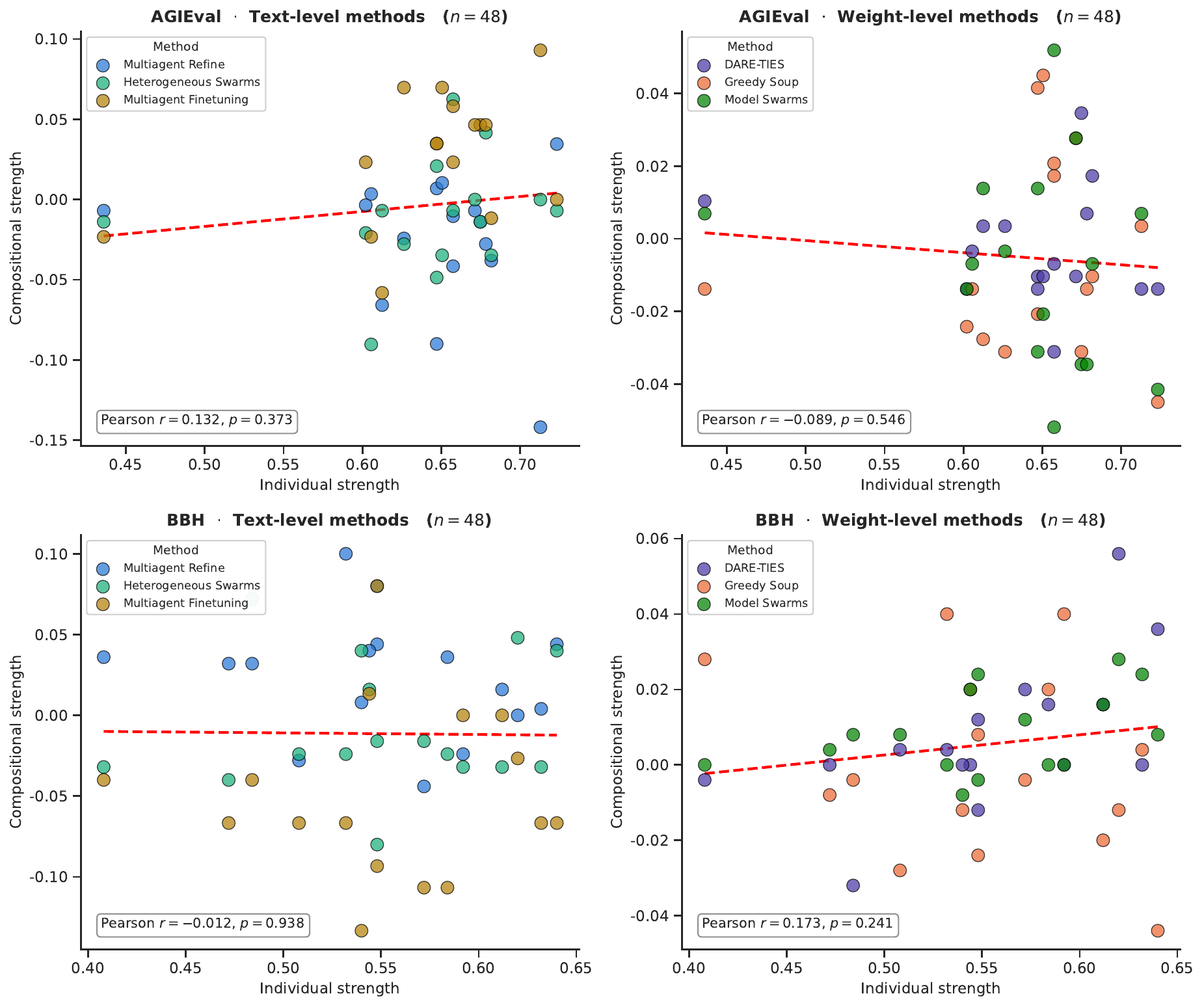}
\caption{\textbf{Individual and Compositional Strength.} Correlation between the individual and compositional strength of LLMs across six collaboration methods (three text-level, three weight-level) and two datasets (48 model-method pairs per subfigure). Across the evaluated settings we do not detect a consistent association between individual and compositional strength (Pearson r $< 0.18$ in all settings). This indicates that evaluating and training AI models for compositional strength is an independent and critical research question. Appendix \ref{sec:fig6_clustering}'s cluster bootstrap and mixed-effects analysis both account for models repeating across methods and confirm the lack of association.}
\label{fig:compositional_strength}
\end{figure}

\noindent \textbf{Individual vs. Compositional Strength} Current evaluations of AI models are mostly measuring \emph{individual strength}: what models can and can not do when operating individually. However, with the emerging paradigm of collaborative and participatory AI, models are no longer merely relying on their own capabilities: it is time to measure and evaluate \emph{compositional strength}, to what extent models are helpful when being employed in compositional systems. We take the first step towards this research question.

Given a pool of $n$ language models $\mathcal{M} = \{\mathbf{m}_1, \cdots, \mathbf{m}_n\}$ and a dataset $\mathcal{D}$, we quantify the individual and compositional strength of each model as follows:

\begin{itemize}[leftmargin=*]
    \item \textbf{Individual Strength}: performance of model $\mathbf{m}_i$ on dataset $\mathcal{D}$, where $\mathbf{m}_i$ individually tackles the problems in $\mathcal{D}$.
    \item \textbf{Compositional Strength}: performance gain when model $\mathbf{m}_i$ is added to a collaborative system. Specifically, $\mathop{\mathbb{E}}_{c \thicksim \mathcal{C}} \Bigl[ \mathrm{performance} \bigl( c (\mathcal{M} ) \mid \mathcal{D} \bigr) - \mathrm{performance}\bigl( c(\mathcal{M} \backslash \mathbf{m}_i) \mid \mathcal{D} \bigr) \Bigr]$, where $\mathcal{C}$ indicates a set of possible collaboration algorithms.
\end{itemize}

We use the top-16 models as $\mathcal{M}$ and investigate the correlation between individual and compositional strength with six collaboration algorithms across two levels (text-level: Multiagent Refine, Heterogeneous Swarms, Multiagent Finetuning; weight-level: Dare-Ties, Greedy Soup, Model Swarms) and two datasets (AGIEval, BBH), yielding 48 model-method pairs per subfigure. We do not detect a consistent association between individual and compositional strength (Pearson $r < 0.18$ in all settings), shown in Figure \ref{fig:compositional_strength}. While existing AI progress is razor-focused on evaluating and improving the \emph{individual} strength of AI models, it is not guaranteed that they will be strong and helpful in future collaborative and participatory systems. It is thus time to go beyond the status quo of optimizing individual strength to beat other models, but for research on evaluating and training AI models by compositional strength.

\vspace*{10pt}
\begin{figure}[!t]
\centering
\includegraphics[width=1.0\textwidth]{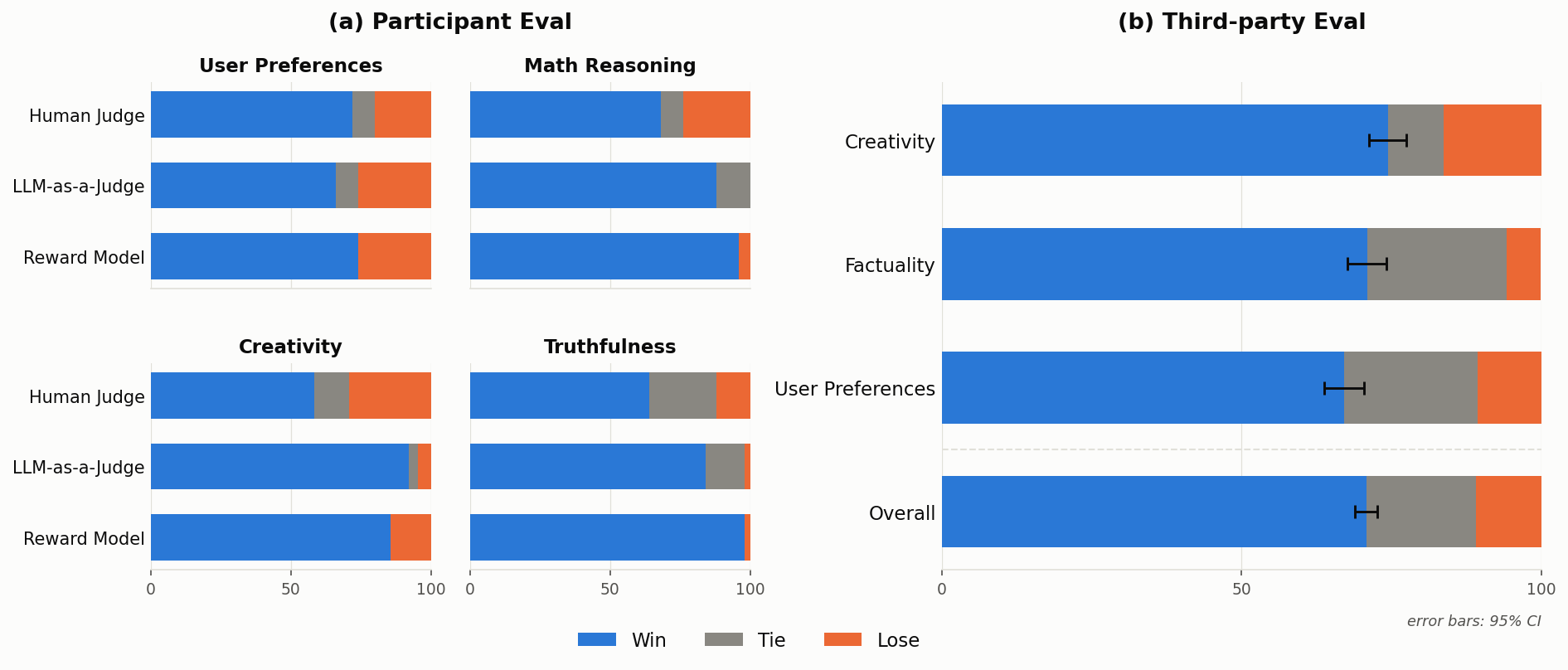}
\caption{\textbf{Evaluation on Participants' Domains.} (a) Participant evaluation: contributors compare their individual model against the compositional system on their own priority domains, using Human Judge, LLM-as-a-Judge, and Reward Model. (b) Third-party evaluation: blind Prolific-based evaluation across three domains (Creativity, Factuality, User Preferences) by evaluators with no connection to the contributed models. Panel (b) aggregates 2250 judgments from 30 annotators over 75 response pairs with response order randomized. The compositional system wins substantially in both evaluation settings, with 95\% CI error bars confirming statistical significance in the third-party evaluation.}
\label{fig:human_eval}
\end{figure}

\noindent \textbf{Getting Back to Participants} Participatory AI will only be possible if diverse stakeholders contribute what they can to build bottom-up AI systems. We posit that one potential incentive for contributions is that the compositional systems they took part in developing will perform better on their original priorities that motivated them to develop their models, thereby advancing their objective at the same time. We invite contributors to an optional human evaluation, where we ask for the data they used to evaluate their individual models and conduct pairwise comparisons between responses from their model and the participatory system, using three evaluation protocols detailed in Appendix \ref{sec:details}: Human (Participant) Judge, LLM-as-a-Judge, and Reward Model.

Results in Figure \ref{fig:human_eval}(a) demonstrate that the participatory and compositional AI systems advance on objectives that participants were originally pursuing. To address potential evaluator bias from contributors judging a system containing their own model, we additionally conduct a blind third-party evaluation on Prolific \cite{prolific}, recruiting evaluators with no connection to the contributed models. Thirty annotators each judged 75 response pairs presented as Response A and Response B with position randomized, yielding 2,250 blind judgments; annotators were told only that both responses were AI-generated, with no information about system identity or provenance. Recruitment, compensation, quality control, and annotator demographics are given in Appendix \ref{sec:details}. We cover three of the four domains: Math Reasoning is excluded from the third-party evaluation as it requires specialized expertise that general crowdworkers cannot reliably assess. Results in Figure \ref{fig:human_eval}(b) show that across 2,250 judgments the compositional system is preferred in 70.8\% of comparisons, against 18.3\% ties and 10.9\% losses, and is preferred in each domain: Creativity (74.4\%), Factuality (70.9\%), and User Preferences (67.1\%), with 95\% CI error bars confirming statistical significance. Annotators were also largely consistent: 68\% of items reached a clear majority verdict, with more than half of annotators agreeing on the same label. Together, both evaluations confirm that contributing to a participatory system benefits participants by developing systems better equipped to address their needs and priorities.

\section{Discussion}

\vspace*{10pt}
\noindent \textbf{Safety of Participatory AI Systems} In this work, we prototype participatory AI with bottom-up, community-sourced, collaborative AI systems built with the contributions of global academics. To bring this paradigm to the wild, we need to account for the safety risks of decentralized contributions. For example, what if some of the contributed models are intentionally or unintentionally biased or malicious? Testing this directly with real adversarially-trained checkpoints \cite{yang2026among}, we find selection-based collaboration methods can be captured: prompt routing's accuracy drops by up to 73\% relative when 2 of 8 pool models are malicious, since the malicious checkpoint is routed to on the large majority of queries, while methods that aggregate across the pool show no comparable degradation (Appendix \ref{sec:malicious}). A related concern is disproportionate rather than malicious contribution: a single actor contributing many copies of one model can fully capture selection-based collaboration methods, though deeper-integration methods resist this better (Appendix \ref{sec:bloc}). We need future research on designing guardrails and ensuring the safety of bottom-up, participatory AI systems \cite{yang2026among}, including the dual-use risk of decentralized contributions (Appendix \ref{sec:dual_use}). Our evaluations are also benchmark-based rather than deployed in high-stakes, real-world, or compute-constrained settings; validating the paradigm there is future work.

\vspace*{10pt}
\noindent \textbf{Why and When Scaling Participation Works} Across our experiments, the benefits and limits of scaling participation trace back to how deeply a collaboration method integrates contributions and how well a pool's diversity matches the task at hand (Appendix \ref{sec:why_succeed}). Selection-based methods, such as prompt routing and trained routing, succeed by identifying a single best contributor per query, so they benefit most when a relevant specialist is in the pool (Appendix \ref{sec:specialist}); the same mechanism, however, is why they degrade as the pool grows, since accurate selection becomes harder as the candidate set expands (Section~\ref{sec:results}). Methods that instead let contributed models adapt to and integrate with one another, whether through iterative refinement, joint training, or parameter merging, alleviate this failure mode, though the best method varies by task and domain (Appendix \ref{sec:method_ranking}). Because a model's standalone strength does not predict its compositional contribution (Section~\ref{sec:results}), curating a pool by benchmark performance alone is not sufficient; compositional strength should become an explicit selection criterion alongside safety-based screening \cite{yang2026among}. Practically, these results suggest scaling participation should default to deeper, adaptive integration methods as the pool grows, reserve selection-based routing for smaller pools with a known relevant specialist, and evaluate prospective contributors on compositional rather than standalone strength.

\vspace*{10pt}
\noindent \textbf{Facilitate Contributions} We propose to scale participation via scaling model collaboration, since model sharing is the default open-source activity in AI. However, \emph{not everyone knows how to train a model}, especially for those who are underrepresented in the AI status quo. We need future research on facilitating diverse contributions from populations with less AI literacy, potentially through natural language-based interfaces for annotation, feedback, and other user-friendly forms of contribution, such as tools that turn natural-language feedback directly into a deployable specialized model \cite{viswanathan2023prompt2model, zhao2024selfguide}.

\vspace*{10pt}
\noindent \textbf{Scope of the Current Prototype} This work is a proof-of-concept of scaling participation, not its full realization. The 61 contributed models are all from academic research labs, and ``participation'' as operationalized here means consenting to include an already-released model. The pool is also concentrated in a few base-model families (Figure \ref{fig:pool_diversity}D), so much of its diversity lies in research objectives, training methods, and post-training recipes rather than in pretraining distributions. This reflects the current open-source landscape, where only a handful of high-quality open foundation models are widely available, and we expect the pool to diversify as more are released. Testing whether this concentration collapses the pool into a hive mind (\cite{jiangartificial}, Appendix \ref{sec:hivemind}) finds the opposite: across three independent open-ended datasets, different models disagree with each other more than a single model disagrees with itself under repeated sampling; Appendix \ref{sec:hivemind} also discusses how to actively combat this risk as the pool scales further. Broader participation from non-academic stakeholders, underrepresented communities, low-resource language speakers, and non-English speakers remains a critical goal, and our diversity analysis (Figure \ref{fig:model_diversity}) suggests such contributions would strengthen participatory systems further rather than saturate their benefits.

\vspace*{10pt}
\noindent \textbf{Toward Collective Decision Making} Participation creates a basis for shared ownership: because diverse stakeholders contributed the modular components, they have a legitimate claim on how the resulting system is used. We emphasize that our work does not implement collective governance. The systems we evaluate were assembled and operated centrally, by us. What the paradigm does provide is a technical foundation on which governance could be built, drawing on existing participatory frameworks for algorithmic governance. For example, WeBuildAI \cite{lee2019webuildai} has each stakeholder's component rank the decision alternatives, then aggregates these rankings into one collective ranking via the Borda rule \cite{brandt2016handbook}, a positional voting method; the natural analog in our paradigm is to aggregate each contributed model's ranking or choice over a decision's alternatives. This is one possible governance structure the paradigm could support, among others \cite{ter2025democratizing, aziz2017justified, gordon2022jury}. Because contributions remain separate, identifiable artifacts rather than being irreversibly absorbed into one monolithic model, contributors retain meaningful leverage: a contributor can decline participation, condition it on particular use cases, or withdraw their model from the pool, and the system can be re-composed without them. Attribution over modular components similarly makes it possible to weight decisions by contribution. Turning these concepts into actual procedures, covering who may contribute, deploy, or decide permitted use cases, is essential future work rather than something our work delivers.

\vspace*{10pt}
\noindent \textbf{Economic Incentives} We envision important socio-technical research on designing economic incentives to facilitate the transition to participatory AI systems. For example, if we have fair attribution algorithms, such as a model's Shapley value \cite{shapley1953value} or other game-theoretic solution concepts, that allocate a system's outcome to each contributor's marginal impact, the proceeds of participatory AI systems could be proportionally distributed to those contributors. This will in turn draw more effort towards the participatory paradigm, turbocharging the transition from unilateral and monolithic models to bottom-up and compositional systems.

\section{Conclusion}
We propose to scale participation in modular AI, where community-sourced AI systems are built with bottom-up contributions from diverse stakeholders. We prototype participatory AI by scaling model collaboration, where up to 32 models trained by different researchers collaborate as compositional AI systems. Participatory AI outperforms monolithic baselines by up to 15.42\% (95\% CI: [10.09\%, 21.13\%]) across diverse tasks, especially on representing human diversity. These bottom-up compositional systems benefit from the diversity of model contributions, not merely compute or model quality, and new skills emerge to solve problems individual models struggle with. Scaling participation offers a proof of concept that modular AI systems built from decentralized contributions can rival centralized, monolithic development; further scaling remains important future work toward an open, collaborative AI future.


\bibliography{sn-bibliography}

\appendix

\newpage


\section*{Funding Statement}
This research was developed in part with funding from the Defense Advanced Research Projects Agency's (DARPA) SciFy program (Agreement No. HR00112520300). The views expressed are those of the author and do not reflect the official policy or position of the Department of Defense or the U.S.~Government. This research was supported by the Coefficient Giving and Amazon Health. Shangbin Feng would like to thank the support of the IBM PhD Fellowship, the Jane Street Graduate Research Fellowship, and the NVIDIA Graduate Fellowship.

\section*{Author Contributions}
S.F. led the project. Conception and design: every author. Soliciting participation: Y.W. and S.F. Result and evaluation: S.F. Result visualization: Y.W. and S.F. Advisory: L.Z., Y.C., and Y.T. All authors contributed to manuscript writing and editing.

\section*{Competing interests}
The authors declare no competing interests.

\section{Related Work}
\label{sec:related_work}

\textbf{Democratizing AI} \ \ Research to democratize AI spans the full lifecycle of AI models, from model training, inference, deployment, and its societal impacts, with four specific sub-areas \citeapp{seger2023democratising, subramonian2024understanding, collective2024}: democratizing AI governance, which focuses on building systems and infrastructures to ensure the development and usage of AI systems reflect the preferences of the general public. This line of research focuses on building socio-technical systems \citeapp{laux2024three, ovadya2025position} to encourage public participation \citeapp{ter2025democratizing}, rein in unilateral control \citeapp{roose2022generative}, and ensure the justice and fairness of AI-related decision making \citeapp{kalluri2020don}; democratizing AI development, which focuses on collaborative and decentralized development of AI models. This line of research focuses on modular models and systems \citeapp{shiflexolmo, zhou2025democratizing}, the pluralistic alignment of AI models \citeapp{sorensenposition, steingruber2025justifications, huang2025democratizing}, and protocols of collaborative development \citeapp{kandpal2023git, borzunov2023petals, marin2025}. There are additional discussions about the democratization of AI usage \citeapp{yu2023ai} and profits \citeapp{miller2021radical}. Our work focuses on the democratization of AI development: existing proposals for collaborative development mostly feature new initiatives that require non-trivial effort for participation \citeapp{kandpal2023git, marin2025} (e.g., moving to a new experimental framework, making major changes to existing code and training procedures), while our proposal of \emph{scaling participation in modular AI systems} is uniquely compatible with the status quo: people, academics, and communities around the world have been training language models that represent their priorities and sharing them with the world\footnote{As of August 5, 2026, there are 2,969,238 open models on Huggingface \citeapp{wolf-etal-2020-transformers}, a popular model sharing infrastructure.}. With a bottom-up and compositional AI system featuring the collaboration of these diverse language models contributed by different people, we could efficiently reuse existing progress in AI and pose a low barrier for participation from stakeholders who already train and release open models.

\vspace*{10pt}
\noindent \textbf{Model Collaboration} Model collaboration is a family of algorithms towards building AI systems through compositional generative modeling, where multiple machine learning models collaborate, compose, and complement each other through diverse collaboration protocols \citeapp{feng2025one, feng2026moco}. The collaboration of diverse large language models could be categorized into four levels, based on where and how information exchange happens across models: \emph{API-level collaboration} features the routing \citeapp{ongroutellm, fenggraphrouter} and cascading \citeapp{dohan2022language, chenfrugalgpt} among multiple LLMs, aiming to select the most fitting and efficiency LM for diverse user queries; \emph{Text-level collaboration} features collaborative problem solving \citeapp{guo2024large}, competition and debate \citeapp{du2023improving, subramaniammultiagent}, response aggregation \citeapp{jiang2023llm, zhao2025majority}, and other paradigms for models to communicate via generated texts; \emph{Logit-level collaboration} employs arithmetic on next-token probability distributions for collaborative decoding with multiple language models \citeapp{liutuning, shen2024learning}; \emph{Weight-level collaboration} features model merging \citeapp{yadavsurvey, he2024localize, tian2026mergemix}, training and merging heterogeneous expert LLMs \citeapp{li2022branch, sukhbaatarbranch}, routing and composition of specialized expert modules \citeapp{muqeeth2024learning, huanglorahub}, arithmetic with model parameters \citeapp{zheng2024weak, feng2025model}, and modular architectures \citeapp{shiflexolmo}. While these algorithms are previously validated with centrally assembled model sets, we bring them to the wild: diverse language models contributed by decentralized stakeholders collaborate to form bottom-up compositional AI systems, turbocharging their potential to build an open and participatory AI future.

\section{Comparison with Prior Paradigms}
\label{sec:comparison_table}

Table \ref{tab:paradigm_comparison} compares scaling participation against closely related prior paradigms across six dimensions.

\begin{sidewaystable}
\centering
\small
\caption{Comparison of AI development paradigms across six dimensions. N/A in columns (d)--(f) indicates the paradigm involves a single organization, so training data sharing, participation barriers, and architecture constraints are not applicable.}
\label{tab:paradigm_comparison}
\begin{tabularx}{\textheight}{
  >{\raggedright\arraybackslash}p{2.8cm}
  >{\raggedright\arraybackslash}X
  >{\raggedright\arraybackslash}X
  >{\raggedright\arraybackslash}X
  >{\raggedright\arraybackslash}X
  >{\raggedright\arraybackslash}X
  >{\raggedright\arraybackslash}X
}
\toprule
\textbf{Paradigm} & \textbf{(a) Where modularity lives} & \textbf{(b) Who contributes what} & \textbf{(c) Compute profile} & \textbf{(d) Training data required?} & \textbf{(e) Barrier to participation} & \textbf{(f) Architecture constraint?} \\
\midrule
Single dense LLM & None; monolithic & Single org trains everything & One-time training; dense inference & N/A & N/A & N/A \\
\addlinespace
MoE LLMs & Architecture: sparse expert routing, jointly trained & Single org trains all experts & One-time training; sparse inference & N/A & N/A & N/A \\
\addlinespace
Collaborative training (e.g., Marin) & None; contributions merged into one model & Multiple orgs contribute data and compute to joint pipeline & Collaborative training; single model inference & Yes & Must join framework, change workflow & Yes \\
\addlinespace
FlexOLMo \citeapp{shiflexolmo} & Architecture: MoE expert modules via branch-train-merge & Multiple orgs contribute domain-specific expert modules & Independent training of experts; sparse expert inference & Not necessarily & Must follow MoE expert framework & Yes (must use MoE architecture) \\
\addlinespace
MoA / Multi-agent (in-house) & Inference-time ensemble or multi-round exchange (debate, routing, aggregation) & A single org selects and orchestrates all models & $k \times$ inference per query, sometimes over multiple rounds; no extra training & No & N/A & No \\
\addlinespace
Model Fusion (in-house) & Weight-space: parameters merged into a single model & A single org trains or acquires all models being merged & One-time merge; single-model inference thereafter & No & N/A & Yes (merged models must share compatible architecture) \\
\addlinespace
\textbf{Ours: Scaling Participation} & Inference + optional post-training of decentralized contributed models & Diverse independent stakeholders contribute their own trained models & Marginal over existing models; parallelizable at inference & No (only trained model required) & N/A, only need consent to include an existing open model & No (any architecture; distillation bridges weight-level methods) \\
\bottomrule
\end{tabularx}
\end{sidewaystable}

\noindent The comparison above also highlights a practical advantage: scaling participation is compatible with the status quo in ways the alternatives are not. Existing collaborative development protocols such as Git-theta \citeapp{kandpal2023git} and Marin \citeapp{marin2025} often require extra effort to engage, where participants need to change their existing workflow, coordinate training procedures, and share private training data with a centralized pipeline. Scaling participation requires none of this: contributors train entirely independently and the only effort needed is to give consent to use their model. This also means the 2.9 million models already on Huggingface \citeapp{wolf-etal-2020-transformers} can retroactively participate without any new effort. Beyond the lower barrier, scaling participation also offers a qualitative advantage: while Marin-style collaborative training permanently absorbs contributions into a single merged model, scaling participation preserves each contributed model as an independent artifact, maintaining the heterogeneity that our diversity experiments show is critical to performance (13.8\% gain). Because contributions enter as separate, identifiable artifacts rather than being absorbed into a single training run, what each contributor brought remains possible to attribute and disentangle, which monolithic collaborative training does not support. One could also combine the two, using Marin-style collaborative training to produce several models and ensembling them at inference. Such a system would recover multi-model inference and some heterogeneity, but the unit of contribution would remain a coordinated pipeline that participants must join in advance, conform to, and operationalize their data within. Scaling participation instead takes independently trained models as they already exist, which is what allows participation to be retroactive and open-ended.

\begin{figure}[t]
\centering
\includegraphics[width=0.9\textwidth]{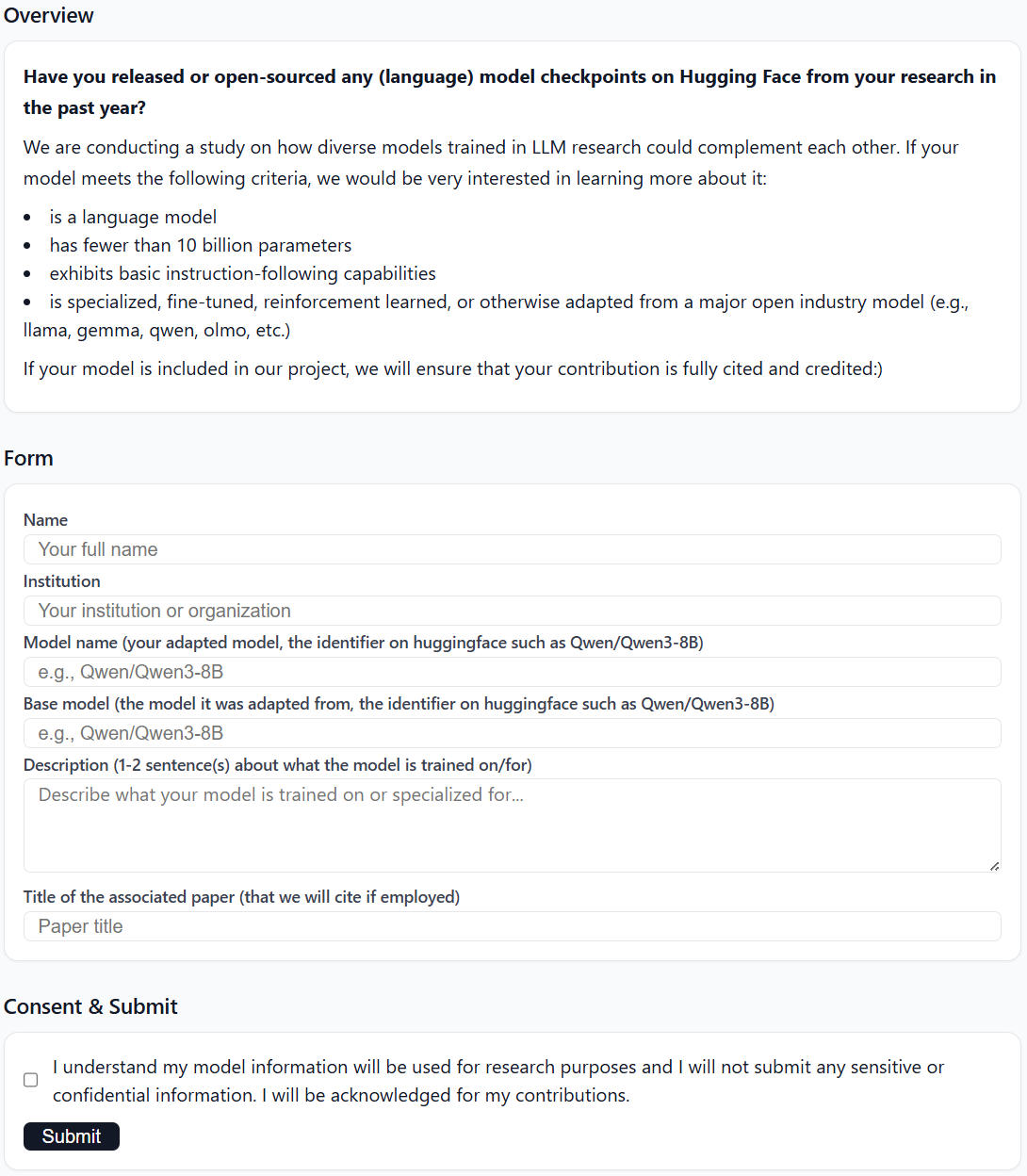}
\caption{\textbf{Soliciting Participation.} The form we used to solicit participation from academics around the world, asking for the language model they trained and released for diverse topics about AI. We are continuing to collect responses at \url{https://yikee.github.io/open-model-collaboration/}.}
\label{fig:solicit_parti}
\end{figure}

\section{List of Participation}

\begin{figure}[t]
\centering
\includegraphics[width=1.0\textwidth]{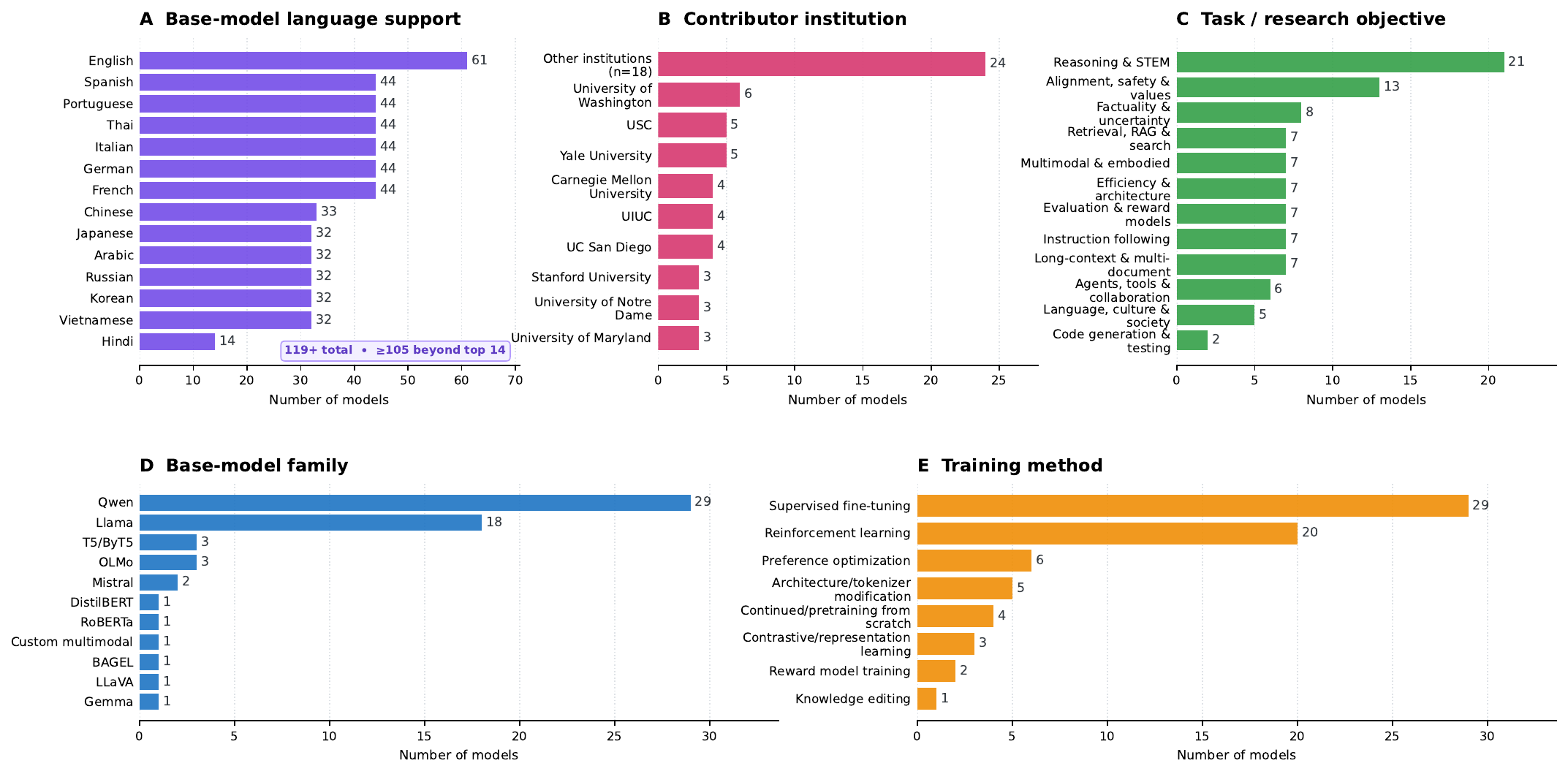}
\caption{\textbf{Diversity composition of the 61 contributed models} across five dimensions: (A) model language support, (B) contributor institution, (C) task/research objective, (D) base-model family, and (E) training method. The pool spans 10 named institutions plus 18 others, diverse languages and dialects, 12 research objectives, 11 base model families, and 8 types of training methods.}
\label{fig:pool_diversity}
\end{figure}

By sending out questionnaires (Figure \ref{fig:solicit_parti}) asking for language models trained in academic research around the world, we obtain 61 open models representing diverse interests and priorities. Figure \ref{fig:pool_diversity} shows the diversity composition of the pool across five dimensions: base-model language support, contributor institution, task/research objective, base-model family, and training method. We are grateful to all who participated in building this new paradigm of bottom-up, compositional AI and credit them in Tables \ref{tab:main_32_1} to \ref{tab:others_2}.

\section{Experiment Details}
\label{sec:details}

\subsection{Datasets and Tasks}
Full citations for the 15 tasks across 5 domains referenced in Section~\ref{sec:experiment_settings}: \textbf{general QA} --- AGIEval \citeapp{zhong2024agieval}, ARC-challenge \citeapp{clark2018think}, MMLU-redux \citeapp{gema2025we}; \textbf{reasoning} --- BigBench-Hard \citeapp{suzgun2023challenging}, GSM8k \citeapp{cobbe2021training}, MATH \citeapp{hendrycks2021measuring}, Sciriff \citeapp{wadden2025sciriff}; \textbf{knowledge} --- WikiDYK \citeapp{zhang2025bidirectional}, PopQA \citeapp{mallen2023not}, BLEND \citeapp{myung2024blend}; \textbf{safety} --- TruthfulQA \citeapp{lin2022truthfulqa}, CocoNot \citeapp{brahman2024art}; \textbf{instruction following} --- AlpacaEval \citeapp{dubois2023alpacafarm}, Wildchat \citeapp{zhaowildchat}, Human Interest \citeapp{feng2025model}.

\subsection{Collaborative Algorithms Details}

We employ the implementations of diverse model collaboration algorithms in MoCo \citeapp{feng2026moco}, a one-stop shop for model collaboration research. We provide details for each algorithm and follow the default settings in MoCo unless otherwise stated. Code and data are available at \href{https://github.com/BunsenFeng/model_collaboration}{https://github.com/BunsenFeng/model\_collaboration}.

\begin{itemize}[leftmargin=*]
    \item \textbf{Prompt Routing} We employ the best LLM on the development set as the routing model. We then employ the following prompt format with this model to route queries: \emph{``You are an AI assistant that routes user questions to the most suitable AI model based on their descriptions. User question: [question] Model descriptions: 1: [description 1] ... n: [description n] Based on the above descriptions, which model is best suited to answer the user's question? Respond with the model number only.''}
    \item \textbf{Trained Router} We employ \textsc{Qwen/Qwen2.5-7B-Instruct} to initialize router training, and employ the reward model \textsc{Skywork/Skywork-Reward-Llama-3.1-8B-v0.2} to score candidate responses from all models and decide on the best model to generate router training data. The prompt format for router training is: \emph{``You are an AI assistant tasked with choosing the best model to answer the user's question based on the model descriptions. Model descriptions: 1: [description 1] ... n: [description n] question: [question] Based on the model descriptions, choose the best model (1-n) to answer the question.''} We LoRA \citeapp{hu2022lora} fine-tune the router on all routing labels generated with the development set for 5 epochs with learning rate $1e-5$ and effective batch size 32.
    \item \textbf{Graph Routing} We employ the graph neural architecture in \citeapp{fenggraphrouter} as the router model. Specifically, we employ SentenceTransformer to extract features from query, model descriptions, and evaluation data, and train the graph neural network with 8 hidden features, $1e-4$ learning rate, $10$ epochs, and batch size 32.
    \item \textbf{Switch Generation} We employ \textsc{Qwen/Qwen2.5-7B-Instruct} to initialize the switcher LM and fine-tune it with 500 generated switching traces, 16 switching rollouts, and employ the reward model \textsc{Skywork/Skywork-Reward-Llama-3.1-8B-v0.2} to guide the selection of rollouts. At inference time, the switcher is called every 100 tokens to decide which model is best suited to generate the next patch of 100 tokens.
    \item \textbf{Multi-Agent Refine} Models employ the following prompt format to refine their own responses based on their models' responses: \emph{``You are part of a team of AI assistants collaborating to answer the user's question. Each assistant provides their own answer: use their answers to refine and improve your own answer. Question: [question] Your previous answer: [answer] Other assistants' answers: - [answer 1] ... - [answer n-1] Please provide a refined answer to the question.''}
    \item \textbf{Multi-Agent Finetuning} We employ 2 rounds of multi-agent finetuning where the supervised fine-tuning has 1 training epoch, 16 effective batch size, and $1e-5$ learning rate.
    \item \textbf{LLM Blender} We initialize both the ranker and fuser LM as \textsc{Qwen/Qwen2.5-7B-Instruct}, and send the top-3 responses decided by the ranker to the fuser for response aggregation.
    \item \textbf{AggLM} We employ the following prompt format for the aggregator LM: \emph{``\#\#\# Problem begin [problem] \#\#\# Problem end and these solution attempts: [answer 1] ...  [answer n] It is possible that any, all, or none of these solutions are correct or complete. Carefully review the provided solutions, using them as starting points—correcting mistakes, filling in gaps, and/or combining useful ideas—to produce a final, comprehensive, and correct solution to the problem. Make sure to provide the final answer in the same format as the solutions.''} We then initialize the aggregator as \textsc{Qwen/Qwen2.5-1.5B-Instruct} and employ GRPO \citeapp{shao2024deepseekmath} with $1e-4$ learning rate, batch size 4, and 1 epoch for training.
    \item \textbf{Sparta Alignment} We execute the competition, judging, and collective alignment process for 3 iterations, with 500 instructions for competition in each iteration and 3 opponents for each model. All models start with the same reputation and gradually evolve based on the outcomes of cross-model competitions.
    \item \textbf{Heterogeneous Swarms} We employ a population of 5 LLM networks and execute the optimization process for 5 iterations, and use 0.2 inertia, 0.3 cognitive coefficient, 0.4 social coefficient, 0.05 repel coefficient, 0.5 step length, and 25\% of development set for optimization in each iteration.
    \item \textbf{Dare-Ties} We employ the MergeKit \citeapp{goddard2024arcee} implementation and assign equal weights to all models.
    \item \textbf{Greedy Soup} We employ linear merging with equal weights as the way to aggregate all models included in each soup setting.
    \item \textbf{Weight Extrapolation} We employ $k=1$ when two models, $k=2$ when four models, and $k=4$ when eight models and above. We employ linear extrapolation of model weights with the top and bottom models.
    \item \textbf{Model Swarms} We execute the optimization process for 10 iterations, and use 0.2 inertia, 0.3 cognitive coefficient, 0.4 social coefficient, 0.05 repel coefficient, 0.5 step length, and 5 patience epochs in particle swarm optimization.
\end{itemize}

\subsection{Experiment Details}
We provide additional details for various experiments and illustrations. All 95\% confidence intervals reported in this paper are computed via bootstrap resampling with 10,000 draws. The resampling unit is the individual evaluation item, drawn i.i.d. with replacement; comparisons between systems are paired, resampling the same items for both; and intervals spanning multiple domains use a stratified bootstrap that resamples within each domain. This item-level bootstrap captures uncertainty from the evaluation set, not from stochasticity across repeated runs: repeating the largest collaborative systems (up to 32 contributed models, some with iterative multi-round or training-based collaboration) many times over for run-level uncertainty estimation would multiply the already-substantial compute budget of this study and was not feasible at this scale.

\noindent \textbf{On knowledge distillation:} for weight-level collaboration, each contributed model serves as teacher and \textsc{Qwen-2.5-7B} as student. Teacher outputs are generated on the training splits of all employed datasets, 12,325 examples in total, and the student is trained on them with LoRA for 3 epochs using AdamW \citeapp{loshchilov2019decoupled} at a learning rate of $1e-5$. The same procedure and budget are used for every contributed model, so distilled checkpoints differ only in their teacher.

\noindent\textbf{Participant Evaluation Protocols.} The participant evaluation (Section \ref{sec:results}, ``Getting Back to Participants'') uses three protocols to compare contributors' individual models against the participatory system on the contributors' own priority domains: \emph{Human  Judge}, directly asking the participant to judge which response is better; \emph{LLM-as-a-Judge}, prompting \textsc{Qwen-2.5-7B} with ``Question: \{\} Answer A: \{\} Answer B: \{\} Please judge which answer is better, A or B, or if it is a tie.'' to infer quality preference; and \emph{Reward Model}, comparing reward scores from \textsc{Skywork/Skywork-Reward-Llama-3.1-8B-v0.2} on the input-output pairs, which usually yields no tie since it gives continuous scalar scores.

\noindent\textbf{Human Diversity Evaluation Protocols.} The human diversity evaluations (Section \ref{sec:results}, ``Evaluations on Reflecting Human Diversity'') use three benchmarks that differ in what they measure and what counts as success. \emph{CulturalBench} \citeapp{chiu2025culturalbench} is a multiple-choice benchmark constructed via human-AI red-teaming, an interactive process where humans probe models to surface questions they get wrong, spanning everyday topics (e.g., food customs, greeting etiquette) across global regions; unlike typical benchmarks, many questions have more than one correct answer, so success requires identifying all locally valid answers rather than defaulting to a single response. \emph{Value Kaleidoscope} \citeapp{sorensen2024value} grounds values, rights, and duties in human-written situations; we evaluate it under the Modular Pluralism setting \citeapp{feng2024modular}, which tests whether a model can identify the values relevant to a situation and correctly judge whether they are supported or opposed, treating value pluralism, multiple values in tension, as the target rather than a single universally correct value. \emph{InfinityChat} \citeapp{jiangartificial} is a set of real-world, open-ended queries that admit a wide range of plausible answers with no single ground truth; since there is no correct answer to check against, we score it under an \emph{open} setting (reward-model quality) and a \emph{diversity} setting (inter-model response homogeneity \citeapp{jiangartificial}). As with any benchmark, each has known limitations \citeapp{raji2021ai}: CulturalBench's country-level framing can understate within-country variation, Value Kaleidoscope's values are model-generated rather than sourced from the situations' original stakeholders, and InfinityChat's reward-model scoring inherits that model's own preferences; we treat all three as complementary but imperfect windows onto human diversity, not a complete measurement of it.

\begin{figure}[t]
\centering
\includegraphics[width=0.62\textwidth]{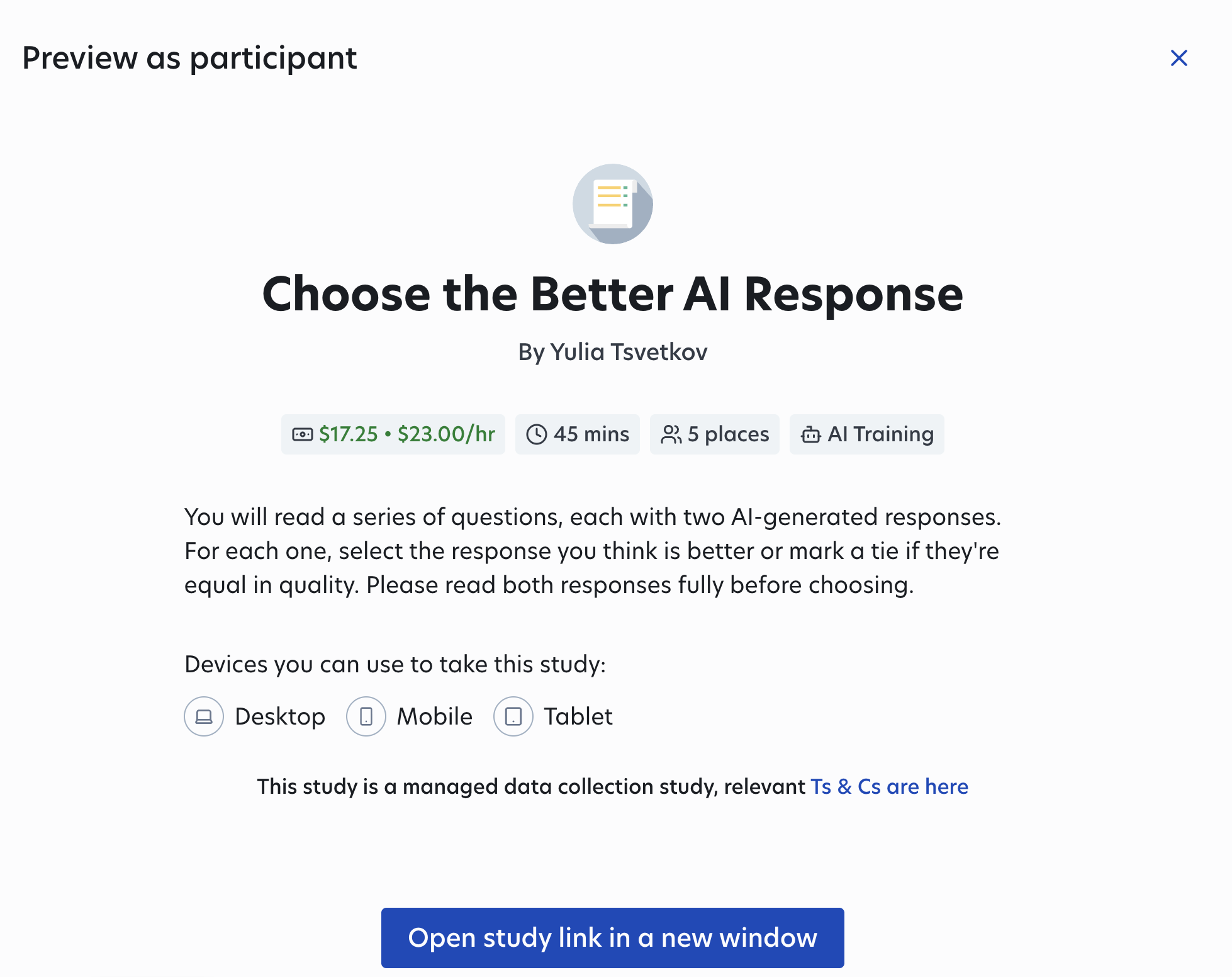}\\[4pt]
\includegraphics[width=0.62\textwidth]{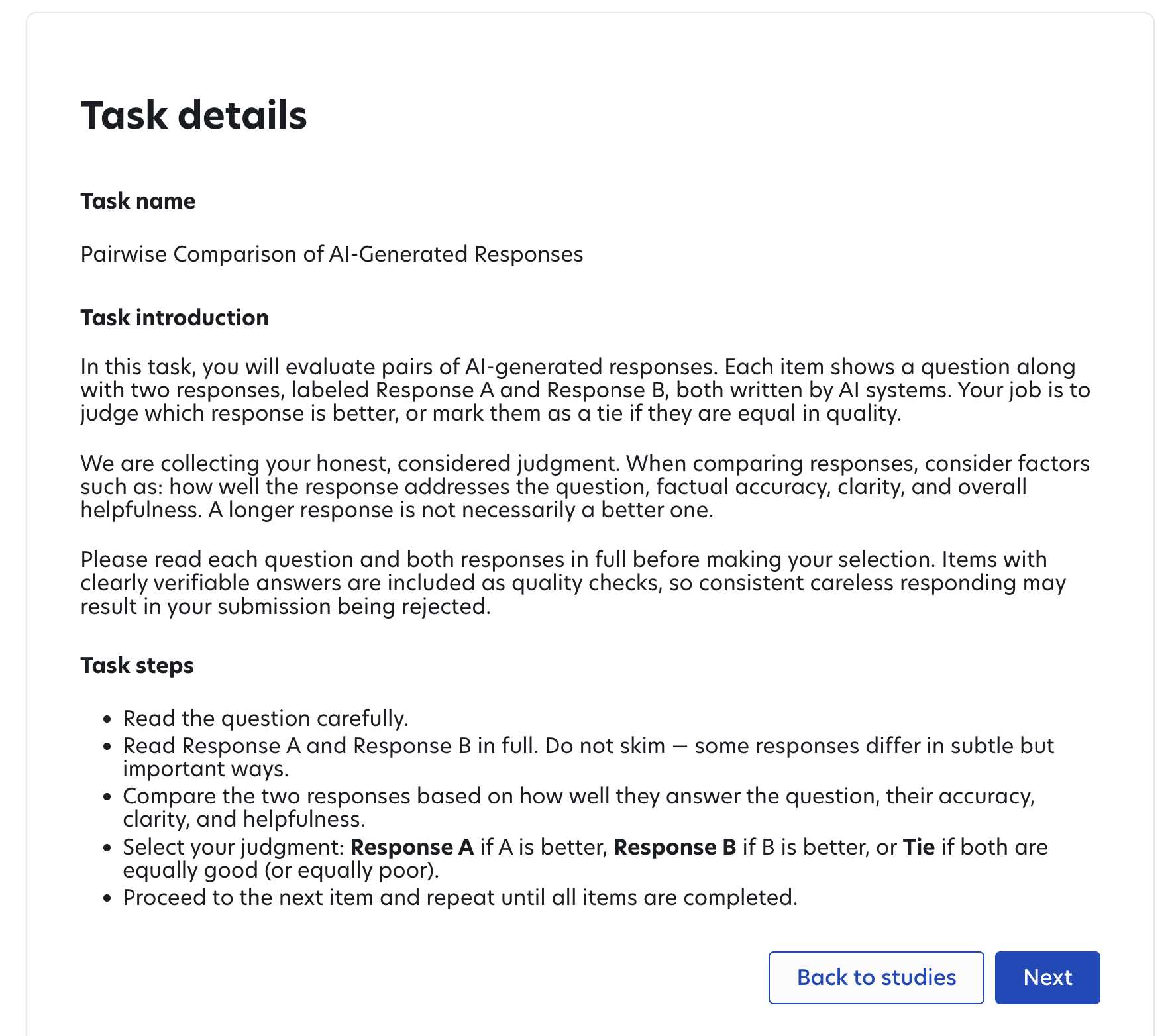}\\[4pt]
\includegraphics[width=\textwidth]{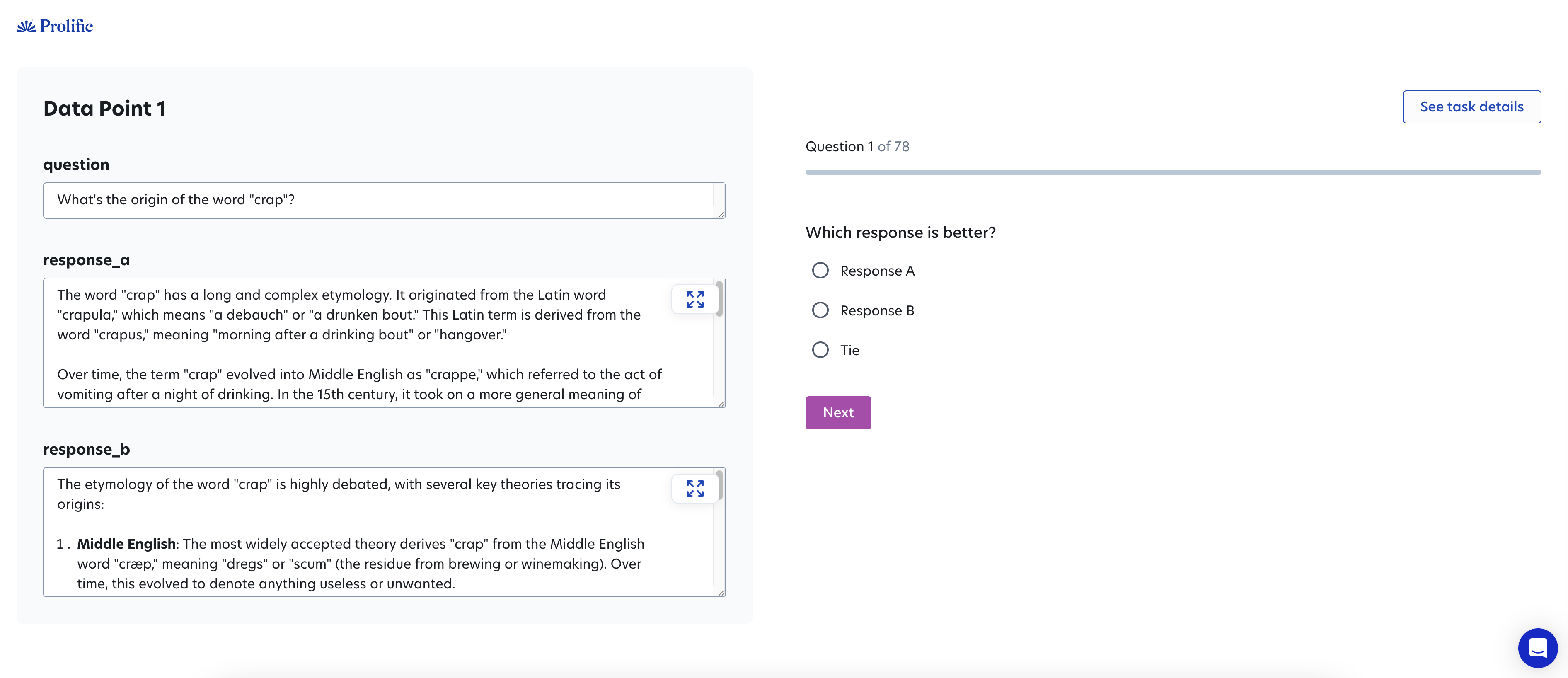}
\caption{\textbf{Third-party evaluation instrument.} Top: the Prolific study listing shown to prospective annotators. Middle: the task instructions. Bottom: the annotation interface for a single item, showing the neutral framing and the win/tie/lose response options.}
\label{fig:prolific_ui}
\end{figure}

\noindent\textbf{On the blind third-party human evaluation.} \emph{Study design.} We constructed 75 evaluation items, each pairing a query with one response from the participatory AI system and one from a contributor's individual model, spanning the three axes reported in Figure \ref{fig:human_eval}(b): creativity, factuality, and user preferences. Responses were presented as ``Response A'' and ``Response B'' with position randomized. Annotators were not told that the responses came from different systems and received no information about system identity, architecture, or provenance; the task was framed neutrally as a pairwise comparison of AI-generated responses, with instructions emphasizing full reading of both responses, judgment on relevance, factual accuracy, clarity, and helpfulness, and that length is not a proxy for quality. The study listing, task instructions, and annotation interface are reproduced in Figure \ref{fig:prolific_ui}. Win, tie, and lose rates were computed over all judgments, and the error bars in Figure \ref{fig:human_eval}(b) denote 95\% confidence intervals computed over annotators. Because every annotator judged the same 75 items, the design is fully crossed and permits a direct check on annotator consistency: 68\% of items reached a clear majority verdict, meaning more than half of the 30 annotators selected the same label.

\emph{Recruitment and compensation.} Annotators were recruited through Prolific \citeapp{prolific} as a managed data-collection study in July 2026. The study was advertised at an estimated 45 minutes and paid US\$17.25 per completed submission, a target rate of US\$23.00 per hour; the median realized completion time was 49.2 minutes (mean 51.1, s.d. 15.0), an effective median rate of approximately US\$21 per hour, exceeding Prolific's fair-pay policy.

\emph{Quality control.} Three attention-check items were interleaved with the 75 evaluation items, for 78 items per annotator. The checks probed complementary failure modes: an instructed-response item, an item pairing an on-topic with an off-topic response, and an item pairing a correct answer with an unambiguous factual absurdity. Of 34 recruited annotators, three failed the attention checks and one did not complete the task as instructed; these submissions were rejected and their slots re-recruited, yielding 30 approved annotators and 2,250 judgments.

\begin{figure}[t]
\centering
\includegraphics[width=\textwidth]{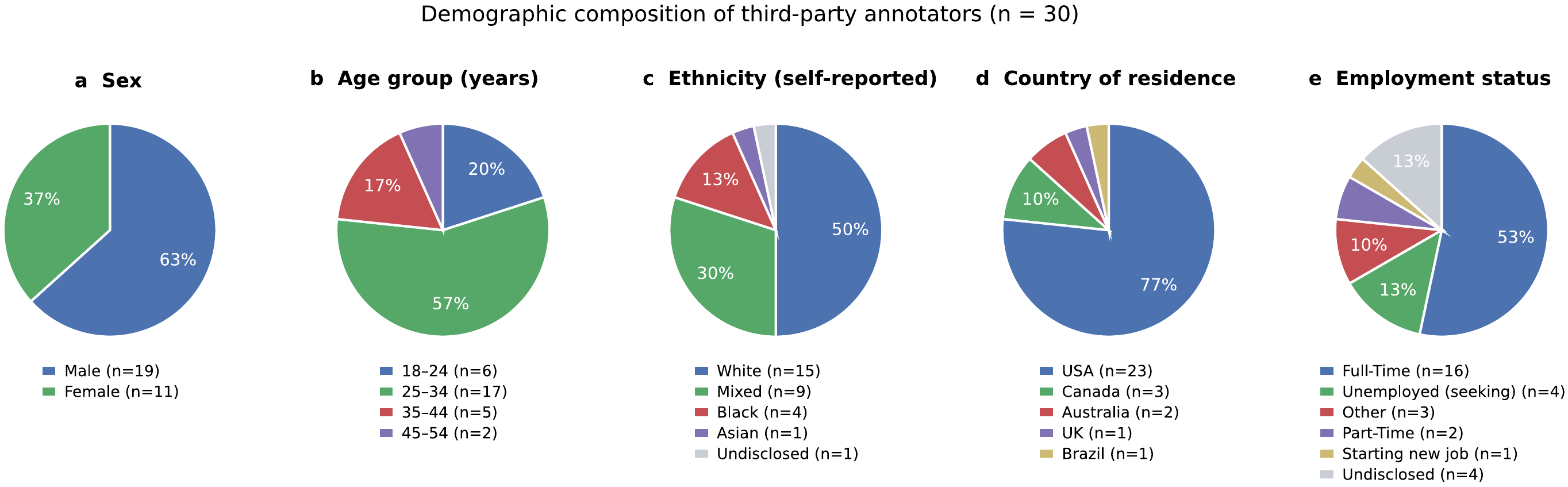}
\caption{\textbf{Annotator demographics} for the 30 approved third-party annotators, across sex, age, self-reported ethnicity, and employment status.}
\label{fig:prolific_demo}
\end{figure}

\emph{Annotator demographics.} The demographic composition of the final pool is shown in Figure \ref{fig:prolific_demo}. The pool was heterogeneous across sex, age, self-reported ethnicity, and employment status, with a mean age of 31.5 years (s.d. 8.0) and half of annotators identifying with an ethnicity other than White. All reported English as their primary language and all were experienced platform workers (median 1,465 prior approved submissions; approval rates $\geq$98\%). This matters for a system whose stated aim is to serve diverse users: the participatory advantage in Figure \ref{fig:human_eval}(b) is estimated over an evaluator pool that varies along these axes rather than a single narrow population.

\emph{Ethics.} All annotators reviewed the study description, including task content and compensation, before accepting, and were compensated through Prolific upon approval.

\noindent\textbf{On the task-data fine-tuned baseline:} \textsc{Llama-3.1-405B} is fine-tuned with QLoRA (4-bit NF4) at rank 16, $\alpha=32$, dropout 0.05, applied to all seven projection matrices, training 623M parameters or 0.15\% of the model. Training runs for 1 epoch over 10,751 examples (instruction following tasks excluded as there are not ground truth responses provided) with an effective batch size of 32 and a cosine schedule at a learning rate of $2e-4$ with 8-bit AdamW, for 336 optimizer steps.

\noindent\textbf{On metric scales and macro-averaging:}\label{sec:metric_scales} Within each evaluation domain, all constituent tasks share the same metric scale (accuracy in [0,1] for general QA, reasoning, knowledge, and safety; reward model scores for instruction following). Domain-level macro-averages are therefore over homogeneous scales. The headline 15.42\% improvement is computed as a relative improvement over the strongest baseline, calculated separately per domain, then averaged across the five domains. Since relative improvements are dimensionless percentages, this average is scale-agnostic and well-defined despite instruction following using a different absolute scale.\begin{itemize}
    \item \textbf{Figure \ref{fig:big}}: We employ the top-32 models based on evaluations on the held-out sets (Section \ref{sec:experiment_settings}), listed in Tables \ref{tab:main_32_1} to \ref{tab:main_32_3}. We report the macro-average performance across datasets in each domain. Most datasets are evaluated with accuracy (of choosing the correct option, outputting the correct final answer; AGIEval, ARC-Challenge, MMLU-redux, BigBench-Hard, GSM8k, MATH, Sciriff, BLEND, TruthfulQA), two knowledge datasets are evaluated by character-level F1-match since there could be a range of possible answers (WikiDYK, PopQA), and three instruction-following datasets are evaluated with reward model scores (specifically \textsc{Skywork/Skywork-Reward-Llama-3.1-8B-v0.2}; AlpacaEval, Wildchat, Human Interest). As a single reward model may reflect idiosyncratic biases, we additionally validate these results with multiple judges, including an open reward model (\textsc{SARM-4B} \citeapp{zhang2026interpretablesarm4b}) and three frontier API models (Gemini-3.5-Flash, Claude-Opus-4.8, and GPT-5.5), with results in Appendix \ref{sec:other_judges}. Each horizontal dashed line represents the best performance across all baselines in this group for visual simplicity: if the scaling trend rises above these horizontal lines, it means scaling participation successfully outperforms all non-modular and/or non-monolithic models in this category.
    \item \textbf{Figure \ref{fig:diversity_evals}}: CulturalBench and Kaleidoscope are evaluated with prediction accuracy, the open setting of InfinityChat with reward model \textsc{Skywork/Skywork-Reward-Llama-3.1-8B-v0.2}, and the diversity setting with the inter-model homogeneity metric \citeapp{jiangartificial}.
    \item \textbf{Figure \ref{fig:model_diversity}}: Setting 1 employs the top-8 models in Table \ref{tab:main_32_1} in that order. Setting 2 and 3 go down the model list in Tables \ref{tab:main_32_1} to \ref{tab:main_32_3} to select models in eligible institutions; Settings 1-3 thus vary diversity qualitatively, by which and how many distinct models, institutions, or countries are represented. Setting 4 instead measures diversity quantitatively: it employs random sampling from the top-32 models, ranked by two statistics computed on each pool's $8\times8$ pairwise agreement matrix on InfinityChat (diagonal 1; off-diagonal entries are the judged between-model agreement rate from Appendix \ref{sec:hivemind}'s protocol, or a model's own within-model self-consistency rate where a pool contains duplicate slots of the same model): mean pairwise disagreement (one minus the average off-diagonal entry) of 0.5080, 0.5487, 0.5500, 0.5933, and the log-determinant of the same matrix, a measure of the volume spanned by the pool's response vectors that drops sharply toward $-\infty$ as near-duplicate models make rows linearly dependent, of $-1.2306$, $-0.9991$, $-0.9900$, $-0.8703$ (rank 4 to rank 1, both statistics monotonic). This confirms that setting 4 is quantitatively increasing in diversity, showing that system performance trends up with both qualitative and quantitative diversity among contributed models.
    \item \textbf{Figure \ref{fig:collaborative_emergence}}: We provide examples of cases where all 32 models could not individually solve the problem, but a collaborative system could, in Tables \ref{tab:emergence_example_1} to \ref{tab:emergence_example_3}. With the collaboration and contribution of multiple models, the ceiling of AI capabilities is greatly improved.
    \item \textbf{Figure \ref{fig:compositional_strength}}: We run Pearson correlation between the individual and compositional strength of the top-16 models, using six collaboration methods (Multiagent Refine, Heterogeneous Swarms, Multiagent Finetuning, DARE-TIES, Greedy Soup, Model Swarms) and two datasets (AGIEval, BBH), yielding 48 model-method pairs per subfigure.
    \item \textbf{Figure \ref{fig:human_eval}}: The four human evaluators in panel (a) are the contributors behind four of the contributed models, one per evaluation domain, and are the participants who responded to our optional invitation to take part in this study. Each judged response pairs from their own model and the compositional system, on the original data and priorities for which they developed their model. Each domain consists of 25 human evaluation problems and 100 problems with automatic evaluatison (LLM-as-a-judge, reward model), summing up to 100 human judgements and 400 automatic evaluations in total.

    \item \textbf{Figure \ref{fig:solicit_parti}}: The collection time range is October 2025 to February 2026, where we obtained 61 contributed models.
\end{itemize}

\section{Additional Experiments}
\label{sec:additional_experiments}

\begin{figure}[t]
\centering
\includegraphics[width=1.0\textwidth]{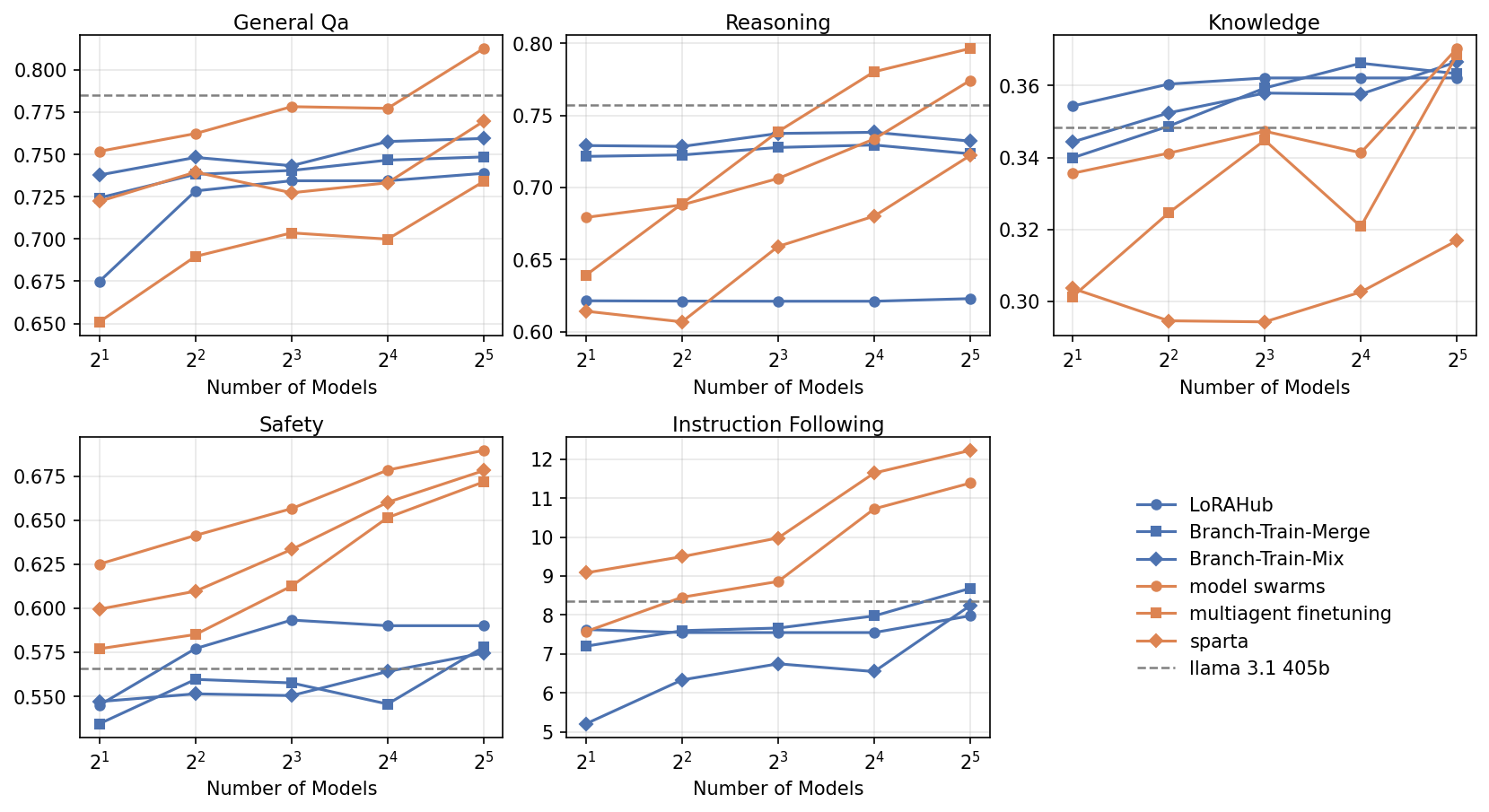}
\caption{\textbf{Comparison with heterogeneous expert recycling methods.} Scaling trends of LoraHub, Branch-Train-Merge, and Branch-Train-MiX compared to three strong methods in the main paper. These methods generally trend up with more models, but are outperformed by the best methods in the main paper.}
\label{fig:lorahub}
\end{figure}

\noindent\textbf{Comparison with Heterogeneous Expert Recycling Methods}
\label{sec:lorahub}
In addition to the 14 collaboration algorithms in the main paper, we compare with three methods from the heterogeneous expert recycling literature: LoraHub \citeapp{huanglorahub}, Branch-Train-Merge \citeapp{li2022branch}, and Branch-Train-MiX \citeapp{sukhbaatarbranch}. These methods specifically focus on combining specialized expert models. We follow the same experimental setup as Section \ref{sec:experiment_settings}, evaluating the scaling trend from 2 to 32 contributed models across five domains and compare them against the three strongest approaches in the main paper. Results in Figure \ref{fig:lorahub} show that all three methods do benefit from scaling participation, confirming that the paradigm generalizes beyond our primary collaboration algorithms; however, they are consistently outperformed by the best methods in the main paper, suggesting that deeper integration strategies such as model swarms and multi-agent finetuning are better suited for orchestrating decentralized, heterogeneous model contributions at scale.

\begin{figure}[t]
\centering
\includegraphics[width=1.0\textwidth]{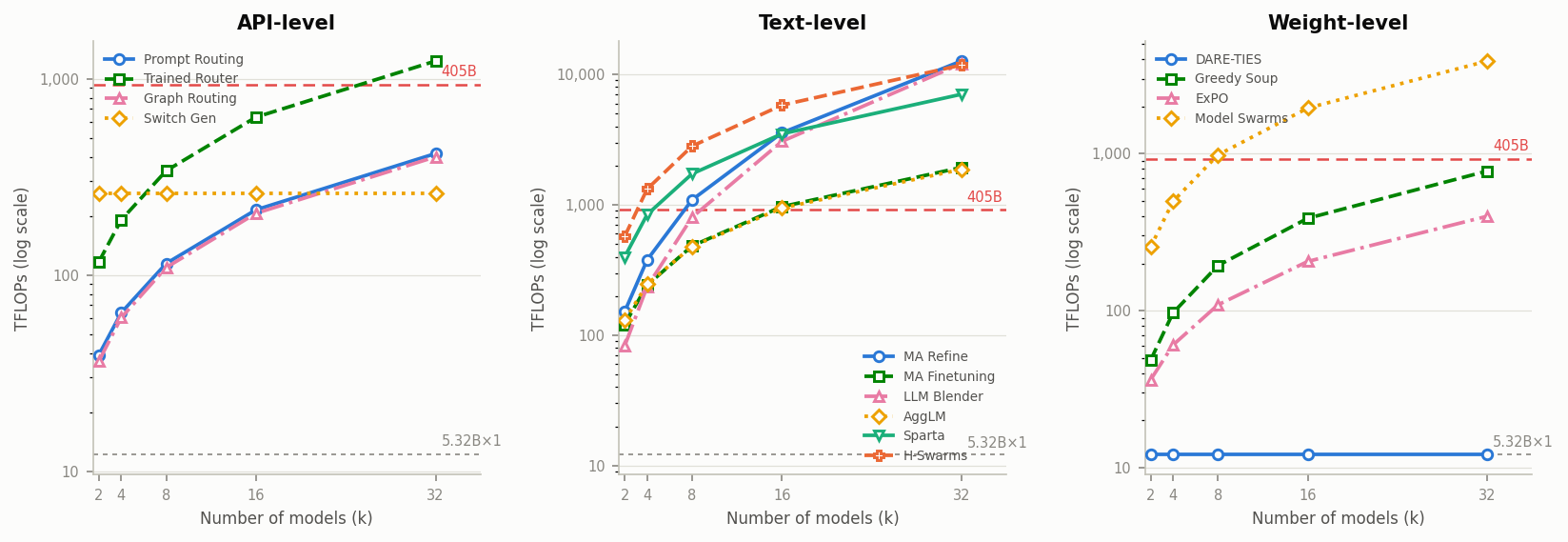}
\caption{\textbf{Amortized FLOPs (post-training + inference) of collaboration methods.} Per-example inference FLOPs plus post-training FLOPs expressed per post-training example, in TFLOPs (log scale), as participation scales from 2 to 32 models. Post-training is a one-time cost; see Figure \ref{fig:flops_inf_train} for the two components separately. The red dashed line marks Llama 3.1 405B (927 TFLOPs). Several methods exceed this at large k.}
\label{fig:flops_total}
\end{figure}

\begin{figure}[t]
\centering
\includegraphics[width=1.0\textwidth]{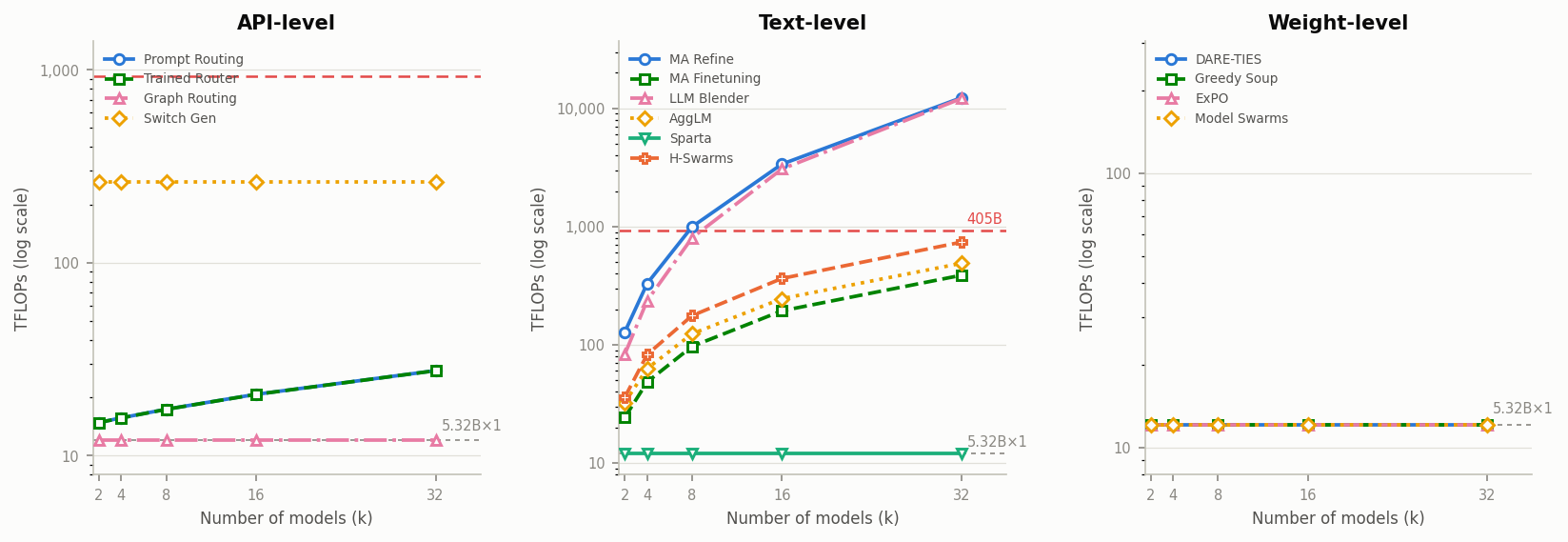}
\vspace{4pt}
\includegraphics[width=1.0\textwidth]{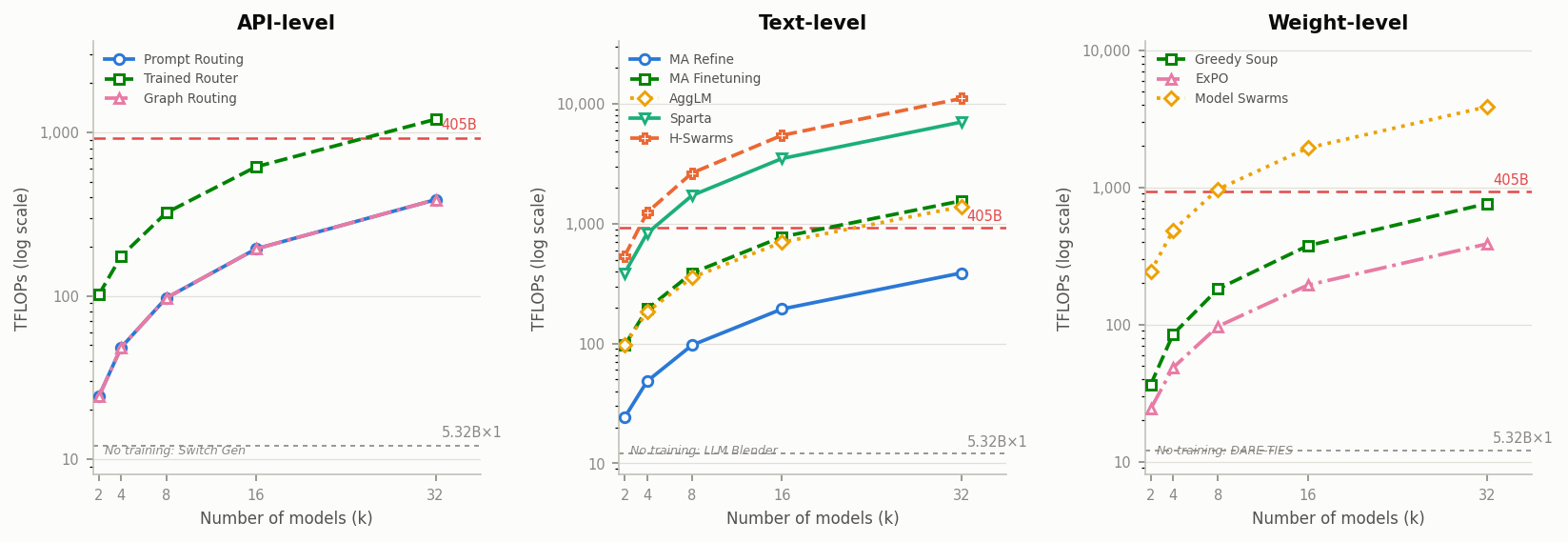}
\caption{\textbf{Per-query inference FLOPs (top) and post-training FLOPs per post-training example (bottom) of collaboration methods separately.} Weight-level methods and Sparta Alignment achieve single-model inference cost (bottom dashed line, 12.11 TFLOPs) regardless of pool size, well below the 405B line. Post-training is a one-time overhead, not incurred per query; it is shown per post-training example so the two panels share a scale.}
\label{fig:flops_inf_train}
\end{figure}

\begin{figure}[t]
\centering
\includegraphics[width=1.0\textwidth]{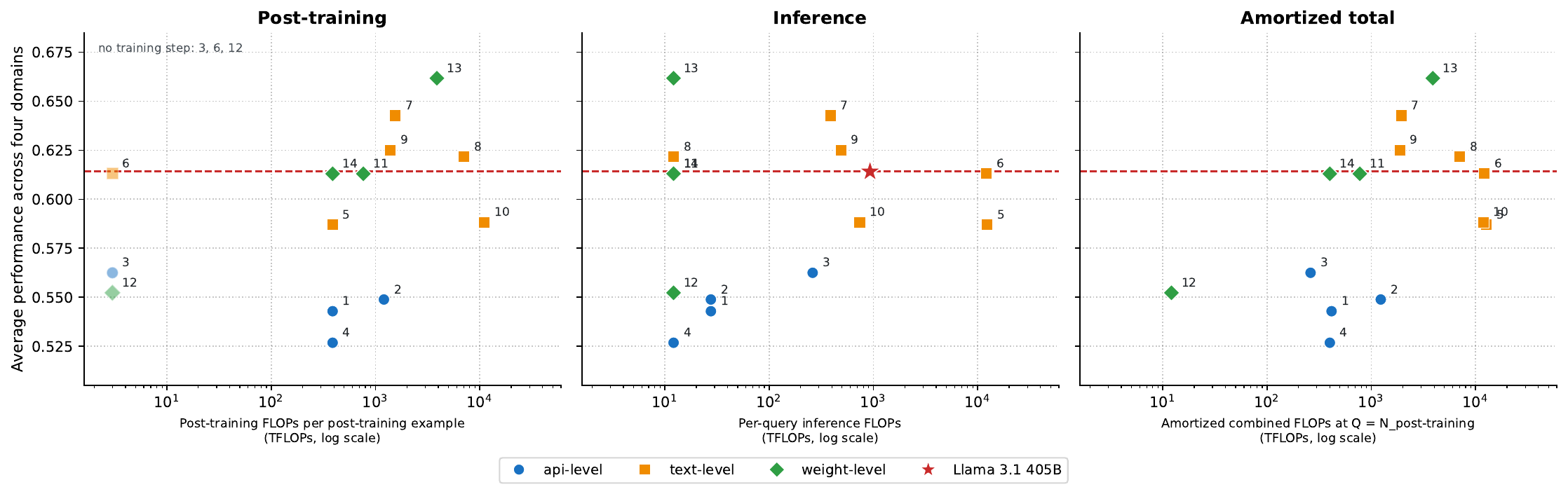}
\caption{\textbf{Cost-normalized comparison of collaboration methods}, at 32 contributed models, against post-training FLOPs per post-training example, per-query inference FLOPs, and their amortized combination at $Q=N_{\text{post-training}}$. The vertical axis averages general QA, reasoning, knowledge, and safety; instruction following is excluded because its reward-model scores use a different scale. The dashed line marks \textsc{Llama-3.1-405B}, whose pre-training cost is not charged on the training and total axes. Methods are: 1 prompt routing, 2 trained router, 3 switch generation, 4 graph routing, 5 multi-agent refine, 6 LLM blender, 7 multi-agent finetuning, 8 sparta alignment, 9 AggLM, 10 heterogeneous swarms, 11 greedy soup, 12 DARE-TIES, 13 model swarms, 14 ExPO.}
\label{fig:cost_normalized}
\end{figure}

\noindent\textbf{FLOPs Analysis}
\label{sec:flops}
We report total training and inference FLOPs for each collaboration method, computed under average prompt length (114 tokens) and maximum response length (1024 tokens) across all datasets, using the standard $2 \times N \times S$ rule of thumb for inference and $6 \times N \times S$ for training. Deployment cost is $C_{\text{total}}(Q) = C_{\text{post-training}} + Q \cdot C_{\text{inference}}$, where post-training is paid once and $Q$ is the number of queries served. Figure \ref{fig:flops_total} plots the inference FLOPs for a single test example together with post-training FLOPs expressed per post-training example, that is, $C_{\text{total}}(Q)/Q$ at a horizon $Q$ equal to the size of the post-training set. This places the two terms on one scale for comparing methods; the amortized term shrinks as $Q$ grows. Figure \ref{fig:flops_inf_train} therefore reports the two components separately, and the cost-normalized analysis in Figure \ref{fig:cost_normalized} reports all three views separately and bases deployment-cost comparisons on the inference panel, which is the recurring cost at deployment. Multi-Agent Refine, Heterogeneous Swarms, and LLM Blender at 32 models exceed a single Llama 3.1 405B inference call on this amortized basis (927 TFLOPs, red dashed line).

However, the picture is more nuanced when training and inference costs are examined separately (Figure \ref{fig:flops_inf_train}). At inference time, weight-level methods (DARE-TIES, Greedy Soup, ExPO, Model Swarms) and Sparta Alignment all run at the cost of a single 5.32B model (12.11 TFLOPs) regardless of how many models participated, roughly 76 times cheaper per query than \textsc{Llama-3.1-405B}. The higher total cost of some methods is driven by one-time training overhead, not per-query inference cost. Furthermore, the pre-training cost of the 32 contributed models is not charged to our system: participants trained these models for their own research purposes and that compute is amortized across those original uses. The relevant cost is only the marginal compute added on top of already-existing models. Finally, because contributed models are independent modules, many methods can run them in parallel, offering potential advantages in wall-clock time and GPU memory over a single dense 405B model.

\noindent\textbf{Model Call Counts and Communication Rounds}
\label{sec:call_counts}
\textit{We emphasize that every call reported in Table \ref{tab:call_counts} is to one of the contributed models, which average 5.32B parameters and are almost all under 10B, an order of magnitude or more smaller than the monolithic baselines (e.g., 405B) these systems are compared against.} Table \ref{tab:call_counts} reports the recurring, per-query inference cost of each collaboration method: the number of underlying model calls made for a single test query as a function of pool size $k$, together with the number of communication rounds for methods that are iterative. One-time training-only calls (e.g., offline preference optimization, particle-swarm search, greedy merging) are a separate, non-recurring cost and are reported in the FLOPs analysis in Figure \ref{fig:flops_inf_train}.

\begin{sidewaystable}
\centering
\small
\caption{Recurring, per-query inference cost in number of model calls and communication rounds for each collaboration method, as a function of pool size $k$. Switch Generation is reported at $L=512$, patch size 100. ``---'' marks configurations not evaluated in the main experiments: graph routing and heterogeneous swarms do not have results for 2 models, as 2 models do not constitute a graph network.}
\label{tab:call_counts}
\begin{tabularx}{\textheight}{
  >{\raggedright\arraybackslash}p{3.4cm}
  >{\centering\arraybackslash}p{2.2cm}
  >{\centering\arraybackslash}p{1.1cm}
  >{\centering\arraybackslash}p{1.1cm}
  >{\centering\arraybackslash}p{1.1cm}
  >{\centering\arraybackslash}p{1.1cm}
  >{\centering\arraybackslash}p{1.1cm}
  >{\raggedright\arraybackslash}X
}
\toprule
\textbf{Method} & \textbf{Formula} & \textbf{k=2} & \textbf{k=4} & \textbf{k=8} & \textbf{k=16} & \textbf{k=32} & \textbf{Communication rounds} \\
\midrule
1. Prompt Routing & 2 & 2 & 2 & 2 & 2 & 2 & none \\
2. Trained Router & 2 & 2 & 2 & 2 & 2 & 2 & none \\
3. Graph Routing & 1 & --- & 1 & 1 & 1 & 1 & none \\
4. Switch Generation & $2 \times 5$ & 10 & 10 & 10 & 10 & 10 & $L=512$ (5 patches) \\
5. Multi-Agent Refine & $2k+1$ & 5 & 9 & 17 & 33 & 65 & 2 (1 initial + 1 refine) \\
6. LLM Blender & $k+2$ & 4 & 6 & 10 & 18 & 34 & 1 listwise rank + 1 fuse \\
7. Multi-Agent Finetuning & $2k$ & 4 & 8 & 16 & 32 & 64 & 2 fixed \\
8. Sparta Alignment & 1 & 1 & 1 & 1 & 1 & 1 & none (offline training only) \\
9. AggLM & $k+1$ & 3 & 5 & 9 & 17 & 33 & 2 \\
10. Heterogeneous Swarms & $k$ & --- & 4 & 8 & 16 & 32 & 1 topological pass (fixed DAG from offline search) \\
11. Greedy Soup & 1 & 1 & 1 & 1 & 1 & 1 & none \\
12. DARE-TIES & 1 & 1 & 1 & 1 & 1 & 1 & none \\
13. Model Swarms & 1 & 1 & 1 & 1 & 1 & 1 & none \\
14. ExPO & 1 & 1 & 1 & 1 & 1 & 1 & none \\
\bottomrule
\end{tabularx}
\end{sidewaystable}

\noindent\textbf{Measured Latency and Memory}
\label{sec:measured_cost}
The call counts above are a recurring inference cost in the number of model invocations; we complement this with directly measured wall-clock latency and GPU memory, at $n=8$ contributed models on AGIEval, repeated 3 times per method on a single node of NVIDIA H200 GPUs. Table \ref{tab:measured_cost} reports, for each method, the median (and interquartile range) latency of a single model call, peak GPU memory, per-query wall-clock time (test-phase only, excluding one-time setup), and a one-time setup cost measured as the time from job start to the first model call.

\begin{sidewaystable}
\centering
\small
\caption{Measured, per-query call latency (median [interquartile range]), peak GPU memory, per-query wall-clock time, and one-time setup cost (time to first model call, bundling loading, conversion, and merging where applicable), at $n=8$ models on \emph{agieval\_tiny}, median over 3 repetitions.}
\label{tab:measured_cost}
\begin{tabularx}{\textheight}{
  >{\raggedright\arraybackslash}p{3.4cm}
  >{\centering\arraybackslash}p{3.0cm}
  >{\centering\arraybackslash}p{1.6cm}
  >{\centering\arraybackslash}p{2.0cm}
  >{\raggedright\arraybackslash}X
}
\toprule
\textbf{Method} & \textbf{Call latency (s)} & \textbf{Peak mem (GB)} & \textbf{Per-query wall-clock (s)} & \textbf{One-time setup (s)} \\
\midrule
Prompt Routing & 0.05 [0.04, 2.52] & 15.35 & 2.05 & 24.3 \\
Trained Router & 2.62 [1.84, 3.74] & 15.49 & 5.01 & 14.6 \\
Graph Routing & 4.19 [0.12, 7.40] & 15.32 & 4.89 & 29.6 \\
Switch Generation & 0.08 [0.04, 1.41] & 15.35 & 8.69 & 19.2 \\
Multi-Agent Refine & 5.45 [2.83, 10.56] & 15.48 & 20.31 & 13.8 \\
LLM Blender & 3.05 [0.45, 8.66] & 15.33 & 8.97 & 15.2 \\
Multi-Agent Finetuning & 4.05 [2.06, 7.64] & 15.63 & 13.35 & 14.4 \\
Sparta Alignment & 12.48 [12.38, 12.65] & 17.17 & 1.56 & 13.9 \\
AggLM & 3.70 [1.64, 9.21] & 15.31 & 9.75 & 37.2 \\
Heterogeneous Swarms & 1.15 [0.40, 2.78] & 15.36 & 15.93 & 15.8 \\
Greedy Soup & 1.67 [1.00, 3.69] & 15.32 & 2.70 & 132.9 \\
DARE-TIES & 2.16 [1.06, 3.98] & 15.32 & 2.83 & 318.8 \\
Model Swarms & 2.77 [1.77, 6.42] & 15.48 & 4.41 & 43.0 \\
ExPO & 1.95 [1.11, 3.84] & 15.32 & 2.60 & 131.0 \\
\bottomrule
\end{tabularx}
\end{sidewaystable}

\noindent\textbf{Cost-Normalized Comparison}
\label{sec:cost_normalized}
We plot performance directly against cost in Figure \ref{fig:cost_normalized}, with each method at 32 contributed models and three separate cost axes: post-training FLOPs per post-training example, per-query inference FLOPs, and their amortized combination at $Q=N_{\text{post-training}}$. The vertical axis averages the four domains scored on a common $[0,1]$ accuracy scale, excluding instruction following, whose reward-model scores are on a different scale and would otherwise dominate the average.

The comparison is favorable in a way that aggregate performance alone does not reveal. On inference cost, which is the recurring cost at deployment, model swarms attains the highest average performance of any method while running at the cost of a single 5.32B model, roughly 76 times cheaper per query than \textsc{Llama-3.1-405B}, which it also outperforms. Sparta alignment, greedy soup, and ExPO sit at the same inference cost and match or exceed the 405B baseline. The methods that are genuinely expensive per query are those issuing many calls with iterative refinement: multi-agent refine and LLM blender cost more than an order of magnitude beyond the 405B baseline while performing no better. Training cost tells a different story, since it is incurred once rather than per query: model swarms and sparta alignment are mid-range, while heterogeneous swarms is the most expensive to optimize without a corresponding gain. We also note that the training axis charges our methods for optimization while charging \textsc{Llama-3.1-405B} nothing for its pre-training, so it understates the comparison in the baseline's favor. The practical conclusion is that cost varies sharply across methods, and that the strongest configurations of scaling participation are also among the cheapest to deploy.

\noindent\textbf{Training-Matched Baseline}
\label{sec:lora_baseline}
We LoRA fine-tune \textsc{Llama-3.1-405B} on the training splits of the employed datasets, and compare it against both its off-the-shelf form and the best collaborative system per domain.

\begin{figure}[t]
\centering
\includegraphics[width=1.0\textwidth]{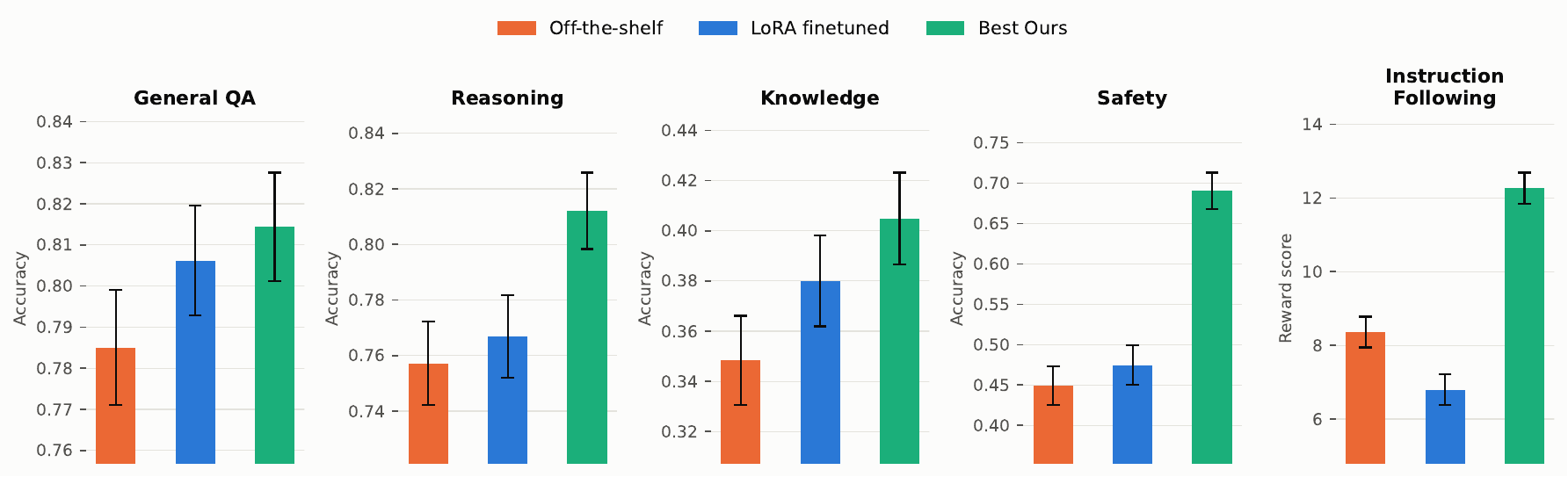}
\caption{\textbf{Task-data fine-tuned baseline comparison.} \textsc{Llama-3.1-405B} off the shelf, finetuned on task data, and the best collaborative system per domain. Training improves the baseline on four of five domains but does not close the gap; on instruction following it degrades performance relative to the released model. Error bars represent 95\% confidence intervals.}
\label{fig:lora_baseline}
\end{figure}

Results in Figure \ref{fig:lora_baseline} show that training does help the baseline, but does not account for the gap. LoRA fine-tuning improves \textsc{Llama-3.1-405B} on general QA, reasoning, knowledge, and safety, yet the best collaborative system remains ahead in every one of those domains. On instruction following the fine-tuned baseline is worse than its off-the-shelf form; one possible explanation is that supervised fine-tuning on a modest set disturbs the extensive preference tuning already present in the released model, though we do not test this. We conclude that this fine-tuning narrows the comparison in some domains but does not explain the advantage of scaling participation. This is a task-data fine-tuned baseline answering the request for minimal training; it is not matched to the collaboration methods in examples, optimization steps, or FLOPs.

\begin{figure}[t]
\centering
\includegraphics[width=1.0\textwidth]{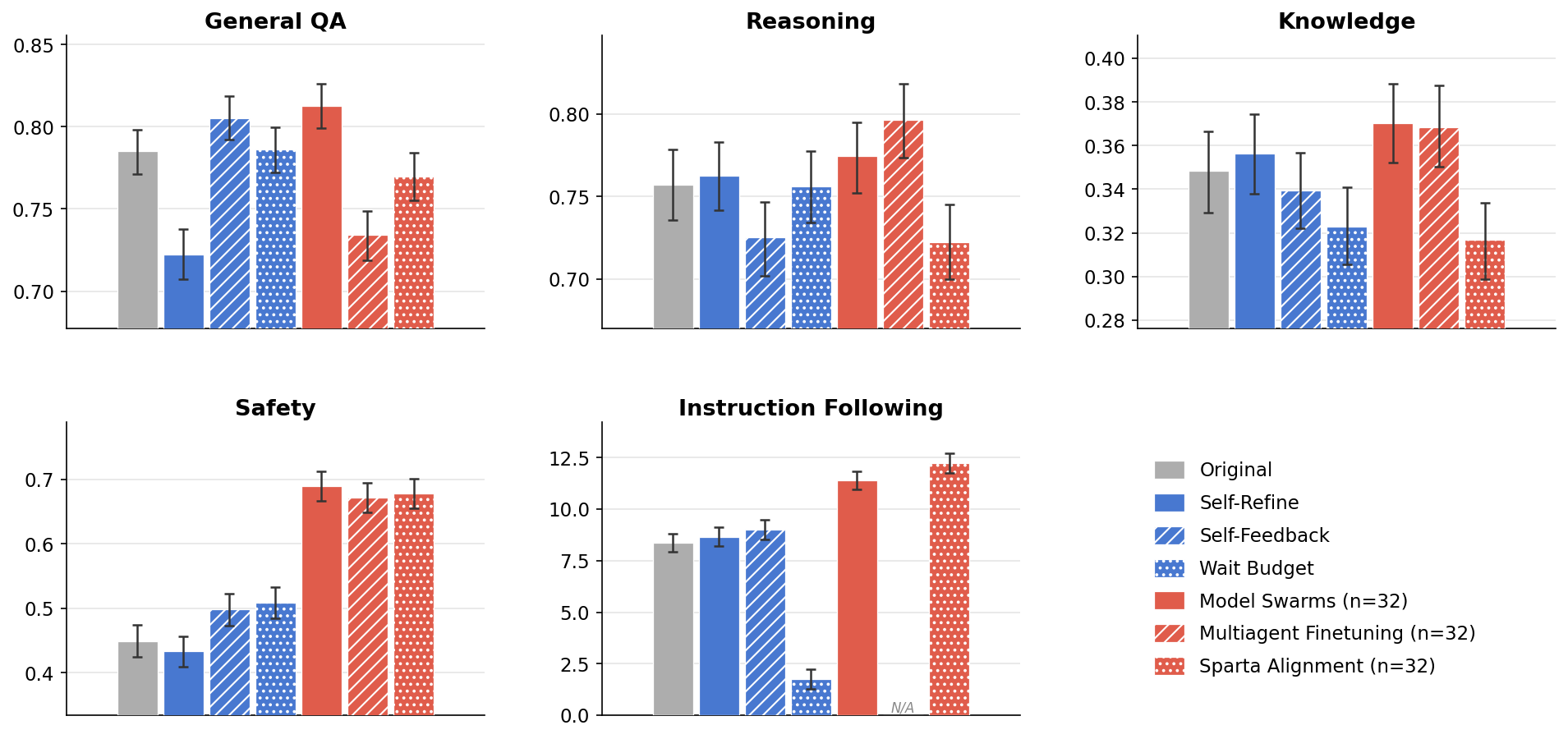}
\caption{\textbf{Compute-matched comparison with test-time scaling baselines.} Our top three participatory methods (Model Swarms, Multiagent Finetuning, Sparta Alignment, all n=32) vs. Self-Refine, Self-Feedback, and Wait Budget applied to LLaMA 3.1 405B. Multiagent finetuning works on tasks with objective ground truths and is thus not compatible with instruction following tasks featuring open-ended generation. The best participatory AI system still outperforms compute-matched dense baselines. The gains therefore cannot be explained by additional compute alone under these compute-matched baselines; together with the fixed-compute diversity ablation, this is consistent with a benefit from contribution diversity. Error bars represent 95\% confidence intervals.}
\label{fig:compute_match}
\end{figure}

\noindent\textbf{Compute-Matched Baseline: Test-Time Scaling}
\label{sec:compute_match}
To directly address the concern that improvements may stem from extra compute rather than participation, we compare the top three methods (Model Swarms, Multiagent Finetuning, and Sparta Alignment, all with 32 models) against three test-time scaling (TTS) approaches applied to the large dense model (LLaMa 3.1 405B): Self-Refine \citeapp{madaan2023self}, in which the model critiques and rewrites its own answer over successive rounds; Self-Feedback \citeapp{feng2024don}, in which the model gives feedback to its own responses; and Wait Budget (s1-style scaling \citeapp{muennighoffs1}), which appends ``Wait'' to the generation to force a longer reasoning trace. These TTS baselines spend comparable extra compute on a single dense model rather than across multiple contributed models. We use the FLOPs analysis in Figure \ref{fig:flops_total} and scale up test-time compute with these methods until they match/exceed the FLOPs of the most expensive method among the three, Sparta Alignment and n=32. Matching uses the same amortized convention as Figure \ref{fig:flops_total}, namely post-training charged per post-training example, $Q=N_{\text{post-training}}$. Results in Figure \ref{fig:compute_match} show that our participatory methods consistently outperform all TTS baselines across evaluation domains, indicating that the gains cannot be explained by additional compute alone under these compute-matched baselines. Together with the fixed-compute diversity ablation (Figure \ref{fig:model_diversity}), this is consistent with a benefit arising from the diversity of decentralized contributions.

\noindent\textbf{Latent-Capability Scaling Axis}
\label{sec:latent_capability}
In addition to FLOPs, observational scaling law (OSL) \citeapp{ruan2024observational} builds a low-dimensional ``latent capability'' axis directly from a model's own benchmark scores, since raw compute does not transfer cleanly across model families with different training efficiencies. We construct an analogous axis for our pool: each of the top-32 models' individual test scores across 13 datasets (AGIEval, ARC-Challenge, BBH, BLEND, CocoNot, GSM8k, MATH, MMLU-redux, PopQA, ScienceMeter, TruthfulQA, WikiDYK, and AlpacaEval). We standardize each column and run PCA, take PC-1' as a general-capability direction, and apply them to each dataset's separately min-max-normalized score, giving every model a non-negative latent-capability score.

\begin{figure}[t]
\centering
\includegraphics[width=1.0\textwidth]{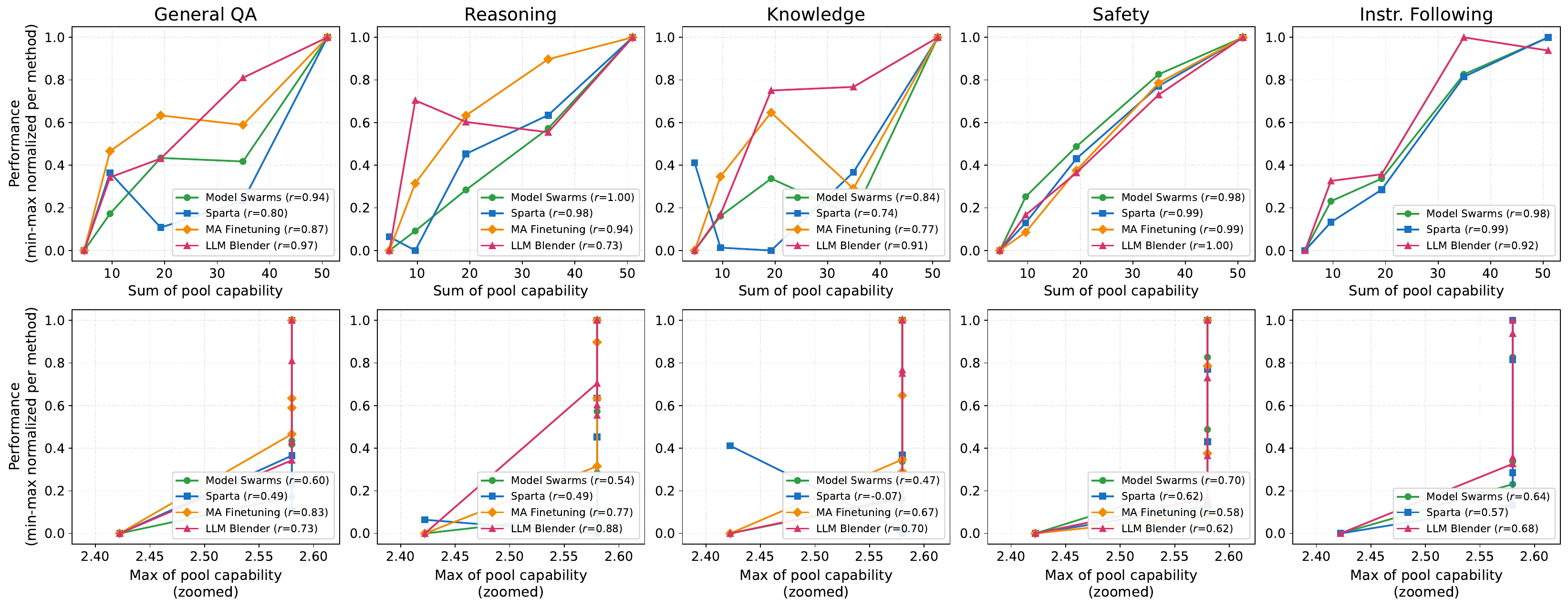}
\caption{\textbf{System performance vs. sum (top) and max (bottom) of pool latent capability.} The four top-ranked methods' performance (min-max normalized per method for comparability) at each pool size ($k=2,4,8,16,32$) against the pool's aggregated latent-capability score, with per-method Pearson $r$ in the legend. The max-aggregate x-axis is zoomed to its narrow observed range. Sum tracks performance closely across all four methods; max saturates by $k=4$ and moves little afterward.}
\label{fig:latent_capability}
\end{figure}

For each pool (2, 4, 8, 16, 32), we aggregate this score by sum and max, and plot against the four top-ranked methods' performance (Model Swarms, Sparta Alignment, Multiagent Finetuning, and LLM Blender, Appendix \ref{sec:method_ranking}) at that pool size (Figure \ref{fig:latent_capability}). The sum-aggregate tracks pool size and performance closely for all four methods: sum increases with pool size, but how closely it tracks performance is a genuine empirical result ($r=0.73$ to $1.00$ across the four methods and five domains, mostly $r>0.9$). The max-aggregate saturates quickly, leaving little room to move afterward; its correlation with performance ($r=-0.07$ to $0.88$, noisier and inconsistent in sign for one method-domain pair) is correspondingly weaker and less informative than sum's.

\noindent\textbf{Full Method Rankings}
\label{sec:method_ranking}
Averaging each method's rank across the five evaluation domains at 32 contributed models (Section \ref{sec:results}), the three best-performing methods overall are model swarms (average rank \#2), sparta alignment (average rank \#4.6), and multiagent finetuning (average rank \#5.4); all three involve training the parameters of participating models, unlike methods that merely route or prompt frozen models. Rankings also vary by domain: multiagent refine \citeapp{du2023improving}, which has models debate and refine each other's answers, ranks best on reasoning but lags on instruction following; heterogeneous swarms \citeapp{fengheterogeneous}, which organizes models in graph structures to divide and conquer problems, ranks best on general QA but lags on safety.

\begin{figure}[t]
\centering
\includegraphics[width=1.0\textwidth]{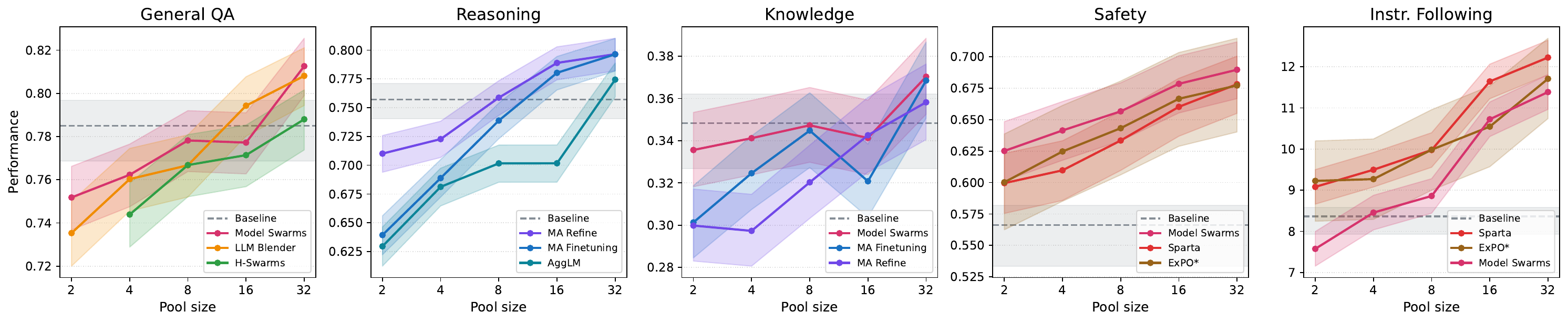}
\caption{\textbf{Scaling curves of top-3 methods and baseline, with 95\% confidence intervals, per domain.} The three best-performing collaboration methods and the strongest baseline per domain, at pool sizes 2, 4, 8, 16, and 32 models; $n=32$ values match Table \ref{tab:ci}.}
\label{fig:ci_top3}
\end{figure}

\noindent\textbf{Confidence Intervals for Main Results}
\label{sec:main_ci}
Adding per-method error bars to Figure \ref{fig:big} would be unreadable given the number of curves. Instead, Table \ref{tab:ci} reports the performance and 95\% bootstrap confidence intervals for the three best-performing collaboration methods and the strongest baseline in each evaluation domain, and Figure \ref{fig:ci_top3} shows the scaling curves of these top-3 methods and baseline, with 95\% confidence intervals at every pool size. The non-overlapping CIs in most domains -- and the aggregate 95\% CI of [10.09\%, 21.13\%] on the headline 15.42\% improvement -- confirm that the gains of scaling participation are statistically significant.

\begin{table}[t]
\centering
\small
\caption{Performance and 95\% bootstrap confidence intervals (CI) for the top three collaboration methods and the strongest baseline per domain (n=32 models). Non-overlapping CIs with the baseline indicate statistically significant improvement.}
\label{tab:ci}
\begin{tabular}{llcc}
\toprule
\textbf{Domain} & \textbf{Method} & \textbf{Performance} & \textbf{95\% CI} \\
\midrule
\multirow{4}{*}{General QA}
  & Model Swarms      & 0.8126 & [0.7967, 0.8234] \\
  & LLM Blender       & 0.8081 & [0.7915, 0.8184] \\
  & H-Swarms          & 0.7880 & [0.7709, 0.7988] \\
  & Baseline          & 0.7850 & [0.7687, 0.7967] \\
\midrule
\multirow{4}{*}{Reasoning}
  & MA Refine         & 0.7965 & [0.7812, 0.8096] \\
  & MA Finetuning     & 0.7954 & [0.7796, 0.8081] \\
  & AggLM             & 0.7744 & [0.7594, 0.7889] \\
  & Baseline          & 0.7572 & [0.7406, 0.7709] \\
\midrule
\multirow{4}{*}{Knowledge}
  & Model Swarms      & 0.3703 & [0.3440, 0.3798] \\
  & MA Finetuning     & 0.3685 & [0.3422, 0.3779] \\
  & MA Refine         & 0.3582 & [0.3340, 0.3695] \\
  & Baseline          & 0.3483 & [0.3268, 0.3622] \\
\midrule
\multirow{4}{*}{Safety}
  & Model Swarms      & 0.6897 & [0.6628, 0.7080] \\
  & Sparta            & 0.6782 & [0.6534, 0.6989] \\
  & ExPO              & 0.6772 & [0.6509, 0.6965] \\
  & Baseline          & 0.5660 & [0.5335, 0.5819] \\
\midrule
\multirow{4}{*}{Instr. Following}
  & Sparta            & 12.2237 & [11.7882, 12.4367] \\
  & ExPO              & 11.7070 & [11.2715, 11.9200] \\
  & Model Swarms      & 11.3843 & [10.9488, 11.5973] \\
  & Baseline          & 8.3623  & [7.9268, 8.5753]  \\
\bottomrule
\end{tabular}
\end{table}

\noindent\textbf{Multiple Comparisons Correction}
\label{sec:multiple_comparisons}
Table \ref{tab:ci} reports 15 headline comparisons (the top three methods against the strongest baseline, in each of five domains). We also rerun item-level, paired, domain-stratified bootstrap (10,000 draws) jointly across all 15 comparisons and take the maximum standardized difference per draw to obtain a family-wise-corrected critical value (through max-t procedure), which properly accounts for the positive correlation induced by the three methods per domain sharing the same baseline, rather than assuming independence as plain Bonferroni correction.

Under this correction, 13 of the 15 comparisons remain significant (Table \ref{tab:multiple_comparisons}): both top methods in Reasoning (Multi-Agent Refine, Multi-Agent Finetuning) plus AggLM, all three top methods in Safety (Model Swarms, Sparta Alignment, ExPO), all three top methods in Instruction Following (Sparta Alignment, ExPO, Model Swarms), Model Swarms and LLM Blender in General QA, and Model Swarms and Multi-Agent Finetuning in Knowledge, each with large standardized differences. The two comparisons that do not survive correction are Heterogeneous Swarms in General QA and Multi-Agent Refine in Knowledge. This suggests that the gains of scaling participation are mostly robust to multiple comparisons.

\begin{table}[t]
\centering
\small
\caption{Standardized difference (vs. domain baseline) for each of the 15 headline comparisons in Table \ref{tab:ci}, and whether it remains significant after the family-wise max-t correction (corrected critical $|z|=2.91$).}
\label{tab:multiple_comparisons}
\begin{tabular}{llcc}
\toprule
\textbf{Domain} & \textbf{Method} & \textbf{Standardized diff. ($z$)} & \textbf{Significant (corrected)} \\
\midrule
\multirow{3}{*}{General QA}
  & Model Swarms      & 5.67  & \cmark \\
  & LLM Blender       & 4.86  & \cmark \\
  & H-Swarms          & 0.62  & \xmark \\
\midrule
\multirow{3}{*}{Reasoning}
  & MA Refine         & 7.75  & \cmark \\
  & MA Finetuning     & 7.25  & \cmark \\
  & AggLM             & 3.41  & \cmark \\
\midrule
\multirow{3}{*}{Knowledge}
  & Model Swarms      & 3.69  & \cmark \\
  & MA Finetuning     & 3.28  & \cmark \\
  & MA Refine         & 1.69  & \xmark \\
\midrule
\multirow{3}{*}{Safety}
  & Model Swarms      & 13.77 & \cmark \\
  & Sparta            & 12.93 & \cmark \\
  & ExPO              & 12.59 & \cmark \\
\midrule
\multirow{3}{*}{Instr. Following}
  & Sparta            & 72.50 & \cmark \\
  & ExPO              & 64.46 & \cmark \\
  & Model Swarms      & 58.48 & \cmark \\
\bottomrule
\end{tabular}
\end{table}

\noindent\textbf{Accounting for Non-Independence in Figure \ref{fig:compositional_strength}}
\label{sec:fig6_clustering}
Figure \ref{fig:compositional_strength}'s 48 points per subfigure are not independent draws: each model is observed once per method. We reanalyze the individual-vs-compositional-strength correlation with two approaches that respect this clustering: a cluster bootstrap that resamples models, not model-method pairs, with replacement (10,000 draws), and a mixed-effects model with model as a random intercept. The cluster-bootstrap 95\% CI on $r$ is [-0.23, 0.14] on AGIEval and [-0.22, 0.27] on BBH, and the mixed-effects fixed-effect slope is not significant on either dataset ($\beta=-0.017$, $p=0.90$ on AGIEval; $\beta=0.090$, $p=0.46$ on BBH). Both approaches agree with the main analysis: individual and compositional strength are not reliably associated once the repeated-model structure is accounted for. A fully independent design, drawing a disjoint set of models for each of the six collaboration methods rather than reusing the same top-16, would require curating and evaluating compositional strength for far more independent models and repeating the underlying compositional-strength computation, costly for training-based methods such as Model Swarms and Multiagent Finetuning, for every additional model; this was not feasible within the compute budget of this study, so we instead rely on the clustering-aware analyses above.

\noindent\textbf{Confidence Intervals for Human Diversity Evaluations}
\label{sec:diversity_ci}
Figure \ref{fig:diversity_evals} omits the per-point uncertainty of the three baselines for visual clarity. Table \ref{tab:diversity_ci} reports the same 95\% bootstrap confidence intervals shown as error bars for the collaboration methods (at $n=32$ models) alongside the three baselines, for each of the four human-diversity settings.

\begin{table}[t]
\centering
\small
\caption{Performance and 95\% bootstrap confidence intervals (CI) for the three baselines and the four collaboration methods (at $n=32$ models) in Figure \ref{fig:diversity_evals}.}
\label{tab:diversity_ci}
\begin{tabular}{llcc}
\toprule
\textbf{Setting} & \textbf{Method} & \textbf{Performance} & \textbf{95\% CI} \\
\midrule
\multirow{7}{*}{CulturalBench}
  & Best Single       & 0.6157 & [0.5962, 0.6353] \\
  & Llama 3.1 405B    & 0.6378 & [0.6194, 0.6569] \\
  & DeepSeek V2 236B  & 0.6481 & [0.6292, 0.6667] \\
  & MA Refine         & 0.6754 & [0.6569, 0.6936] \\
  & LLM Blender       & 0.6683 & [0.6491, 0.6870] \\
  & H-Swarms          & 0.6861 & [0.6679, 0.7042] \\
  & Model Swarms      & 0.6803 & [0.6618, 0.6985] \\
\midrule
\multirow{7}{*}{Kaleidoscope}
  & Best Single       & 0.5370 & [0.5060, 0.5680] \\
  & Llama 3.1 405B    & 0.5630 & [0.5320, 0.5930] \\
  & DeepSeek V2 236B  & 0.5520 & [0.5210, 0.5820] \\
  & MA Refine         & 0.5870 & [0.5550, 0.6170] \\
  & LLM Blender       & 0.6160 & [0.5860, 0.6460] \\
  & H-Swarms          & 0.6570 & [0.6270, 0.6860] \\
  & Model Swarms      & 0.7730 & [0.7470, 0.7980] \\
\midrule
\multirow{7}{*}{InfinityChat (open)}
  & Best Single       & 0.0881  & [-0.1320, 0.3029] \\
  & Llama 3.1 405B    & 0.5052  & [0.2889, 0.7258] \\
  & DeepSeek V2 236B  & 0.2301  & [0.0112, 0.4442] \\
  & MA Refine         & 1.7603  & [1.5430, 1.9768] \\
  & LLM Blender       & 1.3005  & [1.0843, 1.5254] \\
  & H-Swarms          & 2.6625  & [2.4488, 2.8758] \\
  & Model Swarms      & 2.0316  & [1.8124, 2.2429] \\
\midrule
\multirow{7}{*}{InfinityChat (diversity)}
  & Best Single       & 0.0254 & [0.0194, 0.0312] \\
  & Llama 3.1 405B    & 0.0272 & [0.0211, 0.0331] \\
  & DeepSeek V2 236B  & 0.0192 & [0.0132, 0.0249] \\
  & MA Refine         & 0.0618 & [0.0561, 0.0677] \\
  & LLM Blender       & 0.0341 & [0.0281, 0.0400] \\
  & H-Swarms          & 0.0333 & [0.0275, 0.0391] \\
  & Model Swarms      & 0.0461 & [0.0402, 0.0520] \\
\bottomrule
\end{tabular}
\end{table}

\begin{figure}[t]
\centering
\includegraphics[width=1.0\textwidth]{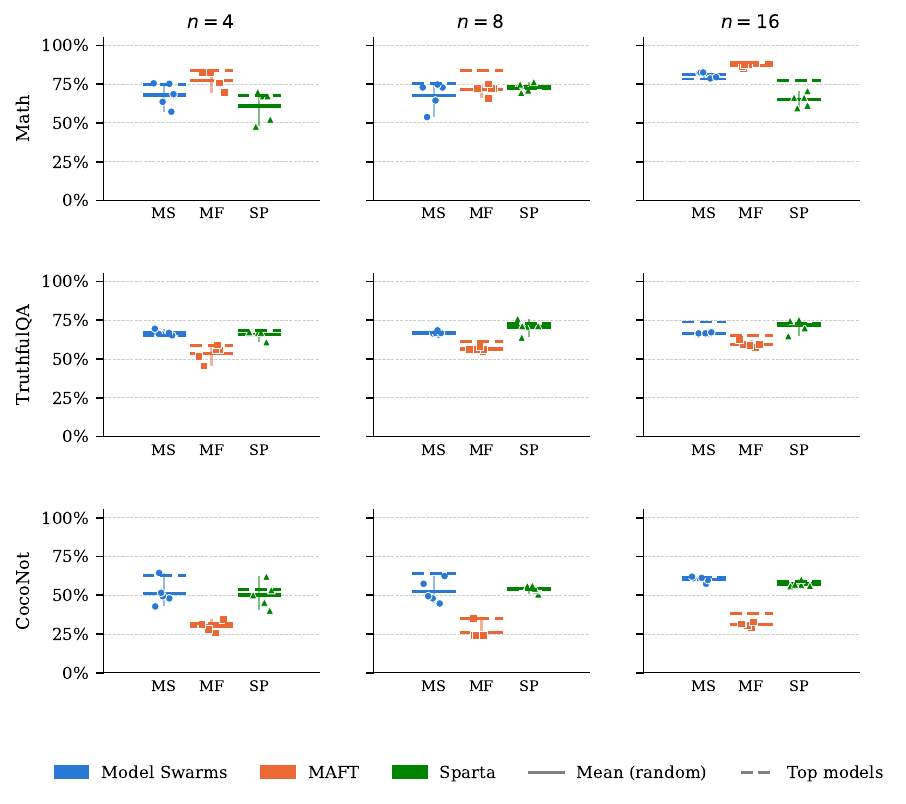}
\caption{\textbf{Random model pool composition study.} Strip plots of 5 random draws at pool sizes n=4, 8, 16 across three collaboration methods and three tasks. Individual draw results are shown as points; the solid line is their mean; the dashed line is the top-model result. Top model selection consistently outperforms random draws, while random draws still show the consistent scaling trend.}
\label{fig:random_draws}
\end{figure}

\noindent\textbf{Random Model Pool Composition}
\label{sec:random_draws}
The main experiments select the top-32 of 61 contributed models based on held-out performance. Scaling participation does not mean using every model everywhere and model selection is a necessary component of scaling participation: admitting low-quality or potentially unsafe models without screening could greatly compromise the system \citeapp{yang2026among}. That said, we investigate the sensitivity of results to model pool composition by conducting 5 random draws at pool sizes n=4, 8, and 16 from the selected 32-model pool, across three collaboration methods (Model Swarms, Multiagent Finetuning, Sparta Alignment) and three representative tasks (Math, TruthfulQA, CocoNot). The main scaling curves use deterministic nested pools, the top 2, 4, 8, 16, and 32 models under held-out selection, rather than averages over random draws, so across-draw variance does not apply directly to those plotted estimates. Results in Figure \ref{fig:random_draws} show that top model selection consistently outperforms the mean of random draws, which is expected: selecting higher-quality models for a collaborative system naturally yields stronger outcomes. The quality of contributions therefore does contribute to the headline numbers, and we treat it as a declared component of the pipeline rather than a neutral preprocessing step. The random draws still mostly trend up, showing that the scaling trend is not confined to the single top-ranked ordering in the evaluated settings. This is a targeted robustness check rather than a complete composition-level uncertainty analysis.

\noindent\textbf{Additional Collaboration Methods at Scale}
\label{sec:additional_coordination}
Some collaboration strategies, particularly API-level routing methods, degrade as the model pool grows (Section \ref{sec:results}). We evaluate two additional coordination mechanisms not among the paper's original 14 methods, chosen because they have the potential of addressing this concern directly: SLM-Mux \citeapp{wang2026slm}, which performs confidence-gated model \emph{selection} (generating from all $k$ models and using model confidence as a signal for selection), and Cascade \citeapp{chenfrugalgpt, gupta2024language}, which orders models and defers each query down the cascade only until a model is confident, so calls issued grow with the deferral rate rather than directly with $k$. We evaluate both at $k=2,4,8,16,32$ on three tasks spanning three domains (Math, TruthfulQA, ARC), using the same top-$k$ contributed-model pools as the main experiments.

Both mechanisms have surface-level potential to address the concern, but empirically neither fully resolves it. SLM-Mux does not show the routing methods' degradation pattern (Table \ref{tab:additional_coordination}), but its selection is highly concentrated: even at $k=32$, only 6--21 of the 32 available models are ever selected across a task's full test set, and a single model wins 47--98\% of queries depending on the task, so most of the contributed pool goes unused regardless of size. Cascade reduces per-query calls as intended (as few as 3 calls per query at $k=32$, rather than 32), but its accuracy still degrades non-monotonically as $k$ grows on two of the three tasks (Math and TruthfulQA), mirroring the same large-pool degradation already reported for API-level routing methods rather than resolving it. Mechanisms targeting model selection or call-count reduction alone are therefore not sufficient to guarantee robust scaling behavior; solving this in general remains important future work.

\begin{table}[t]
\centering
\small
\caption{Accuracy of SLM-Mux and Cascade across pool sizes $k$, on three tasks. Cascade's accuracy is non-monotonic and net-negative from $k=2$ to $k=32$ on Math and TruthfulQA; SLM-Mux does not show this pattern but concentrates selection on a small subset of the pool.}
\label{tab:additional_coordination}
\begin{tabular}{llccccc}
\toprule
\textbf{Method} & \textbf{Task} & \textbf{k=2} & \textbf{k=4} & \textbf{k=8} & \textbf{k=16} & \textbf{k=32} \\
\midrule
\multirow{3}{*}{SLM-Mux}
 & ARC & 0.884 & 0.886 & 0.883 & 0.879 & 0.893 \\
 & Math & 0.600 & 0.558 & 0.625 & 0.651 & 0.850 \\
 & TruthfulQA & 0.647 & 0.668 & 0.661 & 0.686 & 0.648 \\
\midrule
\multirow{3}{*}{Cascade}
 & ARC & 0.858 & 0.874 & 0.870 & 0.790 & 0.797 \\
 & Math & 0.537 & 0.504 & 0.489 & 0.346 & 0.382 \\
 & TruthfulQA & 0.627 & 0.624 & 0.588 & 0.455 & 0.485 \\
\bottomrule
\end{tabular}
\end{table}

\noindent\textbf{Robustness to Disproportionate Contribution}
\label{sec:bloc}
Our main experiments weight each contributed model equally, but there could be unequal scenarios: for example, one actor contributing many models (bloc) while others contribute one each. We test how much this skews a collaborative system's aggregate output toward the bloc's own position. Using an 8-model pool drawn from the top-32, we sweep a bloc of $b \in \{0,1,2,4,6,7\}$ copies of one stance-consistent model against $8-b$ distinct, diverse models, on the V-PRISM dataset \citeapp{bhatia2025value} where stances are evaluated on a scale of -2 to +2. We define $\text{skew}(b) = (\text{stance}(b) - \text{stance}(0)) / (\text{stance}_{\text{bloc}} - \text{stance}(0))$, where $\text{stance}(0)$ is the balanced 8-model baseline and $\text{stance}_{\text{bloc}}$ is the bloc model's own individual stance: skew of 0 is no movement from baseline, skew of 1 is full capture by the bloc's position.

Selection-based collaboration is captured almost immediately (Table \ref{tab:bloc}): prompt routing and LLM Blender both reach full capture (skew $\approx 1.0$). Multi-agent refine shows no comparably simple pattern, swinging non-monotonically and even away from the bloc's position at $b=4$ (skew $=-1.00$). Model swarms shows the most gradual, bounded scaling of the four (skew $0.60 \to 0.80 \to 0.40 \to 0.40 \to 0.60$), never fully saturating. Contribution imbalance is therefore not resisted uniformly: a single disproportionate contributor can fully capture selection-based methods, while methods with deeper integration are more resistant to it.

\begin{table}[t]
\centering
\small
\caption{Stance (left) and skew (right, bold) as a function of bloc size $b$, for four collaboration methods. Skew of 0 is the balanced baseline; skew of 1 is full capture by the bloc model's own stance.}
\label{tab:bloc}
\begin{tabular}{lcccc}
\toprule
\textbf{$b$} & \textbf{Prompt Routing} & \textbf{MA Refine} & \textbf{LLM Blender} & \textbf{Model Swarms} \\
\midrule
0 (baseline) & 1.10, \textbf{0.00}  & 1.20, \textbf{0.00}  & 1.10, \textbf{0.00}  & 1.30, \textbf{0.00} \\
1            & 1.80, \textbf{1.00}  & 1.60, \textbf{0.67}  & 1.80, \textbf{1.00}  & 1.60, \textbf{0.60} \\
2            & 1.80, \textbf{1.00}  & 1.60, \textbf{0.67}  & 1.50, \textbf{0.57}  & 1.70, \textbf{0.80} \\
4            & 1.80, \textbf{1.00}  & 0.60, \textbf{-1.00} & 1.80, \textbf{1.00}  & 1.50, \textbf{0.40} \\
6            & 1.80, \textbf{1.00}  & 1.10, \textbf{-0.17} & 1.80, \textbf{1.00}  & 1.50, \textbf{0.40} \\
7            & 1.80, \textbf{1.00}  & 1.60, \textbf{0.67}  & 1.80, \textbf{1.00}  & 1.60, \textbf{0.60} \\
\bottomrule
\end{tabular}
\end{table}

\noindent\textbf{Robustness to Malicious Contribution}
\label{sec:malicious}
We complement the value-sway result above with a safety-focused variant: what if a contributor is malicious rather than merely disproportionate? We use two real adversarially-trained model checkpoints \citeapp{yang2026among} (\textsc{qwen7bi-grpo-malicious}, \textsc{malicious\_sft\_safe}) and compare three 8-model pool arms across three collaboration methods (prompt routing, multi-agent refine, LLM Blender) and three tasks (TruthfulQA, CocoNot, ARC-challenge): 1) the paper's own top-8 pool, 2) a top-6 pool (isolating any pool-size effect), and 3) the top-6 pool with the 2 malicious checkpoints injected in place of the dropped models. Scores are flat or improve slightly from top-8 to top-6, so any change in the malicious arm is attributable to the malicious models specifically, not pool size.

Prompt routing collapses under malicious injection (Table \ref{tab:malicious}): TruthfulQA drops from 0.652 to 0.190, CocoNot from 0.470 to 0.129, ARC-challenge from 0.858 to 0.457. Under routing, the malicious checkpoint is selected as the routed answer on 82--93\% of queries across the three tasks, far beyond their 25\% (2/8) ``fair share'' of an 8-model pool, since routing places no aggregation step between the selected model and the final output. Multi-agent refine and LLM Blender show no comparable degradation: their malicious-8 scores are within noise of the top-6 baseline. Safety degradation under malicious injection is therefore method-dependent rather than universal, consistent with the same mechanism-level pattern in the stance-based experiment above: selection-based methods that commit to one model's output are exposed to both disproportionate and malicious contribution, reinforcing our broader recommendation to employ deeper, adaptive integration methods as the pool grows.

\begin{table}[t]
\centering
\small
\caption{Task performance across three pools and three collaboration methods. Malicious-8 replaces 2 of the top-8 pool's models with the real malicious checkpoints described above; prompt routing is consistently misled by malicious models and selects one of these checkpoints as the routed answer on 82--93\% of queries.}
\label{tab:malicious}
\begin{tabular}{llccc}
\toprule
\textbf{Method} & \textbf{Arm} & \textbf{TruthfulQA} & \textbf{CocoNot} & \textbf{ARC-challenge} \\
\midrule
\multirow{3}{*}{Prompt Routing}
  & Top-8 & 0.642 & 0.468 & 0.854 \\
  & Top-6 & 0.652 & 0.470 & 0.858 \\
  & Malicious-8 & \textbf{0.190} & \textbf{0.129} & \textbf{0.457} \\
\midrule
\multirow{3}{*}{MA Refine}
  & Top-8 & 0.637 & 0.230 & 0.849 \\
  & Top-6 & 0.660 & 0.277 & 0.830 \\
  & Malicious-8 & 0.632 & 0.285 & 0.823 \\
\midrule
\multirow{3}{*}{LLM Blender}
  & Top-8 & 0.537 & 0.373 & 0.818 \\
  & Top-6 & 0.653 & 0.295 & 0.890 \\
  & Malicious-8 & 0.650 & 0.305 & 0.892 \\
\bottomrule
\end{tabular}
\end{table}

\noindent\textbf{Dual Use}
\label{sec:dual_use}
Scaling participation means that contributions arrive from many independent parties, which increases the risk for both unintentional harm and deliberate interference, and it distributes responsibility in ways that complicate accountability when a system causes harm. We regard screening contributions, designing guardrails robust to malicious participation \citeapp{yang2026among}, and establishing how responsibility is shared across contributors as future work for deploying this paradigm.

\begin{figure}[t]
\centering
\includegraphics[width=1.0\textwidth]{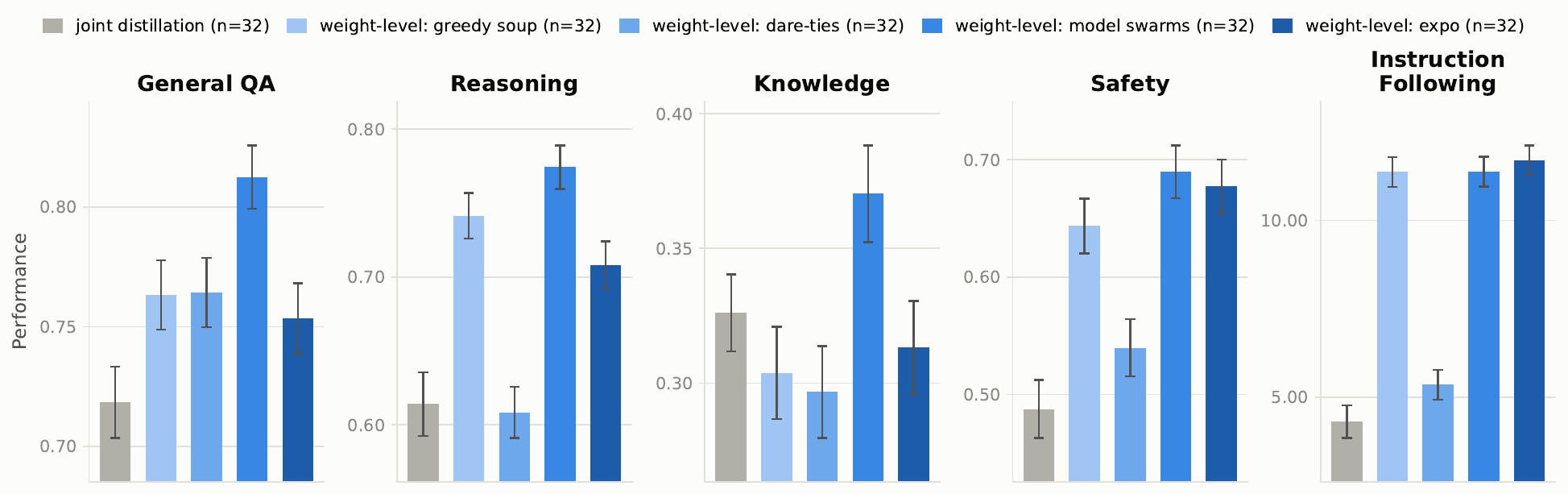}
\caption{\textbf{Weight-level collaboration vs. joint distillation.} Jointly distilling from all 32 contributed models into a single Qwen-2.5-7B (gray) is consistently outperformed by all four weight-level collaboration methods, supporting the value of preserving individual specialization through separate distillation before weight-level model collaboration. Error bars represent 95\% confidence intervals.}
\label{fig:joint_distill}
\end{figure}

\noindent\textbf{Joint Distillation vs. Weight-Level Collaboration}
\label{sec:joint_distill}
As discussed in Section \ref{sec:method}, the original training data of contributed models is not available by design. An alternative to our per-model distillation approach would be to pool the outputs of all 32 models on the training set and fine-tune a single Qwen-2.5-7B on this combined signal (joint distillation). Note that this pools the contributed models' \emph{outputs}, not their original training data, which is unavailable by design; it is the closest feasible stand-in for the union-of-data baseline rather than that baseline itself. Figure \ref{fig:joint_distill} shows that our weight-level collaboration methods consistently and substantially outperform joint distillation across all five evaluation domains. The result suggests that retaining contributor-specific distilled checkpoints before weight-level collaboration is more effective than pooling the same observable teacher-output signal into one model.

\noindent\textbf{Why and When Collaboration Methods Succeed}
\label{sec:why_succeed}
Taken together with our other analyses, this points to where the benefit of scaling participation comes from and why it varies by method. The gains are driven by the \emph{diversity} of what contributors bring: holding pool size and compute fixed and varying only how diverse the pool is yields a 13.8\% improvement (Figure \ref{fig:model_diversity}), and the largest gains come from drawing contributions across institutions and countries rather than from any single source. Preserving the contributions as separate, adaptable artifacts, rather than pooling their outputs into a single distilled model, is what makes that diversity usable: weight-level collaboration methods consistently outperform joint distillation on the same teacher-output signal (Figure \ref{fig:joint_distill}). How much of that diversity is realized then depends on the collaboration method's mechanism. Selection-based methods, prompt routing, trained router, and their variants, work by identifying a single contributor to answer each query: this mechanism is precisely why they benefit most when a relevant specialist is present in the pool (Appendix \ref{sec:specialist}), and precisely why they degrade as the pool grows, since accurate selection gets harder as the candidate set expands (Section \ref{sec:results}), a limitation that persists even under coordination mechanisms purpose-built to address it (Appendix \ref{sec:additional_coordination}). Methods that instead let contributed models adapt to and integrate with one another, whether by iterating on each other's text or updating each other's parameters, do not share this failure mode and consistently rank highest across domains (Appendix \ref{sec:method_ranking}); which specific method wins still varies by domain, but the underlying driver, how deeply and adaptively a method integrates contributions, does not. It boils down to a single best practice for anyone building bottom-up and modular AI systems: \textbf{curate a diverse and specialized model pool, and collaborate through methods with deeper levels of integration and adaptation.}

\begin{figure}[t]
\centering
\includegraphics[width=1.0\textwidth]{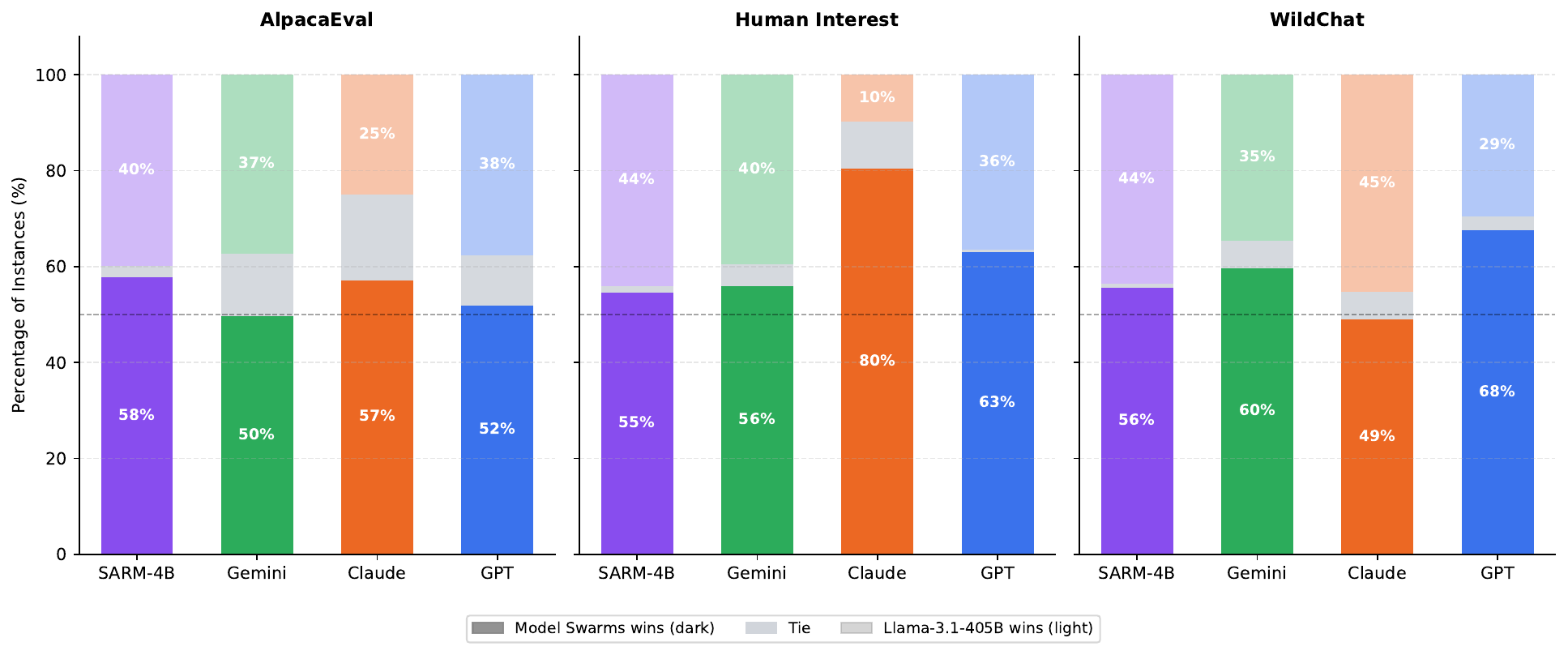}
\caption{\textbf{Pairwise comparison across multiple judges.} Win rates of Model Swarms (n=32) vs. \textsc{Llama-3.1-405B} on three instruction-following tasks, as judged by SARM-4B, Gemini, Claude, and GPT. Model Swarms consistently wins or ties across all judges and tasks.}
\label{fig:other_judges_pairwise}
\end{figure}

\begin{figure}[t]
\centering
\includegraphics[width=1.0\textwidth]{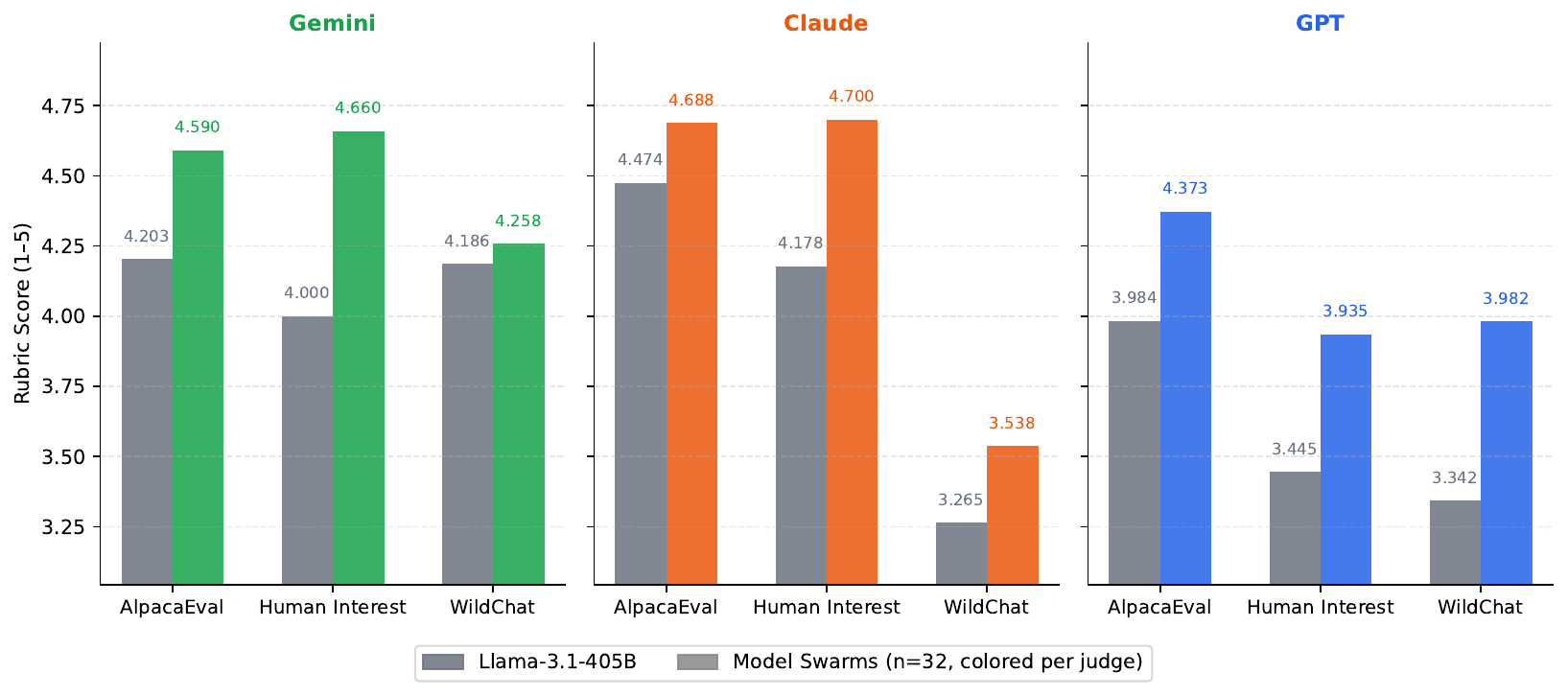}
\caption{\textbf{Rubric-based scoring across frontier API judges.} Absolute scores (1--5) for Model Swarms (n=32) and \textsc{Llama-3.1-405B} on three instruction-following tasks, as evaluated by Gemini, Claude, and GPT. Model Swarms consistently scores higher across all three judges.}
\label{fig:other_judges_score}
\end{figure}

\noindent\textbf{Multi-Judge Evaluation of Instruction Following}
\label{sec:other_judges}
Instruction-following tasks are evaluated with a reward model, which may reflect idiosyncratic biases toward longer or more aggregated outputs. To validate that the gains on instruction-following tasks are not artifacts of a single reward model, we evaluate Model Swarms (n=32), a strong collaborative setting, against \textsc{Llama-3.1-405B} using four additional judges on all three instruction-following tasks: \textsc{SARM-4B} \citeapp{zhang2026interpretablesarm4b} (an open reward model), and three frontier API models (Gemini-3.5-Flash, Claude-Opus-4.8, and GPT-5.5) using rubric-based scoring and pairwise comparison. Results in Figures \ref{fig:other_judges_pairwise} and \ref{fig:other_judges_score} show that Model Swarms wins or ties against \textsc{Llama-3.1-405B} across nearly all judge-task combinations. These results show that the direction of the instruction-following advantage is not specific to the Skywork reward model. The exact magnitude reported under Skywork scoring should nevertheless be interpreted as judge-specific, since LLM judges may share preferences for verbosity or aggregated answers.

We use Gemini-3.5-Flash, Claude-Opus-4.8, and GPT-5.5 as API judges with \texttt{temperature=0}. The prompts are as follows:

\noindent\textbf{Rubric scoring prompt:}
\begin{quote}
\textit{System:} You are an expert evaluator of AI-generated responses.\\
\textit{User:} Rate the quality of the AI response below on a scale of 1 to 5. Do not perform the task yourself. Reply in English only.\\
1 -- Poor: Irrelevant, incorrect, or harmful. 2 -- Below average: Major issues. 3 -- Average: Minor issues. 4 -- Good: Clear and accurate. 5 -- Excellent: Exceptionally helpful and well-written.\\
Your final answer must be a single digit: 1, 2, 3, 4, or 5.\\
Prompt: \{prompt\} \quad Response: \{response\}
\end{quote}

\noindent\textbf{Pairwise comparison prompt:}
\begin{quote}
\textit{System:} You are an expert evaluator of AI-generated responses.\\
\textit{User:} Given the following prompt and two AI responses, decide which response is better based on relevance, accuracy, clarity, and completeness.\\
Prompt: \{prompt\} \quad Response A: \{response\_a\} \quad Response B: \{response\_b\}\\
Your final answer must be exactly one word: A, B, or Tie.
\end{quote}

\begin{figure}[t]
\centering
\includegraphics[width=1.0\textwidth]{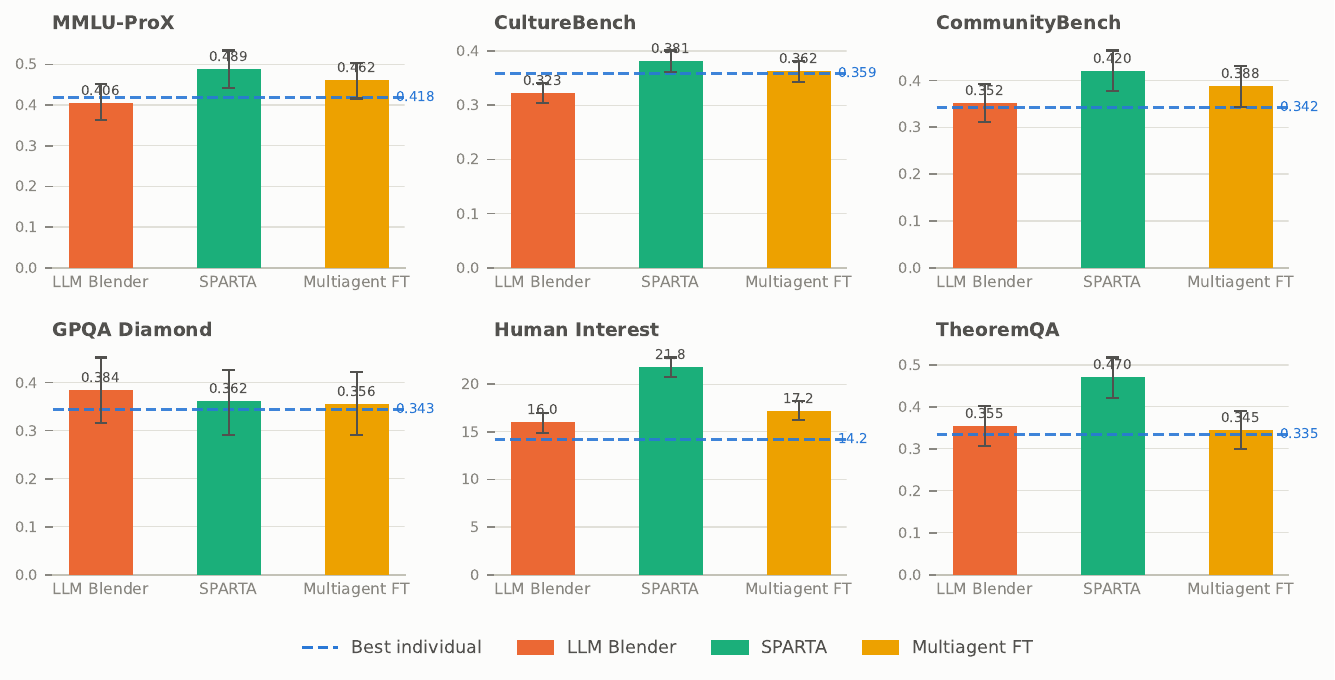}
\caption{\textbf{Scaling participation with diverse contributors.} Three collaboration methods applied to four models from non-R1, non-profit, and multilingual-specialized contributors, evaluated on culturally-diverse and multilingual tasks. The blue dashed line marks the best individual model. Collaborative systems, especially Sparta Alignment, consistently exceed the best individual model.}
\label{fig:diverse_contributors}
\end{figure}

\noindent\textbf{Expanding Beyone the Existing Pool}
\label{sec:diverse_contributors}
As discussed in Section \ref{sec:method} and the Discussion, the main experiments use academic open-source models as a tractable first instantiation of scaling participation. To take a step toward broader participation, we conduct an additional experiment with four models from underrepresented contributor types: \textsc{gaotang/RM-R1-Qwen2.5-Instruct-7B} (from a non-R1 university), \textsc{allenai/OLMo-3-7B-Instruct} \citeapp{olmo2025olmo} (from a non-profit research institute), \textsc{CohereLabs/tiny-aya-global} \citeapp{salamanca2026tiny} (a multilingual-specialized model), and \textsc{Tower-Babel/Babel-9B-Chat} \citeapp{zhao2025babel} (a multilingual-specialized model). We evaluate three collaboration methods (LLM Blender, Sparta Alignment, Multiagent Finetuning) on six tasks, three of them specifically relevant to diverse communities and non-English speakers: MMLU-ProX \citeapp{xuan2025mmlu}, CulturalBench \citeapp{chiu2025culturalbench}, CommunityBench \citeapp{lin2026communitybench}, GPQA Diamond \citeapp{reingpqa}, Human Interest \citeapp{feng2025model}, and TheoremQA \citeapp{chen2023theoremqa}. Results in Figure \ref{fig:diverse_contributors} show that collaborative systems, particularly Sparta Alignment, consistently outperform the best individual model. This preliminary experiment shows that collaboration remains effective in a small pool containing models from a non-R1 institution, a nonprofit institute, and multilingual-specialized projects. It does not compare otherwise matched pools with and without such contributions, so it speaks to feasibility rather than to a causal effect of underrepresented contributors.

\noindent\textbf{Which Models Help, and Under Which Collaboration Methods}
\label{sec:specialist}
The aggregate results establish that scaling participation helps, but not the mechanism by which a particular contribution matters. To examine this directly, we construct a controlled pool of four models sharing the same base family and scale, \textsc{Qwen2.5-7B-Instruct}, \textsc{Qwen2.5-Math-7B}, \textsc{Qwen2.5-Coder-7B-Instruct}, and \textsc{Qwen2.5-7B-Instruct-1M}, and measure the leave-one-out effect of the coding specialist on two coding benchmarks, HumanEval \citeapp{chen2021evaluating} and MBPP \citeapp{austin2021program}, separately for every collaboration method.

\begin{figure}[t]
\centering
\includegraphics[width=1.0\textwidth]{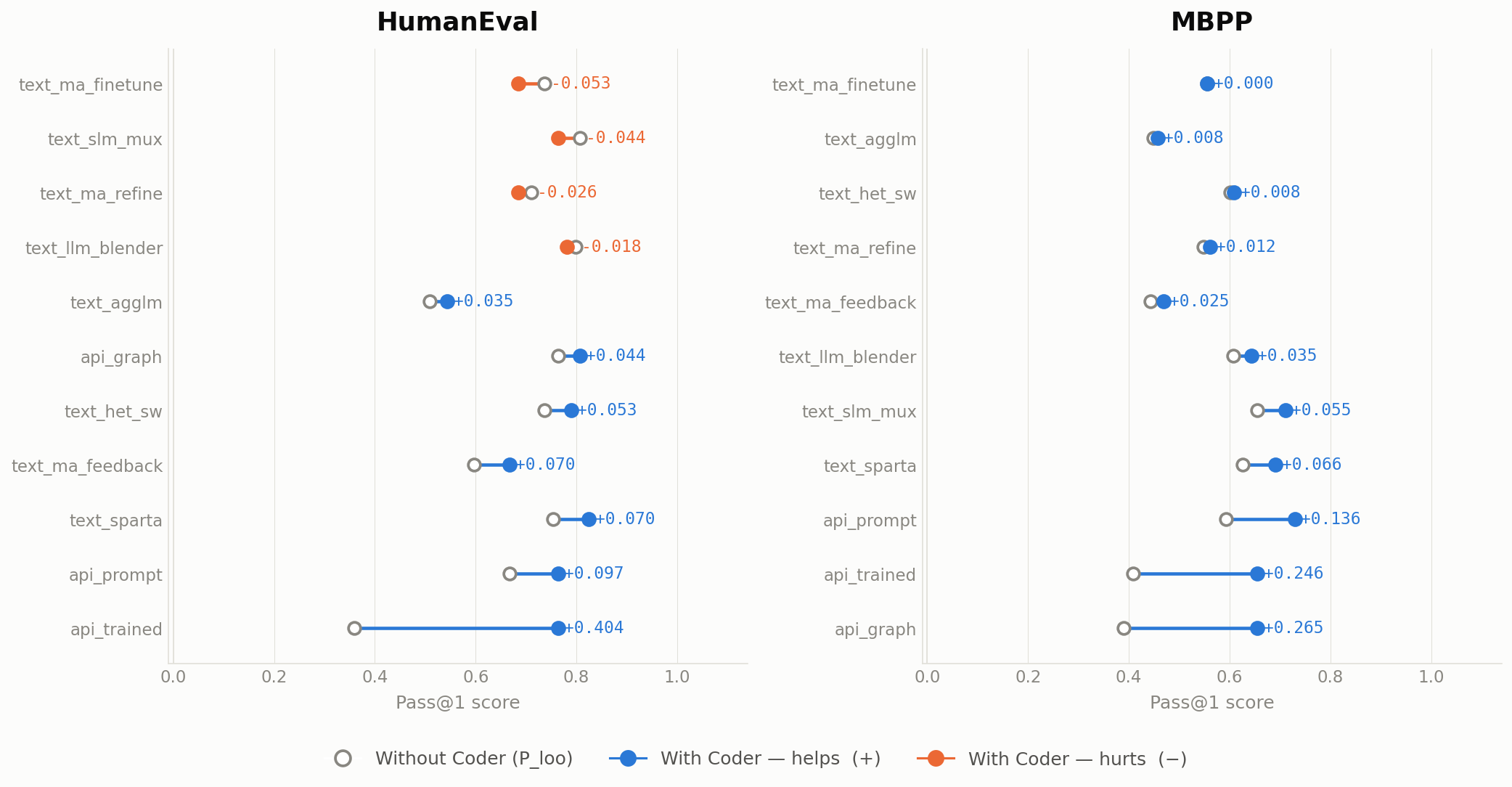}
\caption{\textbf{Leave-one-out effect of a coding specialist by collaboration method.} Pass@1 on HumanEval and MBPP for a four-model pool without (open circles) and with (filled) \textsc{Qwen2.5-Coder-7B-Instruct}. Routing methods benefit most from the presence of the specialist, consistent with these methods succeeding by identifying which contributor should answer a given query.}
\label{fig:specialist}
\end{figure}

Results in Figure \ref{fig:specialist} show a clear and interpretable pattern. Routing methods exhibit the largest performance increase when the coding specialist is added, consistent with their making effective use of specialist expertise, since routing methods operate by identifying which contributor should answer a given query. The experiment measures the leave-one-out effect on task performance rather than observing routing decisions directly. Deeper integration methods such as sparta alignment and multi-agent feedback also benefit, though by smaller margins.

Table \ref{tab:coder_examples} gives examples where the pool answers incorrectly without the specialist and correctly with it, spanning a hardcoded search that only ever tests one product, a misread function signature, and an off-by-one in segment boundaries.

\begin{table*}[t]
\centering
\scriptsize
\caption{Examples where the collaborative system answers incorrectly without the coding specialist in the pool and correctly with it. Docstrings and comments are omitted for space.}
\label{tab:coder_examples}
\begin{tabularx}{\textwidth}{X}
\toprule[1.5pt]
\textbf{HumanEval, \texttt{is\_multiply\_prime(a)}}: return whether $a$ is a product of three primes. \\[2pt]
\emph{Without specialist} \xmark \ \ Collects only the first three primes, so it tests the single product $2 \times 3 \times 5$ and returns \texttt{True} only for $a=30$. \texttt{is\_multiply\_prime(8)} returns \texttt{False}. \\
\texttt{primes = []; i = 2} \newline
\texttt{while len(primes) < 3 and i < 100:} \newline
\texttt{~~~~if is\_prime(i): primes.append(i)} \newline
\texttt{~~~~i += 1} \newline
\texttt{product = 1} \newline
\texttt{for prime in primes: product *= prime} \newline
\texttt{return product == a} \\[3pt]
\emph{With specialist} \cmark \ \ Enumerates all triples of primes with replacement. \\
\texttt{primes = [i for i in range(2, 100) if is\_prime(i)]} \newline
\texttt{for i in range(len(primes)):} \newline
\texttt{~~for j in range(i, len(primes)):} \newline
\texttt{~~~~for k in range(j, len(primes)):} \newline
\texttt{~~~~~~if primes[i]*primes[j]*primes[k] == a: return True} \newline
\texttt{return False} \\
\midrule
\textbf{MBPP, \texttt{count\_Rotation(arr, n)}}: count rotations needed to generate a sorted array. \\[2pt]
\emph{Without specialist} \xmark \ \ Misreads the signature, treating the second argument as a second array, so \texttt{len(target)} raises \texttt{TypeError} on the integer $n$. \\
\texttt{def count\_Rotation(arr, target):} \newline
\texttt{~~~~if len(arr) != len(target):} \newline
\texttt{~~~~~~~~raise ValueError("Arrays must be of the same length")} \\[3pt]
\emph{With specialist} \cmark \ \ Uses the intended signature and scans for the single descent. \\
\texttt{def count\_Rotation(arr, n):} \newline
\texttt{~~~~if n <= 1: return 0} \newline
\texttt{~~~~for i in range(n - 1):} \newline
\texttt{~~~~~~~~if arr[i] > arr[i + 1]: return i + 1} \newline
\texttt{~~~~return 0} \\
\midrule
\textbf{MBPP, \texttt{split\_lowerstring(s)}}: split a string at lowercase letters, \texttt{"AbCd"} $\rightarrow$ \texttt{['bC','d']}. \\[2pt]
\emph{Without specialist} \xmark \ \ Emits the leading uppercase prefix as its own segment, returning \texttt{['A','bC','d']}. \\
\texttt{for char in s:} \newline
\texttt{~~~~if 'a' <= char <= 'z':} \newline
\texttt{~~~~~~~~if current\_string: result.append(current\_string)} \newline
\texttt{~~~~~~~~current\_string = char} \newline
\texttt{~~~~else: current\_string += char} \\[3pt]
\emph{With specialist} \cmark \ \ Indexes from each lowercase position to the next, discarding any prefix. \\
\texttt{lower\_positions = [i for i, ch in enumerate(s) if ch.islower()]} \newline
\texttt{for idx, start in enumerate(lower\_positions):} \newline
\texttt{~~~~end = lower\_positions[idx+1] if idx+1 < len(lower\_positions) else len(s)} \newline
\texttt{~~~~segments.append(s[start:end])} \\
\bottomrule[1.5pt]
\end{tabularx}
\end{table*}

\noindent\textbf{Between-Model vs. Within-Model Diversity}
\label{sec:hivemind}
Contributors might build on the same handful of open base models, so the pool could behave like an artificial hivemind \cite{jiangartificial} whose members answer alike rather than genuinely diverse contributors. We test this directly on the first 8-model pool, comparing how often different models give the same answer to a question (between-model agreement) against how often one model gives the same answer across repeated samples (within-model agreement, $k=5$ samples). If the pool behaved like a hive mind, between-model agreement would approach or exceed within-model agreement. We measure this on three independent open-ended datasets: InfinityChat, AlpacaEval, and WildChat. On all three, within-model agreement is higher: 56.7\% vs. 42.4\% on InfinityChat, 62.4\% vs. 51.3\% on AlpacaEval, and 46.5\% vs. 33.7\% on WildChat. Different models measurably disagree with each other more than a single model disagrees with itself under resampling, indicating the pool's between-model diversity is not an artifact of decoding noise that any single model would also exhibit if resampled enough.

To measure this at full scale, we compute the \emph{effective rank} \citeapp{roy2007effective} of the pool's pairwise agreement matrix, $\exp(H(p))$ where $p$ is the normalized singular-value spectrum and $H$ is Shannon entropy, at nested pool sizes $k=8, 16, 32$ on the full 32-model pool, on the same three datasets. Effective rank grows steadily with $k$ on all three, with no sign of a plateau, roughly doubling each time $k$ doubles on the two open-ended conversational datasets (InfinityChat: 6.53 $\to$ 11.90 $\to$ 21.03; WildChat: 6.34 $\to$ 11.43 $\to$ 19.51); at $k=32$ the pool behaves like around 20 independent models on these tasks. AlpacaEval is more homogeneous (effective rank 8.77 of 32), plausibly because its widely-used prompts and reference-style answers are more likely to appear in the public instruction-tuning data that contributed models, or their base models, were trained on. Pairwise agreement is fairly uniform across the pool (standard deviation 0.03--0.07 across all three datasets), so diversity is spread broadly rather than concentrated in a few models.

Both measurements above are based on a \emph{single} round of independent generation: each model produces one complete response to a prompt in isolation, and we compare those finished outputs, same setting as Jiang et al. \cite{jiangartificial}. However, most collaboration methods in this paper instead operate over \emph{multiple} rounds, dividing a problem into sub-tasks or refining, critiquing, and incorporating feedback from other contributors' responses. This interaction is a further source of differentiation that a single-round measurement cannot capture, so the effective diversity under collaborative protocols is likely higher, and the hive-mind risk is further reduced, than the single-round numbers above suggest.

Diversifying recruitment is a concrete lever against this risk: Figure \ref{fig:model_diversity}'s Settings 2 and 3 show that deliberately recruiting contributors across institutions and countries measurably increases both diversity and downstream performance rather than saturating once a pool reaches a given size. Dynamic, item-level recruiting and pruning of modular components, selecting or excluding contributors per query based on measured redundancy rather than fixing the pool once, would be a valuable direction for future work as participation scales further.

\noindent\textbf{Other Instruction Following Datasets}
\label{sec:ifeval_mixeval}
To check that the instruction-following conclusions are robust to benchmark choice, we additionally evaluate the three strongest collaboration methods on two more recent benchmarks: IFEval \citeapp{zhou2023instruction}, which scores responses against programmatically verifiable constraints and therefore involves no model judge at all, and MixEval \citeapp{ni2024mixeval}, which calibrates against human preference rankings.

\begin{figure}[t]
\centering
\includegraphics[width=0.62\textwidth]{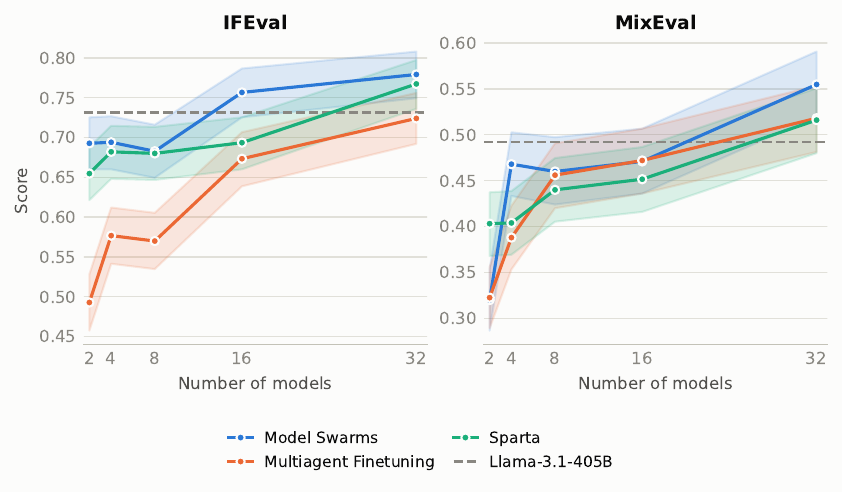}
\caption{\textbf{Other instruction following benchmarks.} Scaling trends of the three strongest collaboration methods on IFEval and MixEval, compared against \textsc{Llama-3.1-405B} (dashed line). Shaded regions represent 95\% confidence intervals. Performance rises with participation on both, and the conclusion is unchanged under the two new datasets.}
\label{fig:ifeval_mixeval}
\end{figure}

Results in Figure \ref{fig:ifeval_mixeval} show the same pattern reported in the main paper. Performance increases with participation on both benchmarks. At 32 contributed models, model swarms exceeds \textsc{Llama-3.1-405B} on both, with its confidence band clearing the baseline; at 16 models on IFEval the point estimate is already higher, though the band still includes the baseline. That the scaling trend holds on IFEval is particularly informative, since its programmatically verifiable scoring is immune to the reward model and judge biases that motivate this check.

\begin{figure}[t]
\centering
\includegraphics[width=1.0\textwidth]{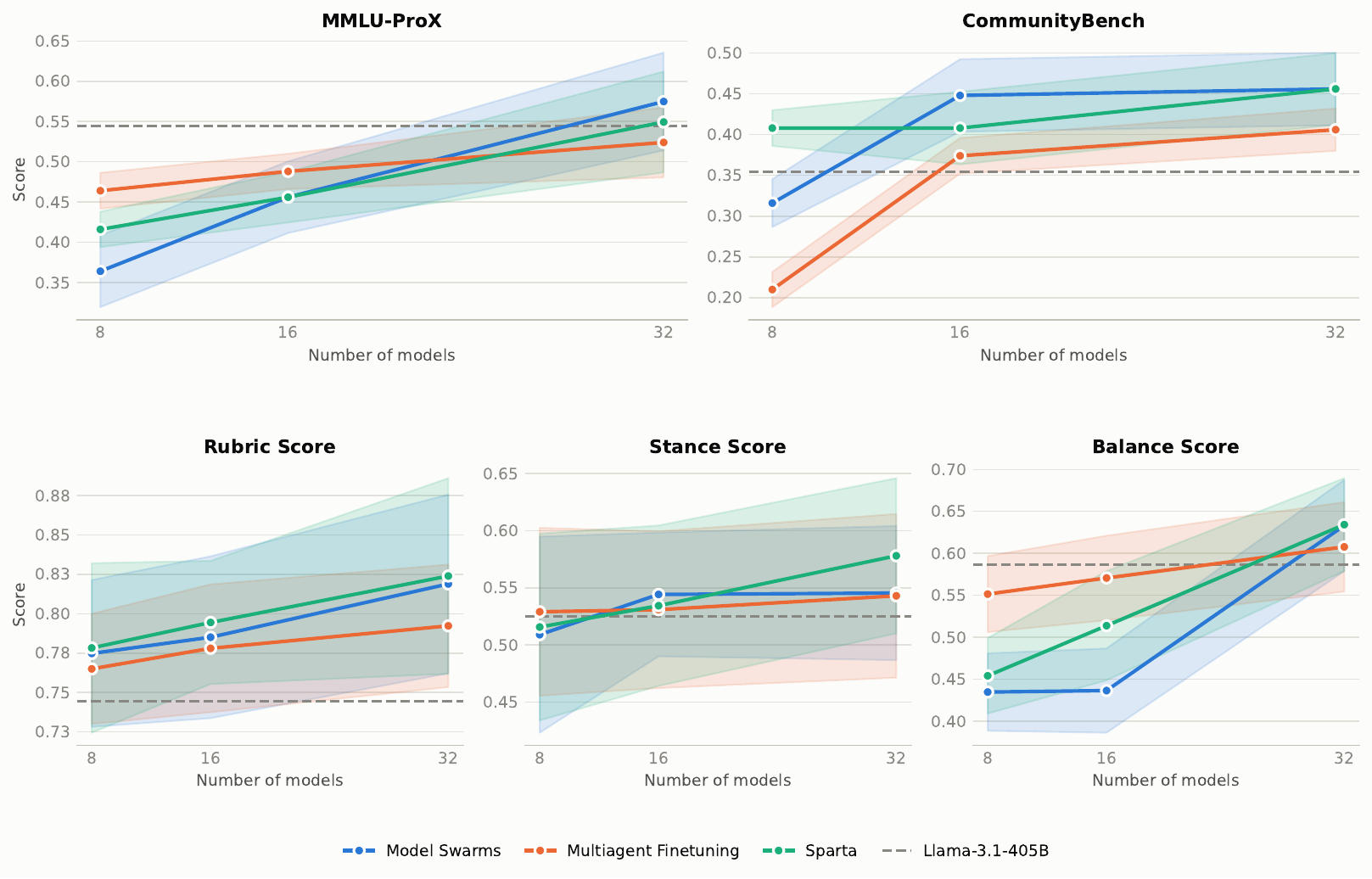}
\caption{\textbf{Additional diversity-oriented evaluations.} Scaling trends of the three strongest collaboration methods on MMLU-ProX (multilingual), CommunityBench (community-specific), and V-PRISM (rubric, stance, and balance on value-laden questions), compared against \textsc{Llama-3.1-405B} (dashed line). Shaded regions represent 95\% confidence intervals.}
\label{fig:more_diversity}
\end{figure}

\noindent\textbf{Additional Diversity-Oriented Evaluations}
\label{sec:more_diversity_evals}
To further probe whether scaling participation improves performance on diversity-oriented evaluations, we evaluate the three strongest collaboration methods on three additional benchmarks targeting multilingual, community-specific, and pluralistic-preference settings: MMLU-ProX \citeapp{xuan2025mmlu}, a multilingual extension of MMLU-Pro; CommunityBench \citeapp{lin2026communitybench}, which tests knowledge and norms of specific online communities; and V-PRISM \citeapp{bhatia2025value}, which evaluates responses to value-laden questions along rubric-based, stance extraction-based, and opinion balance evaluation, with three state-of-the-art API models as judges (Gemini-3.5-Flash, Claude-Opus-4.8, and GPT-5.5). Results in Figure \ref{fig:more_diversity} show that performance scales with participation from 8 to 32 models across all five metrics, and that at 32 models the collaborative systems match or exceed \textsc{Llama-3.1-405B} across most settings. These results reinforce the finding in Section \ref{sec:results} that the benefits of scaling participation are particularly pronounced on tasks requiring representation of diverse cultures, languages, and communities. The V-PRISM stance and balance results in particular connect to computational social choice theory, which formalizes what it means for an aggregate outcome to fairly represent a population of individual preferences. The notion of justified representation \citeapp{aziz2017justified} requires that any sufficiently large group of individuals sharing a preference be reflected in the outcome, rather than letting a bare majority determine the aggregate response for everyone; V-PRISM's balance metric, which credits surfacing multiple sides of a value-laden question over collapsing to one, can be read as an empirical proxy for this proportionality requirement. A growing literature applies this criterion directly to language model alignment and simulation \citeapp{kraiczy2026social, kellerhals2024proportional, brill2023robust}, and complementary human-computer-interaction work performs dissent-preserving aggregation directly in model training \citeapp{gordon2022jury}.

\noindent\textbf{Comparison with Contemporary Open Models}
\label{sec:recent_baselines}
The main experiments compare against open models contemporaneous with the contributed pool, whose base models are predominantly 2024-era: 47 of the 61 contributed models build on \textsc{Qwen2.5} or \textsc{Llama-3.1} (Figure \ref{fig:pool_diversity}D). To characterize where the paradigm stands relative to the current frontier, we additionally evaluate against four state-of-the-art open models released in 2026: \textsc{DeepSeek-V4-Pro}, \textsc{GPT-OSS-120B} \citeapp{openai2025gptoss}, \textsc{Kimi-K2.6}, and \textsc{Nemotron-3-Ultra}. We note these models are considerably larger than the individual contributed models, which average 5.32B parameters, and are among the best open models in 2026.

\begin{figure}[t]
\centering
\includegraphics[width=1.0\textwidth]{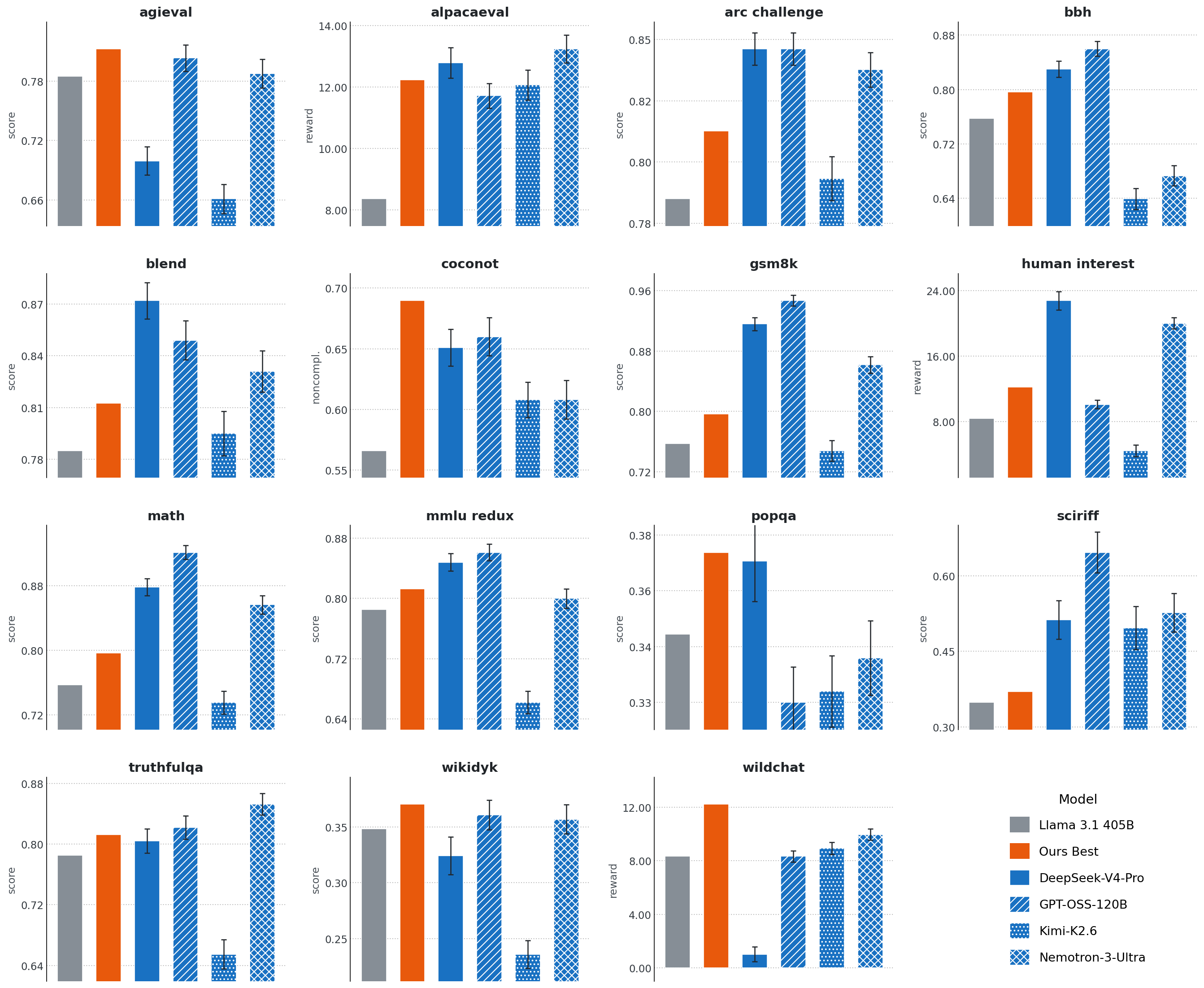}
\caption{\textbf{Comparison with contemporary open models.} Best collaborative system (orange) against \textsc{Llama-3.1-405B} (gray) and four 2026 state-of-the-art open models (blue) across all 15 tasks. Our systems, built from 2024-era contributed models averaging 5.32B parameters, outperform all contemporary baselines on AGIEval, CocoNot, PopQA, WikiDYK, and WildChat, while frontier models lead on reasoning-intensive tasks. Error bars represent 95\% confidence intervals.}
\label{fig:recent_baselines}
\end{figure}

Results in Figure \ref{fig:recent_baselines} show a consistent pattern. Our best collaborative systems outperform every contemporary baseline on AGIEval, CocoNot, PopQA, WikiDYK, and WildChat, that is, on tasks emphasizing safety, factual recall, and open-ended instruction following. On reasoning-intensive tasks such as GSM8k, MATH, and BigBench-Hard, the 2026 frontier models lead, reflecting the substantial reasoning-focused reinforcement learning applied in their post-training. A pool assembled from 2024-era academic models does not uniformly match 2026 frontier systems on those tasks. What the comparison does establish is that such a pool, collaborating, remains competitive with and on several axes exceeds models built with vastly greater resources. Since the scaling trends in Figure \ref{fig:big} have not saturated at 32 models and performance improves along every diversity axis we measure (Figure \ref{fig:model_diversity}), we expect participatory systems built from contemporary contributed models to track the frontier rather than trail it.

\noindent\textbf{Excluding 2025 Contributions}
\label{sec:only2024}
The contributed pool includes some models employing 2025 reasoning and reinforcement learning post-training recipes, which could in principle raise the effective ceiling of the collaborative system independently of collaboration itself. To separate these two effects, we repeat the scaling experiment using only contributed models from 2024, at pool sizes of 8 and 16, with the three strongest collaboration methods. If the gains reported in the main paper were driven by a small number of recent, individually strong contributions, restricting the pool in this way should erase them.

\begin{figure}[t]
\centering
\includegraphics[width=1.0\textwidth]{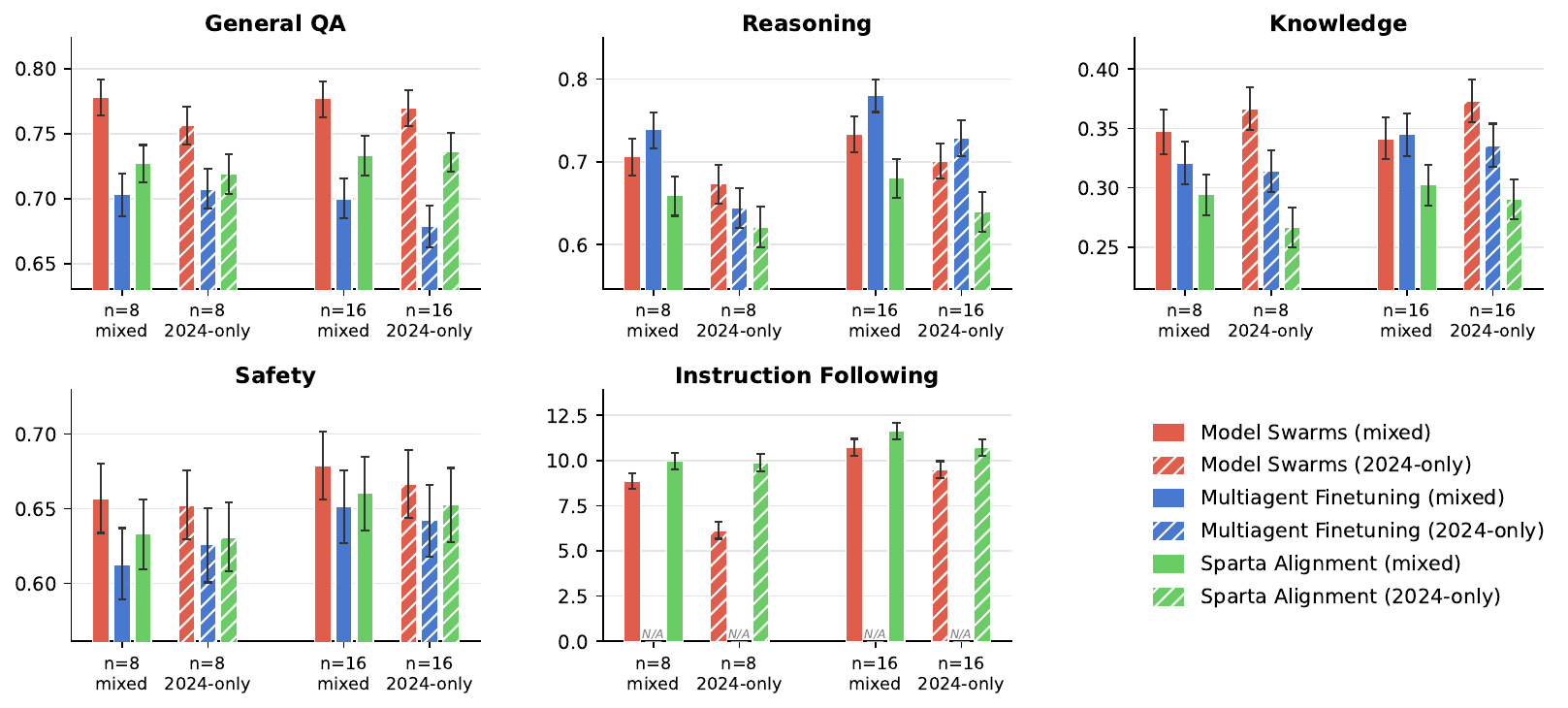}
\caption{\textbf{Excluding recent contributions.} Collaborative systems built from the full contributed pool (solid) versus only 2024 contributed models (hatched), at pool sizes of 8 and 16 across three collaboration methods. Multiagent finetuning requires objective ground truths and is therefore not applicable to instruction following. Most differences fall within overlapping 95\% confidence intervals, so the qualitative trends largely persist in the evaluated settings.}
\label{fig:only2024}
\end{figure}

Results in Figure \ref{fig:only2024} show that they largely persist. Safety is essentially unchanged across all three methods at both pool sizes, and general QA and knowledge are flat. Reasoning tasks degrade with this change: this is consistent with recent contributions carrying reasoning-specific capability, as one would expect given that many employ reasoning-oriented reinforcement learning. Across the remaining settings the differences fall within overlapping 95\% confidence intervals. The qualitative trends therefore largely persist in the evaluated settings when recent contributions are removed. Overlapping confidence intervals do not establish that recent contributions have no effect, and the reasoning decline is consistent with recent reasoning-focused contributions providing additional reasoning capability.

\begin{figure}[t]
\centering
\includegraphics[width=1.0\textwidth]{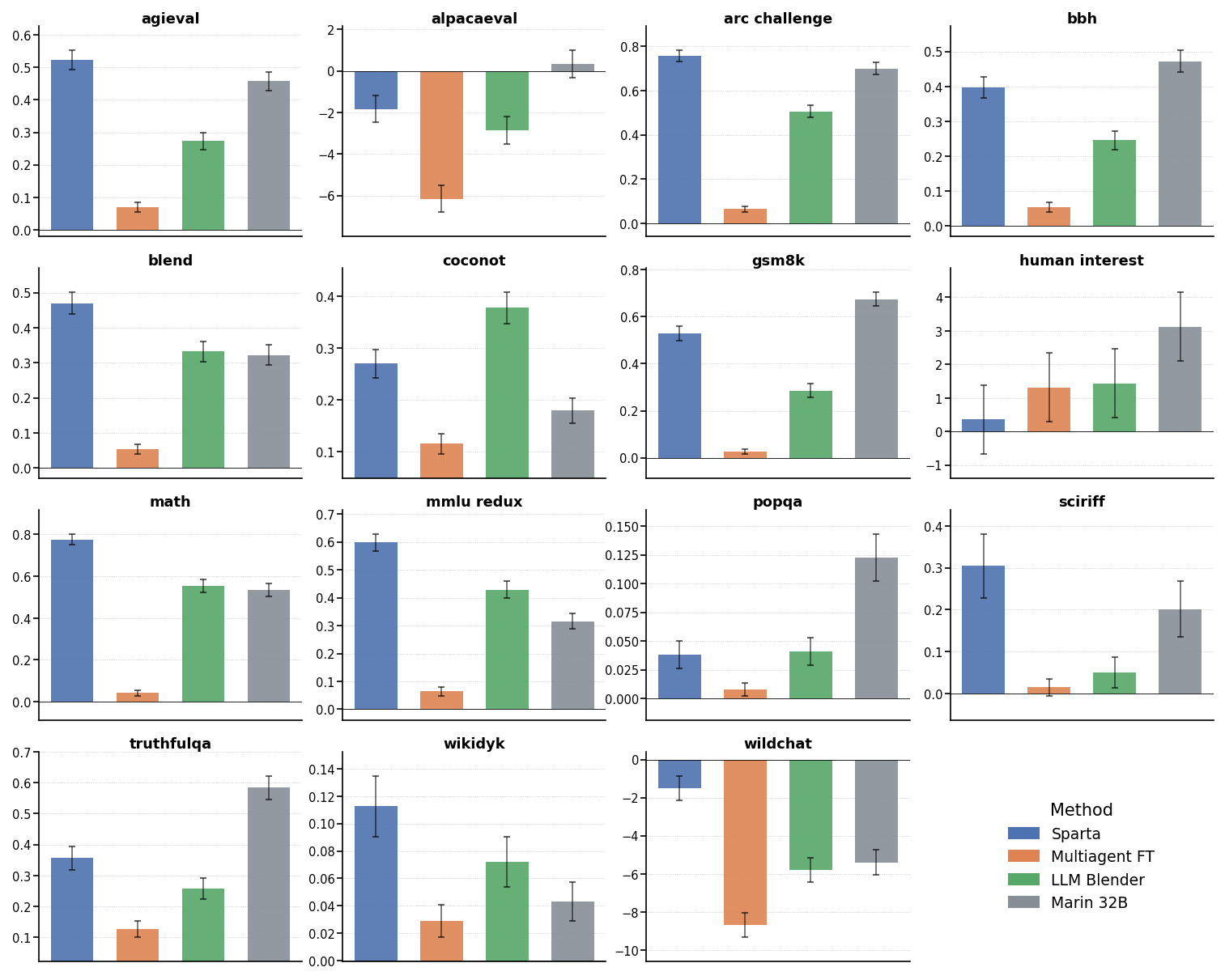}
\caption{\textbf{Comparison initialized from base checkpoints.} Three collaboration methods (\textsc{Sparta Alignment}, \textsc{LLM Blender}, \textsc{Multiagent Finetuning}) built on the four most common base models in our contributor pool, compared against Marin-32B-base across all 15 evaluation tasks. Our collaborative systems outperform Marin-32B-base on 9 out of 15 tasks. The collaboration methods add trained components (fine-tuning, protocol adaptation, and a trained ranker and fuser), so this is not a training-matched comparison and is reported as exploratory. Error bars represent 95\% confidence intervals.}
\label{fig:base_to_base}
\end{figure}

\noindent\textbf{Comparison Initialized from Base Checkpoints}
\label{sec:base_to_base}
Since the post-trained version of Marin-32B is not yet publicly available, a direct comparison with our post-trained collaborative systems on instruction-following and safety tasks would be unfair. We also note that several collaboration methods -- such as Sparta Alignment and Multi-Agent Finetuning -- inherently require models to critique, refine, and follow instructions as part of the collaboration protocol, capabilities that require post-training; comparing these against a base model thus understates their potential in the base-model setting. Nevertheless, we conduct an exploratory comparison initialized from base checkpoints, using the four most common base models in our contributor pool (\textsc{Qwen2.5-7B}, \textsc{Llama-3.1-8B}, \textsc{Mistral-7B-v0.1}, and \textsc{Gemma-2-9B}) to run three collaboration methods (Sparta Alignment, LLM Blender, Multiagent Finetuning) and comparing against Marin-32B-base across all 15 tasks. Results in Figure \ref{fig:base_to_base} show that our collaborative systems outperform Marin-32B-base on 9 out of 15 tasks. All three methods add trained components on top of the base checkpoints: fine-tuning in Multiagent Finetuning, protocol adaptation in Sparta Alignment, and a trained ranker and fuser in LLM Blender. The comparison is therefore initialized from base checkpoints rather than training-matched, and we report it as exploratory. We look forward to a more comprehensive comparison once the post-trained version of Marin-32B is publicly released.

\begin{table*}[t]
    \centering
    \scriptsize
    \setlength{\tabcolsep}{2pt}
    \renewcommand{\arraystretch}{1.1}
    \resizebox{1\linewidth}{!}{
    \begin{tabularx}{\linewidth}{X}
        \toprule[1.5pt]
        \textbf{Question}: A concert is given by a six-member band—guitarist, keyboard player, percussionist, saxophonist, trumpeter, violinist. During the concert, each member performs exactly one solo. The following restrictions apply: The guitarist does not perform the fourth solo. The percussionist performs a solo at some time before the keyboard player does. The keyboard player performs a solo at some time after the violinist does and at some time before the guitarist does. The saxophonist performs a solo at some time after either the percussionist does or the trumpeter does, but not both. Which one of the following CANNOT perform the third: \\
A) guitarist \\
B) keyboard player \\
C) saxophonist \\
D) trumpeter \\
E) violinist \\
        \\
        \textbf{Individual Model Answers}: C: 25 models \xmark, B: 6 models \xmark, E: 1 model \xmark \\
        \\
        \textbf{Participatory System Answer}: Certainly! Let's work through the ordering constraints. \\

Let P, V, K, G, S, and T denote the percussionist, violinist, keyboard player, guitarist, saxophonist, and trumpeter. \\

1. The percussionist performs before the keyboard player: $P < K$. \\
2. The violinist also performs before the keyboard player: $V < K$. \\
3. The keyboard player performs before the guitarist: $K < G$. \\

Because P and V are different musicians, both must perform before K. The guitarist must then perform after P, V, and K. At least three solos must precede the guitarist's, so the guitarist cannot perform third. The rule also expressly prevents the guitarist from performing fourth. \\

To check that the other answer choices really are possible, we can give a valid order for each: \\

- Keyboard player third: P, V, K, S, T, G. \\
- Saxophonist third: P, V, S, K, T, G. \\
- Trumpeter third: P, S, T, V, K, G. \\
- Violinist third: P, S, V, T, K, G. \\

In each order, P and V precede K, K precedes G, and G is not fourth. The saxophonist also performs after exactly one of P and T: S is after P but before T. \\

Thus, every listed musician except the guitarist can perform third. The correct answer is **A) Guitarist**. \cmark \\
        \bottomrule[1.5pt]
    \end{tabularx}
    }
    \vspace*{10pt}
    \caption{Example one of collaborative emergence, where participatory AI solves problems where all 32 modular components fail individually.}
    \label{tab:emergence_example_1}
\end{table*}

\begin{table*}[t]
    \centering
    \scriptsize
    \setlength{\tabcolsep}{2pt}
    \renewcommand{\arraystretch}{1.1}
    \resizebox{1\linewidth}{!}{
    \begin{tabularx}{\linewidth}{X}
        \toprule[1.5pt]
        \textbf{Question}: A maintenance company that takes service requests from three clients—Image, Solide, and Truvest—plans to set targets for its average service response times. Service targets will be set at 3 days, 2 days, or 1 day. Two service targets are set for each client—one for requests received through the maintenance company's website and one for requests received by voicemail. The six targets are set according to the following conditions: None of the clients can have a website target that is longer than its voicemail target. Image's voicemail target must be shorter than the other clients' voicemail targets. Solide's website target must be shorter than Truvest's website target. If none of the clients has a website target of 2 days, which one of the following could be true? \\
A) Image's website target is shorter than Solide's. \\
B) Solide's website target is shorter than Image's. \\
C) Solide's voicemail target is shorter than Truvest's. \\
D) Truvest's website target is shorter than Image's. \\
E) Truvest's voicemail target is shorter than Solide's. \\
\\
\textbf{Individual Model Answers}: E: 18 models \xmark, B: 11 models \xmark, A: 3 models \xmark \\
\\
\textbf{Participatory System Answer}: Certainly! We need to find which statement is possible while satisfying all the target-setting rules. \\

First, no website target is 2 days. A website target can therefore be either 1 day or 3 days. Solide's website target must be shorter than Truvest's, which fixes their website targets: Solide's is 1 day and Truvest's is 3 days. \\

A client's website target cannot be longer than its voicemail target. Since Truvest's website target is already 3 days, Truvest's voicemail target must also be 3 days. \\

Image's voicemail target must be shorter than both Solide's and Truvest's voicemail targets. In particular, it must be shorter than Truvest's 3-day target, so Image's voicemail target is either 1 or 2 days. Image's website target cannot exceed that voicemail target, and a 2-day website target is excluded. Thus, Image's website target must be 1 day. \\

Now consider the choices. Image's and Solide's website targets are both 1 day, ruling out A and B. Truvest's 3-day website target cannot be shorter than Image's 1-day target, ruling out D. Truvest's voicemail target is 3 days and cannot be shorter than Solide's target, ruling out E. \\

Choice C is possible. For example: \\

- Image: website 1 day; voicemail 1 day. \\
- Solide: website 1 day; voicemail 2 days. \\
- Truvest: website 3 days; voicemail 3 days. \\

Every condition holds, and Solide's 2-day voicemail target is shorter than Truvest's 3-day target. Therefore, the correct answer is **C) Solide's voicemail target is shorter than Truvest's**. \cmark \\
        \bottomrule[1.5pt]
    \end{tabularx}
    }
    \vspace*{10pt}
    \caption{Example two of collaborative emergence, where participatory AI solves problems where all 32 modular components fail individually.}
    \label{tab:emergence_example_2}
\end{table*}

\begin{table*}[t]
    \centering
    \scriptsize
    \setlength{\tabcolsep}{2pt}
    \renewcommand{\arraystretch}{1.1}
    \resizebox{1\linewidth}{!}{
    \begin{tabularx}{\linewidth}{X}
        \toprule[1.5pt]
        \textbf{Question}: A software company employs exactly seven sales representatives—Kim, Mahr, Parra, Quinn, Stuckey, Tiao, and Udall—to work in its three sales zones—Zone 1, Zone 2, and Zone 3. Each sales representative works in exactly one of the sales zones, in accordance with the following conditions: Either Parra or Tiao (but not both) works in Zone 1. Either Tiao or Udall (but not both) works in Zone 2. Parra and Quinn work in the same sales zone as each other. Stuckey and Udall work in the same sales zone as each other. There are more of the sales representatives working in Zone 3 than in Zone 2. Quinn CANNOT work in the same sales zone as which one of the following? \\
A) Kim \\
B) Mahr \\
C) Stuckey \\
D) Tiao \\
E) Udall \\

\textbf{Individual Model Answers}: C: 12 models \xmark, E: 11 models \xmark, A: 7 models \xmark, B: 2 models \xmark \\
\\
\textbf{Participatory System Answer}: Certainly! Let's use the zone rules to determine who cannot work with Quinn. \\

Parra and Quinn must work in the same zone. Thus, anyone who cannot work with Parra also cannot work with Quinn. \\

The Zone 1 rule says that exactly one of Parra and Tiao works there. Parra and Tiao therefore cannot share Zone 1, since that would put both of them there. They also cannot share Zone 2 or Zone 3, since that would leave neither of them in Zone 1. Consequently, Parra and Tiao must always work in different zones. Because Quinn works with Parra, Quinn must also work in a different zone from Tiao. \\

We can check that the other four choices are possible. To place Quinn with Kim and Mahr, put Parra, Quinn, Kim, and Mahr in Zone 3; Tiao in Zone 1; and Stuckey and Udall in Zone 2. Exactly one of Parra and Tiao is in Zone 1, exactly one of Tiao and Udall is in Zone 2, both required pairs work together, and Zone 3 has four representatives compared with Zone 2's two. \\

To place Quinn with Stuckey and Udall, put Parra, Quinn, Stuckey, and Udall in Zone 1; Tiao in Zone 2; and Kim and Mahr in Zone 3. This also satisfies both ``exactly one'' rules and both pairing rules. Zone 3 has two representatives, more than Zone 2's one. \\

Kim, Mahr, Stuckey, and Udall can therefore each work with Quinn. Tiao cannot. The correct answer is **D) Tiao**. \cmark \\
        \bottomrule[1.5pt]
    \end{tabularx}
    }
    \vspace*{10pt}
    \caption{Example three of collaborative emergence, where participatory AI solves problems where all 32 modular components fail individually.}
    \label{tab:emergence_example_3}
\end{table*}

\begin{table}[htbp]
    \centering
    \caption{List of the 32 contributed models we employed in Figure \ref{fig:big}, part 1. All model names are identifiers on Huggingface, a model sharing platform. Model descriptions are provided by the original contributors.}
    \label{tab:main_32_1}
    \begin{tabularx}{\textwidth}{
        c 
        >{\hsize=0.66\hsize\raggedright\arraybackslash}X 
        >{\hsize=1.34\hsize\raggedright\arraybackslash}X 
        l 
    }
        \toprule
        \textbf{ID} & \textbf{Model} & \textbf{Description} & \textbf{Paper} \\
        \midrule
        
        1 & 
        chtmp223/Qwen2.5-7B-CLIPPER & 
        We fine-tuned models for the task of narrative claim verification, where the model has to verify whether a claim about a book is True or False given the full book text. & 
        \citeapp{phamclipper} \\
        \addlinespace 
        
        2 & 
        chengq9/ToolRL-Qwen2.5-3B & 
        The model is trained through RL especially for tool use task, focusing on tool learning, parameter filling, etc. & 
        \citeapp{qiantoolrl} \\
        \addlinespace

        3 & 
        AgentFlow/agentflow-planner-7b &
        The framework directly optimizes planner agent within the system in an online fashion using Flow-based Group Refined Policy Optimization (Flow-GRPO) &
        \citeapp{li2025flow} \\
        \addlinespace

        4 &
        nanami/ladder-last16L-llama3.1-8binstruct-sft4k-stage2v03-bsize32-rkl8b &
        Finetuned from llama3.1 to test an efficient language model architecture design. Finetuning enables us to test the architecture on a relatively large model while avoiding expensive train-from-scratch; this is very helpful for architecture research that doesn't deviate too much from the standard Transformer. &
        \citeapp{zhang2025ladder} \\
        \addlinespace

        5 &
        viswavi/qwen2.5\_rlcf &
        This model improves the instruction following capabilities of Qwen-2.5-7B-Instruct using preference tuning on the WildChecklists dataset. This model is described in detail in "Checklists Are Better Than Reward Models For Aligning Language Models". &
        \citeapp{viswanathanchecklists} \\
        \addlinespace

        6 &
        milli19/promptmii-llama-3.1-8b-instruct &
        RL trained for ability to induce instructions &
        \citeapp{xiao2025prompt} \\
        \addlinespace

        7 &
        Zhengping/conditional-probability-regression &
        Estimating conditional probability in everyday scenario &
        \citeapp{wangalways} \\
        \addlinespace

        8 &
        yale-nlp/MDCure-Qwen2-7B-Instruct &
        Model post-trained for various multi-document tasks (e.g., QA, summarization, coreference resolution) that achieves notable improvements in multi-document performance over prior pre-trained baselines. Gains generalize robustly across task/content domains, and existing capabilities of the base model are preserved. &
        \citeapp{liu2025mdcure} \\
        \addlinespace

        9 &
        GritLM/GritLM-7B &
        Generative representational instruction tuned to tackle both generative \& embedding tasks. &
        \citeapp{muennighoff2024generative} \\
        \addlinespace

        10 &
        lime-nlp/Qwen2.5-7B-Instruct-SUM10 &
        The model is trained to reason about its own uncertainty and perform abstention when needed. &
        \citeapp{song2025hallucination} \\
        \addlinespace

        11 &
        geyang627/care-chinese-gemma2-9b &
        care-chinese-gemma2-9b is based on google/gemma-2-9b-it and further fine-tuned on our CARE, enhancing cultural awareness especially in Chinese. &
        \citeapp{guo2025care} \\
        \addlinespace

        12 &
        bespokelabs/Bespoke-Stratos-7B &
        The model is trained for efficient LLM hallucination detection &
        \citeapp{tang2024minicheck} \\
        \addlinespace

        13 & 
        kangdawei/Llama-3.1-8B-Instruct-GenderNeutral-Finetuned &
        Trained on gender neutral dataset for gender bias mitigation &
        \citeapp{wei2025mitigating} \\
        \addlinespace

        14 & 
        DeepRetrieval/DeepRetrieval-PubMed-3B-Llama & 
        The models are trained for rewriting the query for optimized retrieval outcome with BM25 or real search engines (PubMed). More models can be found at https://huggingface.co/DeepRetrieval &
        \citeapp{jiangdeepretrieval} \\
        \addlinespace

        15 &
        yale-nlp/MDCure-Qwen2-1.5B-Instruct &
        Model post-trained for various multi-document tasks (e.g., QA, summarization, coreference resolution) that achieves notable improvements in multi-document performance over prior pre-trained baselines. Gains generalize robustly across task/content domains, and existing capabilities of the base model are preserved. &
        \citeapp{liu2025mdcure} \\
        \addlinespace

        16 &
        Zhaoxuan/PUGC-Mistral-DPO &
        model that aligned with user preferences using preference data souring from user-generated content &
        \citeapp{tan2025aligning} \\
        \addlinespace
        
        \bottomrule
    \end{tabularx}
\end{table}

\begin{table}[htbp]
    \centering
    \caption{List of the 32 contributed models we employed in Figure \ref{fig:big}, part 2. All model names are identifiers on Huggingface, a model sharing platform. Model descriptions are provided by the original contributors.}
    \label{tab:main_32_2}
    \begin{tabularx}{\textwidth}{
        c 
        >{\hsize=0.66\hsize\raggedright\arraybackslash}X 
        >{\hsize=1.34\hsize\raggedright\arraybackslash}X 
        l 
    }
        \toprule
        \textbf{ID} & \textbf{Model} & \textbf{Description} & \textbf{Paper} \\
        \midrule
        
        17 & 
        jwhj/Qwen2.5-Math-1.5B-OREO & 
        A model for math reasoning, trained with OREO (offline reasoning optimization) & 
        \citeapp{wang2025offline} \\
        \addlinespace 
        
        18 &
        PeterJinGo/SearchR1-nq\_hotpotqa\_train-qwen2.5-7b-em-ppo &
        Agentic search &
        \citeapp{jinsearch} \\
        \addlinespace

        19 &
        gasolsun/DynamicRAG-8B &
        The model is trained as a dynamic reranker agent that can determine the appropriate number and order of the retrieved documents. &
        \citeapp{sundynamicrag} \\
        \addlinespace

        20 &
        LLM360/guru-7B &
        A model for general reasoning, trained with GURU dataset &
        \citeapp{chengrevisiting} \\
        \addlinespace

        21 &
        spiral-rl/Spiral-Qwen3-4B &
        This model, SPIRAL, is trained on for multi-agent multi-turn settings to perform complex reasoning tasks by self-play in strategic games like Poker and Tic-Tac-Toe. This self-play method enables it to autonomously develop cognitive skills that transfer to broader domains like math and logic. &
        \citeapp{liu2025spiral} \\
        \addlinespace

        22 &
        uclanlp/brief-pro &
        It is a universal, lightweight compressor that distills relevant evidence for a given query from retrieved documents into a concise summary for seamless integration into in-context RAG. &
        \citeapp{gu2025brief} \\
        \addlinespace

        23 & 
        Rakancorle1/PolicyGuard-4B & 
        PolicyGuard-4B is a 4-billion-parameter lightweight guardrail model trained on PolicyGuardBench, a 60k-example benchmark of policy–trajectory pairs across five web domains. It is designed to detect policy violations in autonomous web agent trajectories, including both full-sequence and prefix-based detection tasks, achieving strong accuracy and cross-domain generalization while maintaining efficient inference &
        \citeapp{wen2025towards} \\
        \addlinespace

        24 &
        ReasoningTransferability/ \newline UniReason-Qwen3-14B-RL &
        This model is a RL-GRPO-tuned version of qwen3-14b focused on math-reasoning capabilities. The model was developed as part of research investigating the transferability of mathematical reasoning skills to general language tasks. &
        \citeapp{huan2025does} \\
        \addlinespace

        25 &
        ypwang61/One-Shot-RLVR-Qwen2.5-Math-1.5B-pi1 &
        Training 1.5B model with one training example to show some features of RL training, like robust to overfitting due to zero-mean advantage, and supports that to a large extent RLVR incentivizes the reasoning capability obtained from pretrain/mid-train stages. &
        \citeapp{wangreinforcement} \\
        \addlinespace

        26 &
        sunblaze-ucb/Qwen2.5-3B-Intuitor-MATH-1EPOCH &
        This model is an Intuitor-fine-tuned version of Qwen2.5-3B trained on the MATH dataset. Intuitor is a reinforcement learning method that fine-tunes large language models (LLMs) using self-certainty—the model’s own internal confidence—as the sole reward. It is built on a novel paradigm we call Reinforcement Learning from Internal Feedback (RLIF). &
        \citeapp{zhao2025learning} \\
        \addlinespace

        27 &
        yale-nlp/Qwen3-8B-SciLit01 &
        A thinking model trained on Synthetic-1, specialized for scientific/STEM tasks. &
        \citeapp{lidemystifying} \\
        \addlinespace

        28 &
        OpenThoughts/OpenThinker3-7B &
        Open source reasoning model trained on OpenThoughts dataset &
        \citeapp{guhaopenthoughts} \\
        \addlinespace

        29 &
        l3lab/L1-Qwen-1.5B-Exact &
        A series of reasoning models trained using reinforcement learning for length control, i.e user can specify desired output length in prompt and model will adhere to it. All models: https://huggingface.co/collections/l3lab/l1-67cacf4e39c176ca4e9890f4 &
        \citeapp{aggarwall1} \\
        \addlinespace
        
        \bottomrule
    \end{tabularx}
\end{table}

\begin{table}[htbp]
    \centering
    \caption{List of the 32 contributed models we employed in Figure \ref{fig:big}, part 3. All model names are identifiers on Huggingface, a model sharing platform. Model descriptions are provided by the original contributors.}
    \label{tab:main_32_3}
    \begin{tabularx}{\textwidth}{
        c 
        >{\hsize=0.66\hsize\raggedright\arraybackslash}X 
        >{\hsize=1.34\hsize\raggedright\arraybackslash}X 
        l 
    }
        \toprule
        \textbf{ID} & \textbf{Model} & \textbf{Description} & \textbf{Paper} \\
        \midrule

        30 &
        fangwu97/DeepSearch-1.5B &
        DeepSearch-1.5B is a 1.5B parameter reasoning model trained with Reinforcement Learning with Verifiable Rewards (RLVR), enhanced by Monte Carlo Tree Search (MCTS). &
        \citeapp{wu2025deepsearch} \\
        \addlinespace

        31 & 
        allegrolab/hubble-8b-500b\_toks-standard-hf &
        The suite of models are fine-tuned from scratch using the Llama architecture. The models are trained on modified datasets to study LLM memorization. The 8B parameter model is trained to 3.7x Chinchilla optimal using 500B tokens. We also release 1B models. &
        \citeapp{wei2025hubble} \\
        \addlinespace

        32 & 
        fcyin/llama2\_7B\_base \newline \_lofit\_truthfulqa &
        This is a Llama-2-7b model fine-tuned on TruthfulQA using Localized Fine-tuning on LLM Representations (LoFiT; https://arxiv.org/abs/2406.01563). This model checkpoint modifies the attention outputs of 96 attention heads (10\% of all attention heads). &
        \citeapp{yin2024lofit} \\
        \addlinespace
        
        \bottomrule
    \end{tabularx}
\end{table}

\begin{table}[htbp]
    \centering
    \caption{List of other contributed models, part 1. All model names are identifiers on Huggingface, a model sharing platform. Model descriptions are provided by the original contributors.}
    \label{tab:others_1}
    \begin{tabularx}{\textwidth}{
        c 
        >{\hsize=0.66\hsize\raggedright\arraybackslash}X 
        >{\hsize=1.34\hsize\raggedright\arraybackslash}X 
        l 
    }
        \toprule
        \textbf{ID} & \textbf{Model} & \textbf{Description} & \textbf{Paper} \\
        \midrule

        33 &
        multimodal-reasoning-lab/Bagel-Zebra-CoT &
        The model is trained on our released Zebra-CoT benchmark. &
        \citeapp{li2025zebra} \\
        \addlinespace

        34 &
        dill-lab/oath-frames-flant5-xl &
        Our model was finetuned to classify social media posts with labels for attitudes towards homelessness. &
        \citeapp{ranjit2024oath} \\
        \addlinespace

        35 &
        tsor13/spectrum-Llama-3.1-8B-v1 &
        Trained for distributional data modeling / steerability &
        \citeapp{sorensenspectrum} \\
        \addlinespace

        36 &
        lecslab/glosslm &
        This model is trained to take as input transcriptions from a language (and/or free translation) and output interlinear gloss (IGT), a form of linguistic annotation used to study the morphosyntactic patterns of a language. &
        \citeapp{ginn2024glosslm} \\
        \addlinespace

        37 &
        UW/OLMo2-8B-SuperBPE-t180k &
        Trained with SuperBPE tokenizer from scratch to test effect of tokenizer &
        \citeapp{liusuperbpe} \\
        \addlinespace

        38 & 
        MLLMMU/LLaVA\_Vanilla &
        We fine-tune a model on our fictitious unlearning benchmark to establish a vanilla baseline for evaluating the effectiveness of different unlearning algorithms. Specifically, these algorithms are applied to the vanilla model, and their performance is compared to assess unlearning quality. &
        \citeapp{liu2025protecting} \\
        \addlinespace

        39 &
        ddvd233/QoQ-Med-VL-7B &
        We tuned Qwen2.5 VL on multimodal medical diagnosis data across 2D/3D images, videos and time series. &
        \citeapp{daiqoq} \\
        \addlinespace

        40 &
        ylu610/FT-T5-3B-Verifier &
        The T5-3B model finetuned on FActScore dataset for the claim verification task. The model is used in our paper: Optimizing Decomposition for Optimal Claim Verification. &
        \citeapp{lu2025optimizing} \\
        \addlinespace

        41 &
        reasonir/ReasonIR-8B &
        The model is trained for reasoning-intensive retrieval. We performed fine-tuning with a contrastive learning objective based on Llama3.1-8B. Note: The model is a language model but is specially adapted to be used as an embedding model in retrieval tasks, so it may not fit if the scope is strictly a language model for generation tasks. &
        \citeapp{shaoreasonir} \\
        \addlinespace

        42 &
        nbalepur/Llama-3.1-8B-PT-DPO-BeaverTails &
        Trained on the BeaverTails dataset to respond to user queries, but also can take in a user persona as an input (e.g. "I am a vegetarian") to personalize the response &
        \citeapp{balepur2025whose} \\
        \addlinespace

        43 &
        guangyangmusic/legato &
        We train a VLM for optical music recognition, i.e., recognizing music scores from images. The vision encoder is adapted from Llama-3.2-11B-Vision and the text decoder is trained from scratch. &
        \citeapp{yanglegato} \\
        \addlinespace

        44 &
        Shiyu-Lab/roberta-base-watermark-embed &
        The model is trained to generate text embeddings that are insensitive to semantic-preserving changes but sensitive to semantic-distorting changes. &
        \citeapp{andefending} \\
        \addlinespace

        45 &
        JiaaqiLiu/VIPER-R1-RL &
        The model is trained to generate text embeddings that are insensitive to semantic-preserving changes but sensitive to semantic-distorting changes. &
        \citeapp{liu2025mimicking} \\
        \addlinespace

        46 &
        l3-unc/qwen2.5-7b\_edited\_fact\_check\_v2 &
        These models are knowledge edited models using MEMIT to generate counterfactuals models for faithfulness metric meta-evaluation. &
        \citeapp{zaman2025causal} \\
        \addlinespace

        47 &
        Jennny/llama3\_8b \newline \_helpful\_rm\_full &
        The models are trained as Bradley-Terry reward models on top of a SFT meta-llama/Llama-3.1-8B model to provide reward scores across helpfulness, honesty, and truthfulness dimensions. &
        \citeapp{shen2025simultaneous} \\
        \addlinespace

        48 &
        KaiserWhoLearns/ \newline PTvsSFT\_OLMo1b &
        This is the model checkpoint release for Amuro \& Char: Analyzing the Relationship between Pre-Training and Fine-Tuning of Large Language Models, each model checkpoint is fine-tuned for one task. &
        \citeapp{sun2025amuro} \\
        \addlinespace
        
        \bottomrule
    \end{tabularx}
\end{table}

\begin{table}[htbp]
    \centering
    \caption{List of other contributed models, part 2. All model names are identifiers on Huggingface, a model sharing platform. Model descriptions are provided by the original contributors.}
    \label{tab:others_2}
    \begin{tabularx}{\textwidth}{
        c 
        >{\hsize=0.66\hsize\raggedright\arraybackslash}X 
        >{\hsize=1.34\hsize\raggedright\arraybackslash}X 
        l 
    }
        \toprule
        \textbf{ID} & \textbf{Model} & \textbf{Description} & \textbf{Paper} \\
        \midrule

        49 &
        LMMs-Lab-Turtle/SelfRewarded-R1-7B &
        A model trained specialized for generating visual caption before generating the final answer. &
        \citeapp{li2025self} \\
        \addlinespace

        50 &
        allenai/OLMo-2-0325-32B-Instruct &
        Pretrained on 6 trillion tokens of web text, academic papers, code, math-heavy content, and more. Instruction tuned on Tulu 3 data, including for general SFT, DPO, RLVR. &
        \citeapp{olmo20242} \\
        \addlinespace

        51 &
        yale-nlp/MDCureRM &
        This is a quality scoring model for multi-document post-training data. It can be used in a RM-as-a-Judge fashion to determine the training utility of a particular multi-document QA sample; scoring is based on a predefined rubric with 6 criteria. &
        \citeapp{liu2025mdcure} \\
        \addlinespace

        52 &
        yale-nlp/MDCure-LLAMA3.1-8B-Instruct &
        Post-training adaptation of Llama3.1-8B-Instruct for various multi-document tasks (e.g., QA, summarization, coreference resolution) that achieves notable improvements in multi-document performance over prior pre-trained baselines. Gains generalize robustly across task/content domains. &
        \citeapp{liu2025mdcure} \\
        \addlinespace

        53 &
        izzcw/minecollab\_ \newline filtered\_crafting &
        I trained the model to collaborate better in long context embodied tasks (e.g. Minecraft) by using tools and learning to share information. &
        \citeapp{white2025collaborating} \\
        \addlinespace

        54 &
        kkkevinkkk/Llama-3-8B-CR-DPO &
        This is trained to resolve knowledge conflicts between contextual knowledge and parametric knowledge in the RAG set-up. &
        \citeapp{huangtrust} \\
        \addlinespace

        55 &
        chentong00/Qwen3-8B-GRPO-Binary-RAR &
        The model is finetuned on Qwen3-8B to reduce extrinsic hallucination. We use RL (GRPO) with a binary retrieval-augmented reward and train on the chentong00/binary-rar-wildchat-8k dataset. &
        \citeapp{chen2025train} \\
        \addlinespace

        56 &
        Shiyu-Lab/HarnessLLM \newline \_RL\_Qwen3\_4B &
        The model was trained with RL for automatic test cases generation. &
        \citeapp{liu2025harnessllm} \\
        \addlinespace

        57 &
        CNCL-Penn-State/CrPO-llama-3.1-8b-instruct-cre &
        The model is optimized for better output creativity. &
        \citeapp{ismayilzada2025creative} \\
        \addlinespace

        58 &
        hkust-nlp/drkernel-14b &
        Triton Kernel Generations &
        \citeapp{liu2026dr} \\
        \addlinespace

        59 &
        visual-ds/distilbert-base-uncased-hatexplain &
        The rationales of the HateXplain dataset contain words not included in the original `distilbert-base-uncased` model’s vocabulary because they are offensive and hate speech words. However, when training a model to incorporate rationales, including these tokens in the vocabulary may be important. Otherwise, the results would be underestimated. In the train portion of the dataset, we filtered the most popular out-of-vocabulary tokens (those with more than ten occurrences), added them to the models’ vocabularies, and fine-tuned the model in this portion. We used a masked language modeling probability of 0.15 with a batch size of 8 for 15 epochs in a GPU NVIDIA GeForce GTX 1070. &
        \citeapp{resck2024exploring} \\
        \addlinespace

        60 &
        oaimli/qwen\_sft\_scitrek &
        We fine-tuned Qwen2.5-7B-Instruct-1M on our constructed dataset SciTrek (https://arxiv.org/abs/2509.21028), to improve the performance on long-context numerical reasoning over scientific articles. &
        \citeapp{li2025gets} \\
        \addlinespace

        61 &
        bkhmsi/micro-llama-1b &
        We partition the layers of a pretrained language model into four expert modules aligned with well-studied cognitive networks in the human brain. The result is a more interpretable and steerable model that is more aligned to human behavior without sacrificing performance on reasoning benchmarks. &
        \citeapp{alkhamissi2025mixture} \\
        \addlinespace
        
        \bottomrule
    \end{tabularx}
\end{table}

\clearpage
\bibliographystyleapp{sn-nature}
\bibliographyapp{sn-bibliography}

\end{document}